\documentclass[
 amsmath,
 amssymb,
 aps,
]{revtex4-2}

\usepackage{graphicx} 
\usepackage{dcolumn} 
\usepackage{bm} 

\usepackage{hyperref} 

\usepackage{ascmac}

\usepackage[linesnumbered, ruled, vlined]{algorithm2e}

\usepackage{xcolor}

\DeclareMathOperator*{\argmin}{arg\,min}

\usepackage{slashed}

\usepackage{tikz}
\usetikzlibrary{quantikz}

\allowdisplaybreaks

\usepackage{MnSymbol}

\usepackage{soul}

\makeatletter
\newcommand{\vast}{\bBigg@{3.0}}
\newcommand{\Vast}{\bBigg@{4.0}}

\makeatother

\usepackage{accents}
\newlength{\dhatheight}

\usepackage[whole]{bxcjkjatype}

\usepackage{bbold}

\usepackage{amsthm}

\theoremstyle{definition}

\theoremstyle{remark}

\usepackage[compat=1.1.0]{tikz-feynhand}

\usepackage{simpler-wick}

\usepackage{braket}

\usepackage{xr}
\begin{document}

\preprint{APS/123-QED}

\title{Quantum Advantage in Variational Bayes Inference}

\author{Hideyuki Miyahara}
\email{miyahara@g.ucla.edu, hmiyahara512@gmail.com}

\affiliation{
Department of Electrical and Computer Engineering, \\
Henry Samueli School of Engineering and Applied Science, \\
University of California, Los Angeles, California 90095
}


\author{Vwani Roychowdhury}

\email{vwani@g.ucla.edu}

\affiliation{
Department of Electrical and Computer Engineering, \\
Henry Samueli School of Engineering and Applied Science, \\
University of California, Los Angeles, California 90095
}

\date{\today}

\begin{abstract}
  Variational Bayes (VB) inference algorithm is used widely to estimate both the parameters and the unobserved hidden variables in generative statistical models. The algorithm --inspired by variational methods used in computational physics-- is iterative and can get easily stuck in local minima, even when classical techniques, such as deterministic annealing (DA), are used.
  We study a variational Bayes (VB) inference algorithm based on a non-traditional quantum annealing approach -- referred to as quantum annealing variational Bayes (QAVB) inference -- and show that there is indeed a quantum advantage to QAVB over its classical counterparts.
  In particular, we show that such better performance is rooted in key concepts from quantum mechanics: (i) the ground state of the Hamiltonian of a quantum system -- defined from the given variational Bayes (VB) problem -- corresponds to an optimal solution for the minimization problem of the variational free energy at very low temperatures; (ii) such a ground state can be achieved by a technique paralleling the quantum annealing process; and (iii) starting from this ground state, the optimal solution to the VB problem can be achieved by increasing the heat bath temperature to unity, and thereby avoiding local minima introduced by spontaneous symmetry breaking observed in classical physics based VB algorithms.
  We also show that the update equations of QAVB can be potentially implemented using $\lceil \log K \rceil$ qubits and $\mathcal{O} (K)$ operations per step. Thus, QAVB can match the time complexity of existing VB algorithms, while delivering higher performance.
\end{abstract}


\maketitle


\section{Introduction}
Quantum machine learning (QML) primarily deals with quantum algorithms and quantum-inspired algorithms for data analysis, and is an emerging research field that is forming new bridges between the traditional fields of physics and machine learning.
Several QML frameworks, such as quantum principal component analysis (qPCA)~\cite{Lloyd_001} and quantum recommendation systems~\cite{Kerenidis_001}, have been introduced that show significant quantum speedups while achieving the same performance as the corresponding classical algorithms.
These quantum algorithms, in turn, were later shown to have classical counterparts, and randomized algorithms with the same time complexity were derived~\cite{Tang_001, Tang_002}. This discovery process showed an encouraging synergy where the principles of quantum mechanics can also facilitate the design of better classical algorithms.
Interest in QML has also been fueled by the emergence of noisy intermediate-scale quantum (NISQ) devices.
The low fidelity and limited scale of such devices prevent implementations of well-known algorithms such as the Shor's factorization algorithm or combinatorial optimization algorithms based on quantum annealing. However, efficient QML algorithms for conventional ML tasks such as, dimensionality reduction, clustering, classification, and Bayesian inference, could likely be implemented on NISQ devices and show potential quantum advantages in speed or accuracy.
For example, variational quantum classifiers (VQC) and quantum circuit learning (QCL) frameworks have been proposed that hold the promise of time and hardware efficient training and realizations of conventional classifiers \cite{Mitarai_001, Schuld_001, Miyahara_006}.
Recent results, however, show that simple kernel method based classical classifiers are guaranteed to have better performance than their quantum counterparts.
Furthermore, there is no analytical guarantee or numerical evidence suggesting that the variational quantum algorithms will even have reasonable performance, especially for high-dimensional datasets where these algorithms are expected to have speedup advantages.

The above mentioned examples underscore the general trend in the QML field: existing algorithms can potentially speed up classical algorithms, but \textit{QML algorithms that outperform their classical ML counterparts are very rare} or nonexistent.
Thus, the search for QML algorithms that either perform better than any classical algorithm without incurring significant computational overheads, or exhibit significant speedups (for the same performance) continues to remain an active area of interest.

In this paper, we address the problem of variational Bayes (VB) inference, which is a popular technique in ML, and explore how quantum mechanics can help design an algorithm with better performance than the existing classical techniques.
In fact, principles from classical statistical physics have already inspired a genre of algorithms for VB.
The history of optimization algorithms motivated by physics dates back to simulated annealing (SA)~\cite{Kirkpatrick_001}, which utilizes a thermostat to overcome the local optima problem in optimization and SA has been applied to several ML tasks~\cite{Lavielle_001, Albert_001}.
SA approaches, however, have a well-known drawback in that it requires an infinitely long annealing schedule to guarantee the global optimum of an optimization problem or at least a very long annealing schedule to reach its equilibrium state at a finite temperature.
To fix such drawbacks of SA, deterministic annealing (DA) was developed and applied to several machine learning problems ~\cite{Rose_001}.
For example, by applying DA to variational Bayes (VB) inference~\cite{Bishop_001, Murphy_001}, deterministic annealing variational Bayes (DAVB) inference~\cite{Katahira_001} was proposed.
However, it has been shown that DA and DAVB can get stuck in a local optima relatively easily or a saddle point (as shown in Fig.~\ref{main_numerical_001_001}(c)), and details of this phenomenon are further discussed later in this paper.

More recently, by following the trend in QML, a quantum annealing variational Bayes (QAVB) inference framework -- a quantum-mechanical extension of VB and DAVB -- was proposed in~\cite{Miyahara_002}, and the study showed that QAVB outperformed both VB and DAVB in several numerical examples. Other than numerical examples, concrete mechanisms that enable QAVB to achieve better performance than its classical counterparts were not given, and numerical results providing evidence for such potential mechanisms were  not presented. Moreover, if QAVB is implemented classically then each iteration step requires $\mathcal{O} (K^3)$ operations (where the categorical hidden variable in the VB problem has $K$ possible values), as compared to $\mathcal{O} (K)$ computations required by classical VB algorithms. Thus, any performance enhancements offered by QAVB seems to have an associated computational price. This increased computational cost stems from classical simulations of a quantum system, which requires repeated diagonalization of the underlying Hamiltonian. Thus, a natural question, especially in the context of QML, is whether the update equations of QAVB can be simulated using quantum devices, where no such diagonalization would be necessary.

In this paper, we first introduce the VB problem.
Then we explain the motivation behind the incorporation of a non-traditional quantum annealing (QA) approach, and the framework of QAVB and formulate a mechanism by which QAVB could show better performance than both VB and DAVB. As in the traditional QA case, our non-traditional QA considers the evolution of a quantum system under a time-varying Hamiltonian; however, the evolution dynamics is now driven by relaxation under the mean-field (MF) approximation, as opposed to relaxation under the Schr\"{o}dinger equation.
We provide both numerical evidence and analytical proofs supporting this mechanism. From an analytical perspective, we show how the ground-state dynamics introduced by our non-traditional QA can also be analyzed by techniques similar to those used in the well-known adiabatic theorem that characterizes the traditional QA process where the Hamiltonian is time-varying. In order to support our theoretical and mechanism related results, we provide numerical results on two synthetic datasets, created using a generative model, where all the hidden variables and parameters are specified. This allows us to compare the performance of any algorithm to that of the ground truth optimal solutions. As predicted, our results show that QAVB (a single run, independent of initial conditions) finds good estimates that are very close to that of the underlying generative models, but VB and DAVB find them with low probability.  Moreover, these numerical results  show that the QA part of QAVB is critically important for optimal parameter estimation and is the key to obtaining better performance than classical algorithms. For results on  higher dimensional datasets where QAVB outperforms VB and DAVB please refer to Ref.~\cite{Miyahara_002}.
Then, we show that the QAVB update steps are completely positive and trace-preserving (CPTP) maps. Since it is known that a CPTP map can be implemented on quantum systems, we thus show that QAVB can be implemented using NISQ devices, comprising only $\lceil \log K \rceil$ qubits.

\section{Variational Bayes inference}
Suppose that we have $N$ observable data points $Z^\mathrm{obs} \coloneqq \{ z_i^\mathrm{obs} \}_{i=1}^N$ that are the output of an unknown generative model $p_\mathrm{gen}^z (z)$: $z_i^\mathrm{obs} \sim p_\mathrm{gen}^z (\cdot)$ with additional dynamics that are not necessarily observed.
One of the important approaches in ML is to assume that the generative model can be well approximated by a parameterized model that outputs both the observable data points $Z^\mathrm{obs}$, as well as an associated set of unobservable or hidden data points, $\Sigma \coloneqq \{ \sigma_i \}_{i=1}^N$, where $\sigma_i \in \{ 1, 2, \dots, K \}$ is a categorical variable with $K$ outcomes.
These hidden variables often have interpretable meanings and can be used to predict other outcomes associated with the dataset. The task then is to estimate the parameters of the generative model, and the posterior distributions of the unobservable variables from the observable data.

More specifically, we first introduce the underlying model via a distribution $p^{z, \sigma| \theta} (z, \sigma| \theta)$, which is the conditional probability distribution of $z$ and $\sigma$ when $\theta$ is given, and $p_\mathrm{pr}^\theta (\theta)$, which is the prior distribution of $\theta$.
Here, $\theta$ is the set of parameters that characterize the conditional distribution and $\Sigma \coloneqq \{ \sigma_i \}_{i=1}^N$ is the set of unobservable variables.
For the above modeling to be successful, $p^{z, \sigma| \theta} (z, \sigma| \theta)$ and $p_\mathrm{gen}^z (z)$ have to satisfy $p_\mathrm{gen}^z (\cdot) \approx \sum_{\sigma \in S^\sigma} p^{z, \sigma| \theta} (\cdot, \sigma| \theta_*)$, where $\theta_*$ is an optimal parameter and $S^\sigma$ is the domain of $\sigma$.
For later convenience, we also define $p^{Z, \Sigma| \theta} (Z, \Sigma| \theta) \coloneqq \prod_{i=1}^N p^{z, \sigma| \theta} (z_i, \sigma_i| \theta)$.

Then, VB is an algorithm to compute the posterior distribution of $\theta$ and the hidden variables $\Sigma$ in the above setup.
In particular, the posterior distribution $p^{\Sigma, \theta| Z} (\Sigma, \theta | Z^\mathrm{obs}) \coloneqq \frac{p^{Z, \Sigma| \theta} (Z^\mathrm{obs}, \Sigma| \theta) p_\mathrm{pr}^\theta (\theta)}{p^Z (Z^\mathrm{obs})}$ is computationally intractable, as it is difficult to compute $p^Z (Z^\mathrm{obs})$.
Note that $p^Z (Z^\mathrm{obs}) \coloneqq \sum_{\Sigma \in S^\Sigma} \int_{\theta \in S^\theta} d\theta \, p^{Z, \Sigma| \theta} (Z^\mathrm{obs}, \Sigma| \theta) p_\mathrm{pr}^\theta (\theta)$.
Then, in VB, we try to approximate $p^{\Sigma, \theta| Z} (\Sigma, \theta | Z^\mathrm{obs})$ by introducing a variational function $q^{\Sigma, \theta} (\Sigma, \theta)$ and minimizing the Kullback-Leibler (KL) divergence between $q^{\Sigma, \theta} (\Sigma, \theta)$ and $p^{\Sigma, \theta| Z} (\Sigma, \theta | Z^\mathrm{obs})$.
Specifically, we solve
\begin{align}
  q_*^{\Sigma, \theta} (\Sigma, \theta) &= \argmin_{q^{\Sigma, \theta} (\Sigma, \theta)} \mathrm{KL} \Big( q^{\Sigma, \theta} (\Sigma, \theta) \Big\| p^{\Sigma, \theta| Z} (\Sigma, \theta | Z^\mathrm{obs}) \Big), \label{main_optimization_problem_VB_001_001}
\end{align}
where  the KL divergence between $p (x)$ and $q (x)$, defined over their domain $S^x$,  is given by
\begin{align}
  \mathrm{KL} \Big( p (x) \Big\| q (x) \Big) &\coloneqq \sum_{x \in S^x} p (x) [\ln p (x) - \ln q (x)]. \label{main_def_KL_001_001}
\end{align}
 In Eq.~\eqref{main_def_KL_001_001}, $x$ is assumed to be discrete, but almost the same definition is applicable for a continuous variable by replacing the summation by an integral. Furthermore, after making the assumption of MF where $q^{\Sigma, \theta} (\Sigma, \theta)= q^{\Sigma} (\Sigma)q^{\theta} (\theta)$,  the optimization problem in the right-hand side of Eq.~\eqref{main_optimization_problem_VB_001_001} is solved iteratively by setting
\begin{align}
q_{t+1}^{\Sigma} (\Sigma) &\propto \exp \bigg( \int_{\theta \in S^\theta} d\theta \, q_{t+1}^\theta(\theta) \ln \Big( p^{Z, \Sigma| \theta} (Z^\mathrm{obs}, \Sigma| \theta) p_\mathrm{pr}^\theta (\theta) \Big) \bigg), \\
q_{t+1}^{\theta} (\theta) &\propto \exp \Bigg( \sum_{\Sigma \in S^{\Sigma}} q_t^{\Sigma}(\Sigma) \ln \Big( p^{Z, \Sigma| \theta} (Z^\mathrm{obs}, \Sigma| \theta) p_\mathrm{pr}^\theta (\theta) \Big) \Bigg),
\end{align}
where $q_t^\Sigma (\Sigma)$ and $q_t^\theta(\theta)$ are the distributions of $\Sigma$ and $\theta$ at the $t$-th iteration, respectively~\cite{Bishop_001, Murphy_001}.
Once we get the posterior distributions of $\theta$ and $\Sigma$, we can utilize them for inference problems.

\section{Motivations of quantization and a noncommutative term}
The optimization problem in Eq.~\eqref{main_optimization_problem_VB_001_001}, however, is still highly non-convex with multiple local minima and finding good solutions is a challenging task.
We explain this difficulty of VB from the viewpoint of quantum statistical mechanics, and then show how this seeming escalation of complexity introduced by viewing a Bayesian problem as a quantum system leads to a better solution to the original VB problem.
In statistical physics, the probability $w_n (\beta)$~\footnote{See Sec.~\ref{supp_canonical_distribution_001} in the supplemental information (SI) for the details of the canonical distribution} of finding a system in a configuration with energy $\varepsilon_n$ is given by $w_n (\beta) \coloneqq e^{- \beta \varepsilon_n} / \mathcal{Z} (\beta)$, where $\beta \coloneqq (k_\mathrm{B} T)^{-1}$, $k_\mathrm{B}$ is the Boltzmann constant, $T$ is the temperature of a heat bath to which the system is attached, and $\mathcal{Z} (\beta) \coloneqq \sum_{n = 0}^\infty e^{- \beta \varepsilon_n}$.
We can now reverse directions, and given the VB problem we can construct a virtual physical system such that $p^{Z, \Sigma| \theta} (Z^\mathrm{obs}, \Sigma| \theta)$ is the probability of it being in configuration $\{ Z^\mathrm{obs}, \Sigma \}$ conditioned by $\theta$.
Then, this system is defined by energy levels $\varepsilon_{\Sigma| \theta} = -\frac{1}{\beta} \ln p^{Z, \Sigma| \theta} (Z^\mathrm{obs}, \Sigma| \theta)$.
Since the next step is to construct a virtual quantum system, it is more convenient to use the concept of a Hamiltonian, which specifies the energy level corresponding to every configuration of a system; for our classical system the Hamiltonian is identical to the energy levels.
We first define two Hamiltonians corresponding to the probabilities, $p^{Z, \Sigma| \theta} (Z^\mathrm{obs}, \Sigma| \theta)$ and $p_\mathrm{pr}^\theta (\theta)$:
\begin{align}
H_\mathrm{cl}^{\Sigma| \theta} &\coloneqq - \ln p^{Z, \Sigma| \theta} (Z^\mathrm{obs}, \Sigma| \theta), \label{main_classical_Hamiltonian_001_001} \\
H_\mathrm{pr}^\theta &\coloneqq - \ln p_\mathrm{pr}^\theta (\theta). \label{main_classical_Hamiltonian_001_002}
\end{align}

Next we convert this classical physical systems to quantum ones using the canonical quantization approach~\cite{Sakurai_001}. We denote the projection operator on $\Sigma$ and $\theta$ by $\hat{P}^{\Sigma, \theta} \coloneqq | \Sigma, \theta \rangle \langle \Sigma, \theta |$; then we can write the Hamiltonian operators of Eqs.~\eqref{main_classical_Hamiltonian_001_001} and \eqref{main_classical_Hamiltonian_001_002} as
\begin{align}
\hat{H}_\mathrm{cl}^{\Sigma| \theta} &\coloneqq \sum_{\Sigma \in S^\Sigma} \int_{\theta \in S^\theta} d \theta \, H_\mathrm{cl}^{\Sigma| \theta} \hat{P}^{\Sigma, \theta}, \label{main_classical_Hamiltonian_002_001} \\
\hat{H}_\mathrm{pr}^\theta &\coloneqq \sum_{\Sigma \in S^\Sigma} \int_{\theta \in S^\theta} d \theta \, H_\mathrm{pr}^\theta \hat{P}^{\Sigma, \theta}, \label{main_classical_Hamiltonian_002_002}
\end{align}
where $S^\Sigma$ and $S^\theta$ are the domains of $\Sigma$ and $\theta$, respectively.
Note that the dimension of the Hamiltonian is $K^N$ if $\theta$ is not quantized, and infinity if $\theta$ is quantized.
These Hamiltonians are still diagonal and thus the system is still classical. Each, diagonal element is by definition, $\varepsilon_{\Sigma| \theta} = -\ln p^{Z, \Sigma| \theta} (Z^\mathrm{obs}, \Sigma| \theta)$. Since we are soon going to develop the framework for estimating these probabilities by defining a non-diagonal Hamiltonian, it is useful to introduce the notation of the Gibbs operator
\begin{align}
\hat{f} (\beta_\mathrm{pr}, \beta) &\coloneqq \exp \left( - \beta_\mathrm{pr} \hat{H}_\mathrm{pr}^\theta - \beta \hat{H}_\mathrm{cl}^{\Sigma| \theta}  \right),
\end{align}
and rewrite the the probabilities back in terms of the Hamiltonian notation. For simplicity, we consider the case of a non-informative prior distribution, so that $\hat{H}_\mathrm{pr}^\theta$ is not a function of $\theta$.
Since we are still dealing with a diagonal Hamiltonian, we can rewrite Eq.~\eqref{main_optimization_problem_VB_001_001} in the Hamiltonian formulation:
\begin{align}
\hat{\rho}_*^{\Sigma, \theta} (\Sigma, \theta) &= \argmin_{\hat{\rho}^{\Sigma, \theta}} \mathcal{S} \bigg( \hat{\rho}^{\Sigma, \theta} \bigg\| \frac{e^{- \beta \hat{H}_\mathrm{cl}^{\Sigma| \theta}}}{\mathcal{Z} (\beta)} \bigg) \bigg|_{\beta = 1}, \label{main_optimization_problem_VB_002_001}
\end{align}
where $\mathcal{Z} (\beta)$ is the partition function at $\beta$: $\mathcal{Z} (\beta) \coloneqq \mathrm{Tr} \Big[ e^{- \beta \hat{H}_\mathrm{cl}^{\Sigma| \theta}} \Big]$ and $\beta$ is the inverse temperature.
Furthermore, $\mathcal{S} (\cdot \| \cdot)$ is the quantum relative entropy, which is a quantum extension of the KL divergence, Eq.~\eqref{main_def_KL_001_001}, given by
\begin{align}
\mathcal{S} ( \hat{\rho} \| \hat{\sigma} ) &\coloneqq \mathrm{Tr} [ \hat{\rho} \ln \hat{\rho} - \hat{\rho} \ln \hat{\sigma} ]. \label{main_def_quantum_KL_001_001}
\end{align}
The optimization problem in Eq.~\eqref{main_optimization_problem_VB_002_001} is as difficult as Eq.~\eqref{main_optimization_problem_VB_001_001}, since they are equivalent.

Let us first consider a simpler problem by taking the limit $\beta = \infty$ in Eq.~\eqref{main_optimization_problem_VB_002_001}; then we have
a\begin{align}
| 0; \mathrm{cl} \rangle \langle 0; \mathrm{cl} | &= \argmin_{\hat{\rho}^{\Sigma, \theta}} \lim_{\beta \to \infty} \mathcal{S} \bigg( \hat{\rho}^{\Sigma, \theta} \bigg\| \frac{e^{- \beta \hat{H}_\mathrm{cl}^{\Sigma| \theta}}}{\mathcal{Z} (\beta)} \bigg), \label{main_optimization_problem_VB_003_001}
\end{align}
where $| 0; \mathrm{cl} \rangle$ is the ground state of $\hat{H}_\mathrm{cl}^{\Sigma| \theta}$.
Equation~\eqref{main_optimization_problem_VB_003_001} implies that, at $\beta = \infty$, Eq.~\eqref{main_optimization_problem_VB_001_001} becomes the problem of finding the ground state of the data-defined  Hamiltonian in Eq.~\ref{main_classical_Hamiltonian_002_001}.
As explained next, we can use a variant of the quantum annealing technique to approximate such a ground state, denoted as $|0; \mathrm{cl} \rangle$.

\begin{figure*}[htbp]
\centering
\includegraphics[scale=0.70]{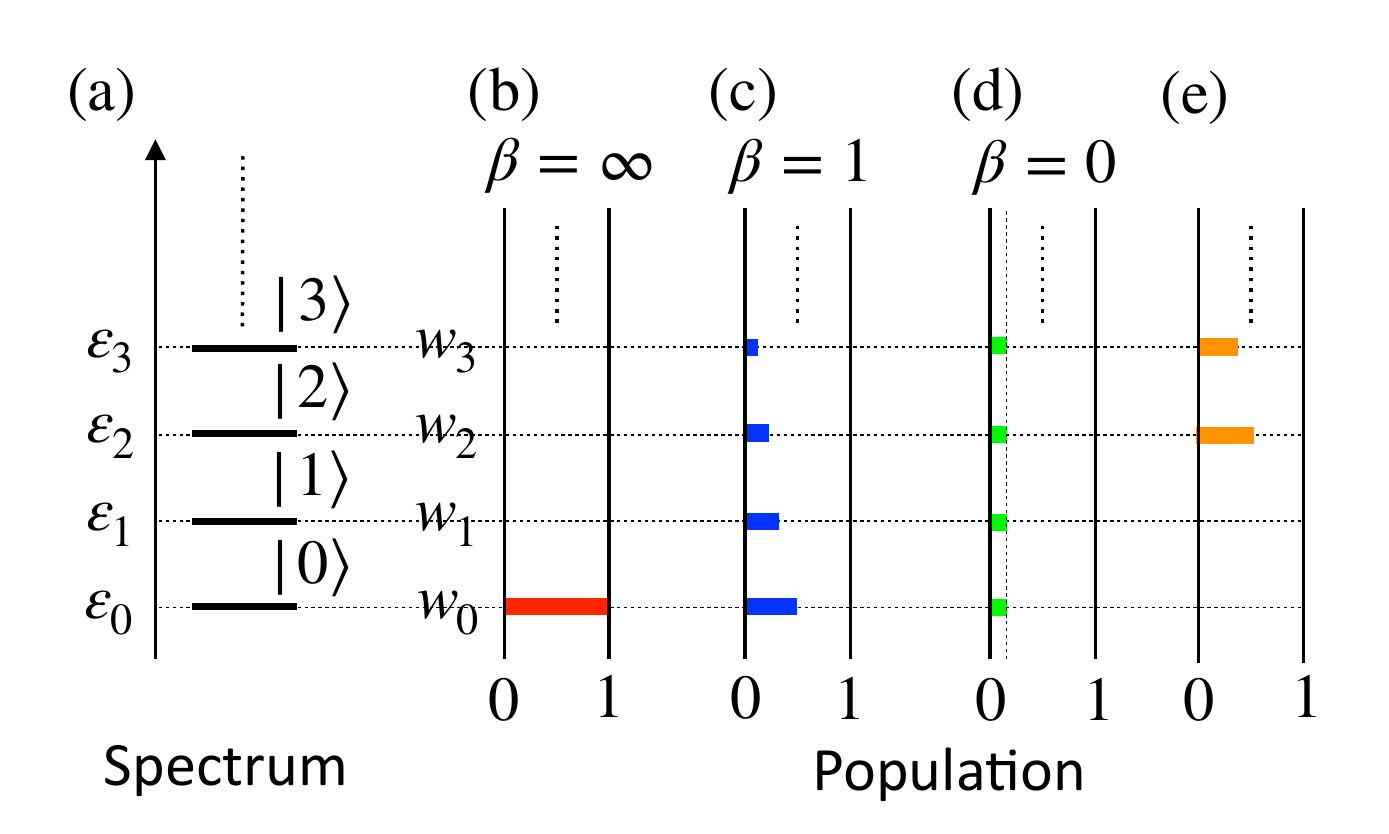}
\includegraphics[scale=0.70]{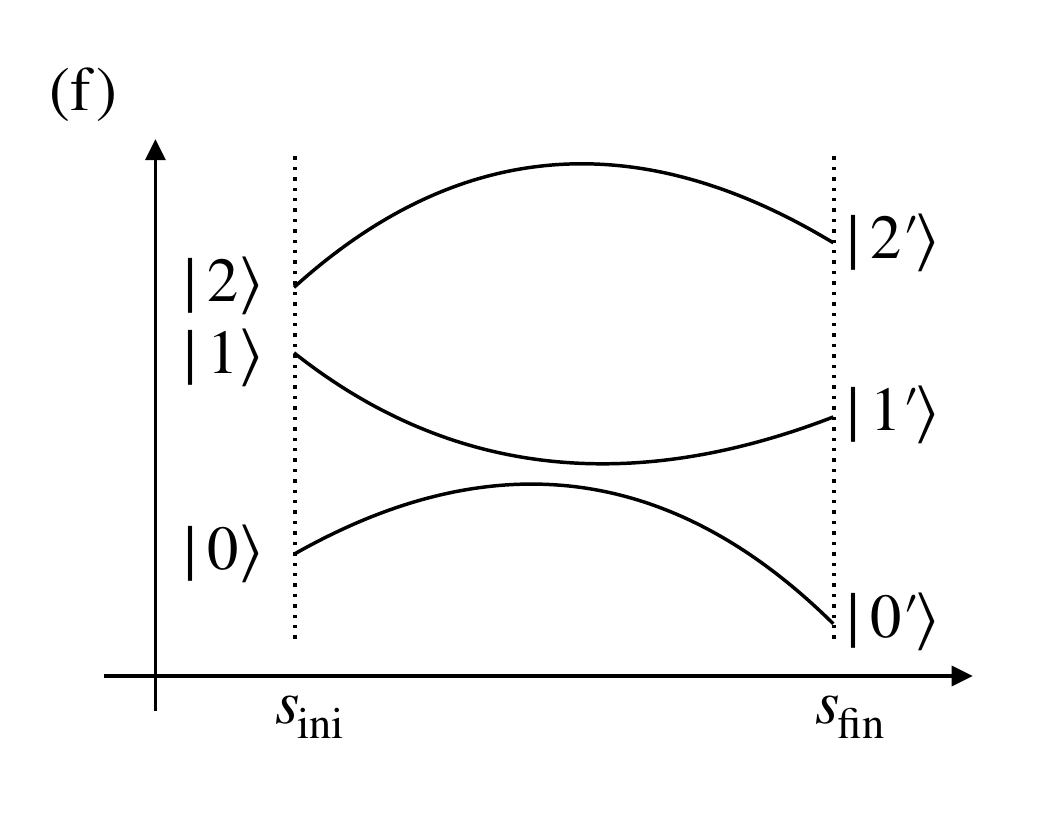}
\caption{\textbf{Quantum advantage in VB explained using schematics.} (a) a typical energy spectrum of $\hat{H} = \sum_{n=0}^\infty \varepsilon_n | n \rangle \langle n |$, (b, c, d) populations of the canonical distributions at $\beta = \frac{1}{T}= \infty, 1, 0$, respectively, (where $T$ is the temperature of the bath attached to the system), and (e) that of a typical non-canonical distribution. We denote the energy level of $|n\rangle$ by $\varepsilon_n$ for $n = 0, 1, 2, \dots$ and assume that $\varepsilon_0 \le \varepsilon_1 \le \dots \le \varepsilon_n$. A mixed state is written as $\hat{\rho} = \sum_{n=0}^\infty w_n | n \rangle \langle n |$ where $w_n \ge 0$ for $n = 0, 1, 2, \dots$ and $\sum_{n=0}^\infty w_n = 1$. At $\beta = \infty$ we have $w_0 = 1$ and $w_n = 0$ for $n = 1, 2, \dots$  while we have $w_n = \mathrm{const.}$ for $n = 0, 1, 2, \dots$ at $\beta = 0$. (f) Schematic of the change of the energy spectrum of $\hat{H} (s)$ from $s_\mathrm{ini}$ to $s_\mathrm{fin}$. By construction, \textit{the optimal solution to the VB problem corresponds to the canonical distribution of the corresponding physical system at $\beta=1$}. If one could start with the system in a canonical distribution at zero temperature ($\beta >>1$), which is the ground state,  then one could raise temperature slowly to reach the canonical distribution at $\beta=1$ and hence, obtain the optimum solution to the VB problem. \\ \textit{QAVB uses a variant of quantum annealing to approximate the ground state at close to $\beta = \infty$, and then increases the temperature to $\beta =1$ leading to a closer approximation to the canonical distribution}. Moreover, it requires only a single run (especially with $s_0 =1$ as in Algorithm~\ref{main_algo_QAVB_001_001}) without any dependence on initialization. In contrast, other methods based on classical statistics or Monte Carlo methods, the challenge is to start with a canonical distribution at any $\beta_0$ (as opposed to QAVB that allows one to approximate the canonical distribution at $T=0$ or $\beta_0=\infty$) and avoid having to raise the temperature. For example, the deterministic annealing method, either starts with a  very high initial temperature (where the canonical distribution is trivially known, i.e., uniform) and  gets stuck at a saddle point, or starts with a random initialization of the distribution at a finite temperature (which would not be the canonical distribution for that temperature)   getting easily stuck in local minima and leading to different estimations sensitive to the initial choice. }
\label{main_shcematic_001_001}
\end{figure*}

We next consider the relationship between the populations of the canonical distributions at $\beta = \infty$, and at $\beta = 0$ and discuss how such canonical distributions might evolve as $\beta$ is changed adiabatically to $\beta=1$. Note that a canonical distribution at $\beta=1$ corresponds to an optimal solution to the VB problem.
Figs.~\ref{main_shcematic_001_001}(b) and (c), show  schematic representations of populations of the canonical distributions at $\beta = \infty,$ and at $\beta= 1$, respectively.
As shown in Figs.~\ref{main_shcematic_001_001}(b) and (c), once we get $|0; \mathrm{cl} \rangle$, we obtain $\hat{\rho}_*^{\Sigma, \theta} (\Sigma, \theta)$ in Eq.~\eqref{main_optimization_problem_VB_002_001} by pumping up the population of the ground state to those of excited states deterministically. If one starts at a very high temperature (i.e., $\beta \approx 0$), as often done in DAVB, then the initial canonical distribution is uniform, and it is well known that when one reduces temperature then it gets stuck in a saddle point, far from the canonical distribution. On the other hand, if one starts at finite temperature, then one has to assume a non-canonical distribution as  initial condition, and then the algorithm gets easily stuck in a local minimum.
In other words, it is difficult to obtain the canonical distribution at $\beta = 1$ from that at $\beta = 0$ or a non-canonical distribution, as shown in Fig.~\ref{main_shcematic_001_001}(e).
Thus, if the ground state is available, it helps us to solve the VB problem, Eq.~\eqref{main_optimization_problem_VB_001_001}.

\section{Quantum annealing variational Bayes (QAVB) inference}
We describe QAVB by following Ref.~\cite{Miyahara_002}. In general,
QA~\cite{Finnila_001, Kadowaki_001, Farhi_001} is a method to find the ground state of a given Hamiltonian by using the adiabatic theorem, as shown in Fig.~\ref{main_shcematic_001_001}(f).
If we can design a parametrized Hamiltonian $\hat{H} (s)$ such that $\hat{H} (s_\mathrm{ini})$ is solvable and $\hat{H} (s_\mathrm{fin}) = \hat{H}_\mathrm{pr}^\theta + \hat{H}_\mathrm{cl}^{\Sigma| \theta}$, then one can apply QA to approximate the desired ground state. In the case of QA the dynamics is described by the Schr\"odinger equation; then the adiabatic theorem holds for the dynamics of a time-dependent system.
However, in our case, the state evolution follows the MF equation and a similar property is not known. The analysis adiabatic evolution in QAVB is one of the goals of  this paper and addressed in a later section.

In the rest of this paper, we formulate QAVB by adding a noncommutative term to the Hamiltonians of VB, Eqs.~\eqref{main_classical_Hamiltonian_002_001} and \eqref{main_classical_Hamiltonian_002_002}, and confirm its validity. By using Eqs.~\eqref{main_classical_Hamiltonian_002_001} and \eqref{main_classical_Hamiltonian_002_002}, we then define the following Gibbs operator:
\begin{align}
\hat{f} (\beta, s) &\coloneqq \exp \left( - \hat{H}_\mathrm{pr}^\theta - \beta (1 - s) \hat{H}_\mathrm{cl}^{\Sigma| \theta} - \beta s \hat{H}_\mathrm{qu}^\Sigma \right). \label{main_Gibbs_operator_quantum_001_001}
\end{align}
Here the third term of Eq.~\eqref{main_Gibbs_operator_quantum_001_001} is given by $\hat{H}_\mathrm{qu}^{\Sigma} \coloneqq \sum_{i=1}^N \hat{H}_\mathrm{qu}^{\sigma_i}$,
and each term in the right-hand side satisfies the following noncommutative relation:
\begin{align}
\bigg[ \hat{H}_\mathrm{qu}^{\sigma_i}, \bigg( \overset{i-1}{\underset{j = 1}{\otimes}} \hat{I}^{\sigma_j} \bigg) \otimes \hat{\sigma}_i \otimes \bigg( \overset{N}{\underset{j = i+1}{\otimes}} \hat{I}^{\sigma_j} \bigg) \otimes \hat{I}^\theta \bigg] \ne 0. \label{main_noncommutative_relation_001_001}
\end{align}
Here, $\hat{\sigma}_i$ is a matrix such that $\hat{\sigma}_i | \sigma_i \rangle = \sigma_i | \sigma_i \rangle$ and $\hat{I}^{(\cdot)}$ is the identity operator for the corresponding Hilbert space.
Using Eqs.~\eqref{main_def_quantum_KL_001_001} and \eqref{main_Gibbs_operator_quantum_001_001}, we consider the following quantum relative entropy:
\begin{align}
\mathcal{S} \bigg( \hat{\rho}^{\Sigma, \theta} \bigg \| \frac{\hat{f} (\beta, s)}{\mathcal{Z} (\beta, s)} \bigg) &\coloneqq \mathrm{Tr}_{\Sigma, \theta} \bigg[ \hat{\rho}^{\Sigma, \theta} \bigg\{\ln \hat{\rho}^{\Sigma, \theta} - \ln \frac{\hat{f} (\beta, s)}{\mathcal{Z} (\beta, s)} \bigg\} \bigg], \label{main_quantum_relative_entropy_001_001}
\end{align}
where $\mathcal{Z} (\beta, s)$ is the partition function given by $\mathcal{Z} (\beta, s) \coloneqq \mathrm{Tr}_{\Sigma, \theta} \Big[ \hat{f} (\beta, s) \Big]$.
By minimizing Eq.~\eqref{main_quantum_relative_entropy_001_001} with respect to $\hat{\rho}^{\Sigma, \theta}$, we can estimate the distribution of $\theta$.
However, the minimization problem of Eq.~\eqref{main_quantum_relative_entropy_001_001} is quite difficult; then we utilize the following decomposition:
\begin{align}
  \hat{\rho}^{\Sigma, \theta} \approx \hat{\rho}^{\Sigma} \otimes \hat{\rho}^{\theta}. \label{main_MF_approximation_001_001}
\end{align}
Equation~\eqref{main_MF_approximation_001_001} is often called the MF approximation.
By performing the variational calculation of Eq.~\eqref{main_quantum_relative_entropy_001_001} with Eq.~\eqref{main_MF_approximation_001_001}, we obtain the following update equations:
\begin{align}
\hat{\rho}_{t+1}^\Sigma &\propto \exp \left( \mathrm{Tr}_{\theta} \left[ \Big(\hat{I}^\Sigma \otimes \hat{\rho}_{t+1}^\theta \Big) \ln \hat{f} (\beta_t, s_t) \right] \right), \label{main_QAVB_E-step_001_001} \\
\hat{\rho}_{t+1}^\theta &\propto \exp \left( \mathrm{Tr}_{\Sigma} \left[ \Big( \hat{\rho}_t^\Sigma \otimes \hat{I}^\theta \Big) \ln \hat{f} (\beta_t, s_t) \right] \right). \label{main_QAVB_M-step_001_001}
\end{align}
Finally, we summarize this algorithm in Algo.~\ref{main_algo_QAVB_001_001}. Note that the setting of $s_0 =1$ in the algorithm ensures that there is no dependence of the results on the initial choice of $\hat{\rho}_{0}^\Sigma$, and hence \textit{this variant of the QAVB is executed only once for a given problem}. In contrast, for DAVB and VB (also for QAVB where $s_0 <1$; see Ref.~\cite{Miyahara_002}) results are highly sensitive to the initial conditions and good outcomes are obtained with low probability.
\begin{algorithm}[t]
  \DontPrintSemicolon
  \caption{Quantum annealing variational Bayes (QAVB) inference with $s_0 = 1$.}
  \label{main_algo_QAVB_001_001}
  set $t \leftarrow 0$ and $p_\mathrm{pr}^\theta (\theta)$ \;
  fix annealing schedules $\{ s_t \}$ and $\{ \beta_t \}$ such that $s_0=1$ and $\beta_0 \gg 1.0$ \;
  \While{convergence criterion is not satisfied}{
  compute $\hat{\rho}_{t+1}^\theta$ in Eq.~\eqref{main_QAVB_M-step_001_001} \;
  compute $\hat{\rho}_{t+1}^\Sigma$ in Eq.~\eqref{main_QAVB_E-step_001_001} \;
  update $t \leftarrow t+1$ \;
  }
\end{algorithm}

\begin{figure}[t]
\centering
\includegraphics[scale=0.50]{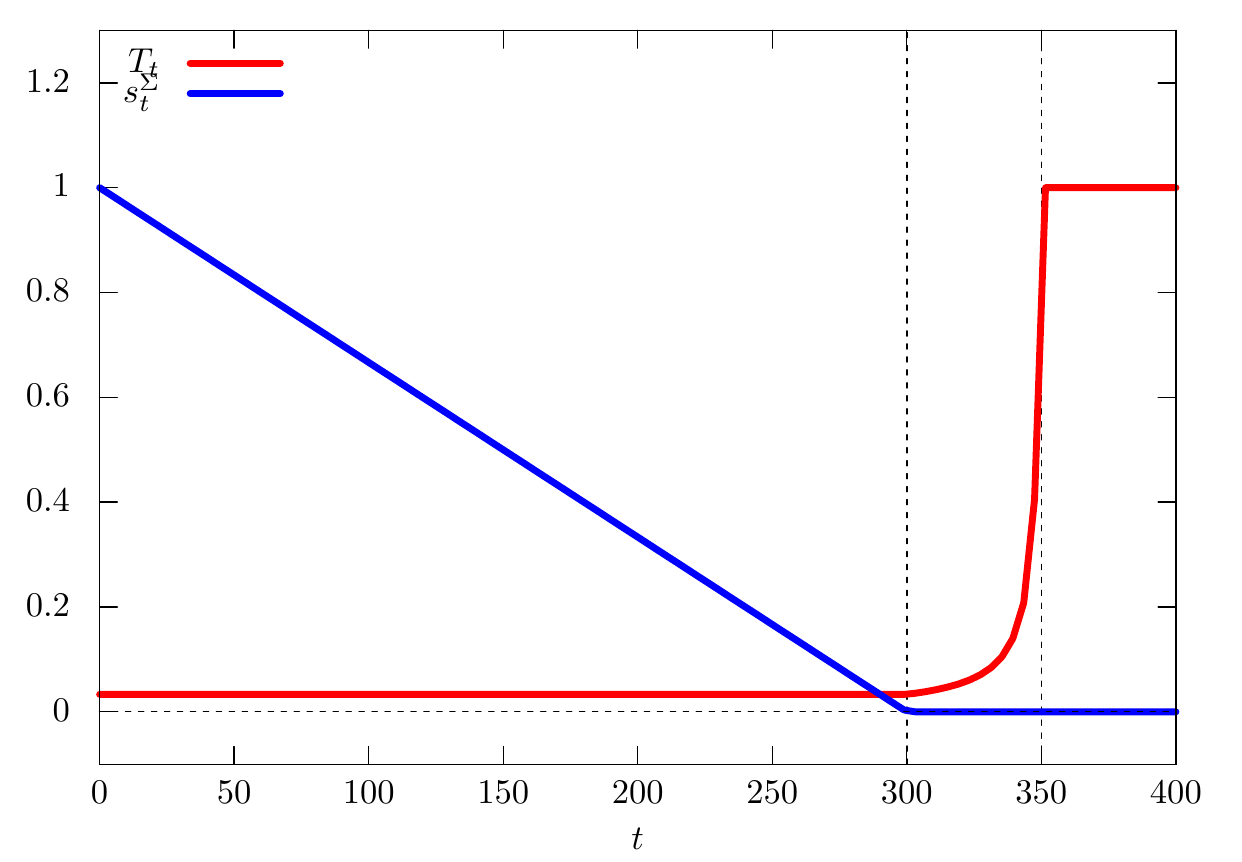}
\includegraphics[scale=0.50]{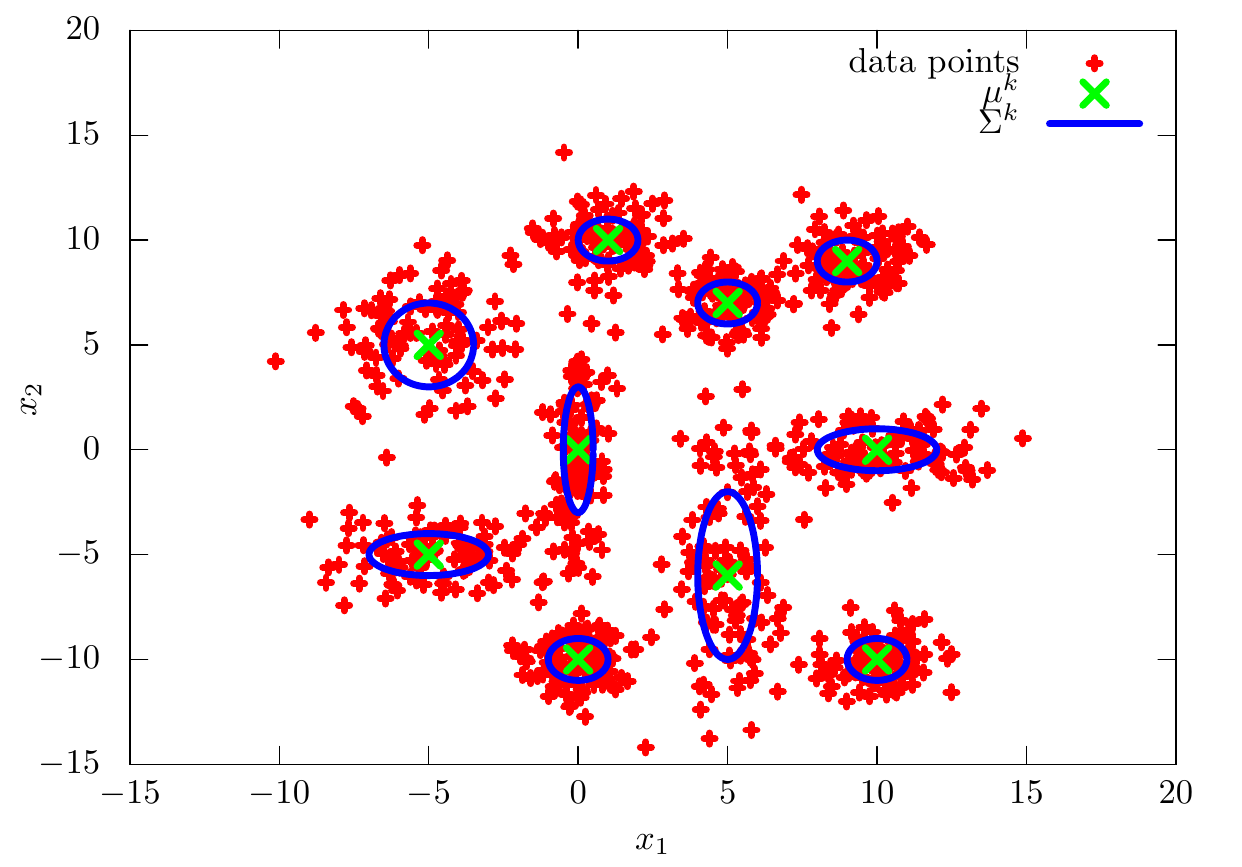}
\includegraphics[scale=0.50]{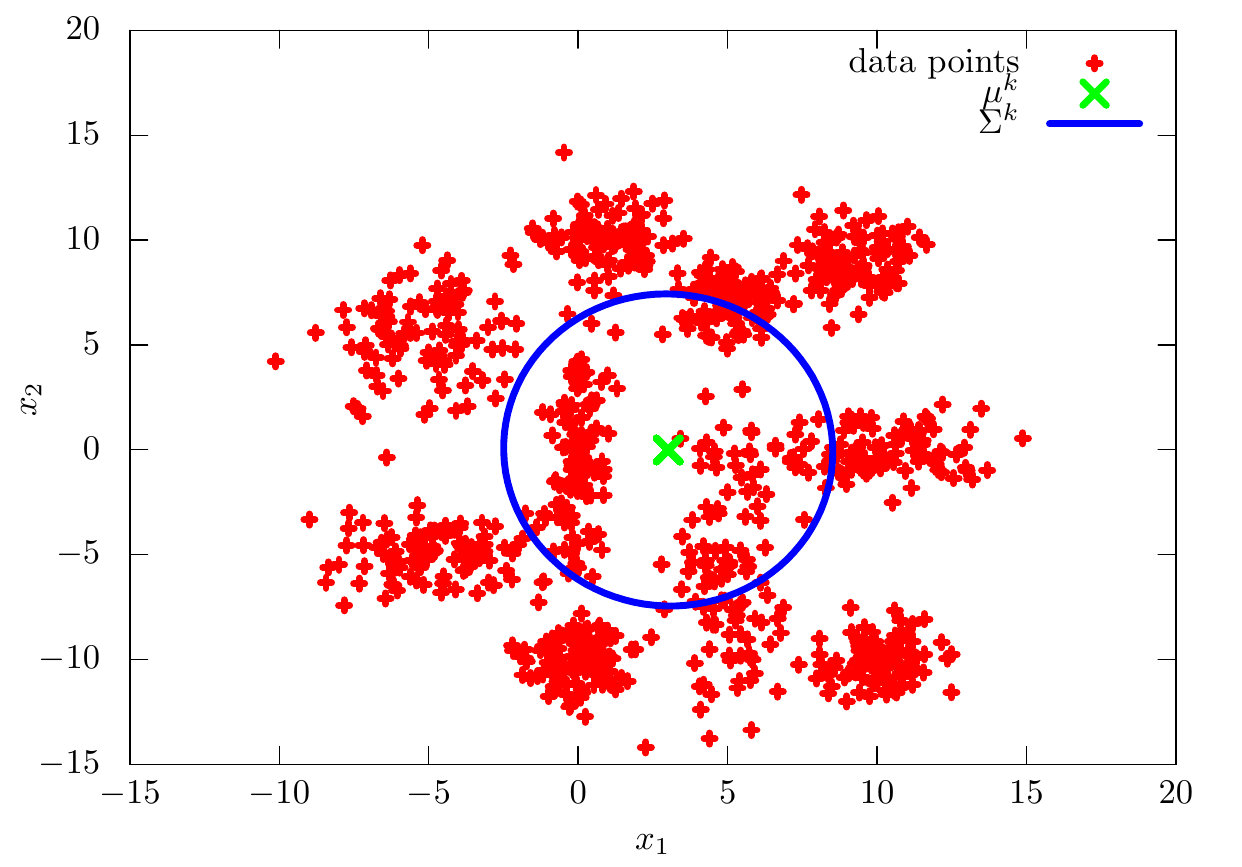}
\caption{(a) Annealing schedules described by Eqs.~\eqref{main_annealing_schedule_QAVB_001_001} and \eqref{main_annealing_schedule_QAVB_001_002} with $\beta_0 = 30.0$, $s_0 = 1.0$, $\tau_1 = 300$, and $\tau_2 = 350$. (b) Two-dimensional data set generated by ten Gaussian functions. Each data point has a label. (c) Gaussian functions at step 112 estimated by DAVB with $\beta_0 = 0.0010$, $s_0 = 0.0$, $\tau_1 = 10$, and $\tau_2 = 100.$ Only one Gaussian mode dominates; the rest have $\pi_i \approx  0$. This shows that when DAVB starts at a high temperature and is slowly cooled, it gets stuck in a saddle point.}
\label{main_numerical_001_001}
\end{figure}

There are multiple candidates for $H_\mathrm{qu}^\Sigma$ that satisfies Eq.~\eqref{main_noncommutative_relation_001_001}.
In numerical simulations, we use the following $\hat{H}_\mathrm{qu}^{\sigma_i}$:
\begin{align}
\hat{H}_\mathrm{qu}^{\sigma_i} &\coloneqq \bigg( \overset{i-1}{\underset{j = 1}{\otimes}} \hat{I}^{\sigma_j} \bigg) \otimes \Bigg( \sum_{k = 1}^K (| \sigma_i = k \rangle \langle \sigma_i = k + 1 | \nonumber \\
& \quad + | \sigma_i = k + 1\rangle \langle \sigma_i = k |) \Bigg) \otimes \bigg( \overset{N}{\underset{j = i+1}{\otimes}} \hat{I}^{\sigma_j} \bigg) \otimes \hat{I}^\theta,
\end{align}
where $| \sigma_i = K + 1 \rangle = | \sigma_i = 1 \rangle$.
To run QAVB, we also need to fix an annealing schedule; so, it is quite important to construct an efficient one.
However, there are an infinite number of possible annealing schedules; so we need to limit ourselves.
In Ref.~\cite{Miyahara_002}, the following annealing schedules for $s_t$ and $\beta_t = 1 / T_t$, where $T_t$ is the temperature of the bath to which the system is attached at time $t$, are adopted:
\begin{align}
s_t &= s_0 \times \mathrm{max} (1 - t / \tau_\mathrm{1}, 0.0), \label{main_annealing_schedule_QAVB_001_001} \\
\beta_t &=
\begin{cases}
\beta_0 & (t \le \tau_\mathrm{1}), \\
1 + \frac{ (\beta_0 - 1) (\tau_\mathrm{2} - t)}{\tau_\mathrm{2} - \tau_\mathrm{1}} & (\tau_\mathrm{1} \le t \le \tau_\mathrm{2}), \\
1.0 & (t \ge \tau_\mathrm{2}).
\end{cases} \label{main_annealing_schedule_QAVB_001_002}
\end{align}
Note that Eqs.~\eqref{main_annealing_schedule_QAVB_001_001} and \eqref{main_annealing_schedule_QAVB_001_002} are characterized by four parameters: $s_0$, $\beta_0$, $\tau_1$, and $\tau_2$.
Furthermore, the performance of QAVB on $s_0$ and $\beta_0$ is investigated in Ref.~\cite{Miyahara_002} and it shows that $s_0 = 1.0$ and $\beta_0 = 30.0$ are effective.
In Fig.~\ref{main_numerical_001_001}(a), we plot the annealing schedules described by Eqs.~\eqref{main_annealing_schedule_QAVB_001_001} and \eqref{main_annealing_schedule_QAVB_001_002} with $\beta_0 = 30.0$, $s_0 = 1.0$, $\tau_1 = 300$, and $\tau_2 = 350$.

\section{Mechanisms of QAVB}
To discuss the dynamics of an estimate by QAVB, we focus on the annealing schedule described by Eqs.~\eqref{main_annealing_schedule_QAVB_001_001} and \eqref{main_annealing_schedule_QAVB_001_002} with $\beta_0 \gg 1.0$ and $s_0 = 1.0$ since, as we see later, QAVB with this annealing schedule shows high performance.
The annealing schedule can be divided into two parts.
First quantum fluctuations are gradually decreased until they disappear at low temperature, and then the temperature $\beta$ is raised to 1, at which point the cost functions of QAVB and VB are identical.
We develop a highly likely  mechanism of QAVB on the basis of this decomposition as follows.

Due to the nature of the canonical distribution, the ground state of $\hat{H}_\mathrm{cl}^{\Sigma| \theta}$ dominates the density operator at finite but large $\beta >> 1$.
In the QA part of the annealing schedule, the state is expected to gradually vary from the ground state of the Hamiltonian, $\hat{H}_\mathrm{qu}^\Sigma$, that has a trivial ground state (by design) to that of the Hamiltonian of interest, $\hat{H}_\mathrm{cl}^{\Sigma| \theta}$ in Eq.~\eqref{main_classical_Hamiltonian_001_001}. Of course, given the parameterized form of $\hat{\rho}^{\Sigma, \theta}$ used in VB, and the MF approximation, $\hat{\rho}^{\Sigma, \theta}= \hat{\rho}^{\Sigma} \otimes \hat{\rho}^{\theta}$ one can only approximate the ground state. Picking more expressive functional forms or, as shown in the numerical section, increasing the number of clusters $K$ in the GMM estimation problem can improve the expressive power and lead to better approximation and improved performance.

Furthermore, the ground state corresponds to the hard clustering assignment, in the sense that each datapoint is assigned to exactly one categorical value. This follows from the observation that $\hat{H}_\mathrm{cl}^{\Sigma| \theta}$ is diagonal and hence its ground state corresponds to a diagonal element, where $\Sigma$ is fixed, which implies that each datapoint is assigned to a single hidden categorical value. Such \textit{optimal hard clustering is also an important problem in machine learning, and thus it is useful to   obtain or closely approximate the ground state}.

Then, in the second part of the annealing schedule, we raise the temperature to obtain the state that minimizes the cost function of VB, Eq.~\eqref{main_quantum_relative_entropy_001_001} with $\beta = 1.0$ and $s = 0.0$.
From the viewpoint of physics, saddle points are associated with spontaneous symmetry breaking (SSB).
We often come across SSB in the process of decreasing temperature; on the other hand, all the experiments and theoretical analysis so far have shown that there is no SSB in the process of increasing temperature.
Thus, we can expect that, if we have the ground state at $T \approx 0 \ (\beta = 1 / T \approx \infty)$, then we can have the canonical distribution at any $\beta$ just by decreasing $\beta$.
This discussion is also expected to hold for QAVB.
In this paper, we validate this discussion by looking at the estimates before and after raising temperature (see Table~\ref{main_table_success_rate_001_001}).

DAVB was also developed on the idea that an optimal estimate is continuously connected and a global optimum would be obtained by changing temperature gradually. The update steps are identical to that of QAVB when $s_t = 0$; see Supporting Information for derivations.
However, if we start DAVB with high temperature (i.e. $\beta \approx 0$), we cannot avoid SSB and, if we start it with low temperature, then the final estimate  depends strongly on the initial configuration.
Such deficiencies motivated us to develop QAVB and analyze its dynamics. We show here why QAVB has a different dynamics, allowing it to outperform other methods.

\section{Numerical simulations}
For demonstrating the quantum advantage in VB inference, and showcase the dynamics of QAVB, we apply QAVB to the clustering problem using the Gaussian mixture model (GMM).
In the numerical simulations, two datasets are investigated: two-dimensional and three-dimensional datasets generated by the GMM~\footnote{See Sec.~\ref{supp_sec_GMM_001} for the details of the GMM.
The prior and posterior distributions for the GMM are described in Sec.~\ref{supp_sec_GMM_prior_posterior_001_001}.}. These low dimensional datasets are sufficient to demonstrate the various factors that contribute to the successful dynamics of QAVB. For applications of QAVB to higher dimensional datasets, please see  Ref.~\cite{Miyahara_002}.
In Fig.~\ref{main_numerical_001_001}(b), the first dataset is shown.
To quantify performance, we define the prediction rate, or \textit{success rate}, as the ratio of how many hidden variables are correctly estimated (i.e., how many data points are assigned to the same Gaussian as in the model that generated the data) to the total number of data points.
Note that there is an arbitrariness on the permutation of hidden variables; thus, we use the maximum value with respect to the permutation as the prediction rate.

\begin{figure}[t]
\includegraphics[scale=0.450]{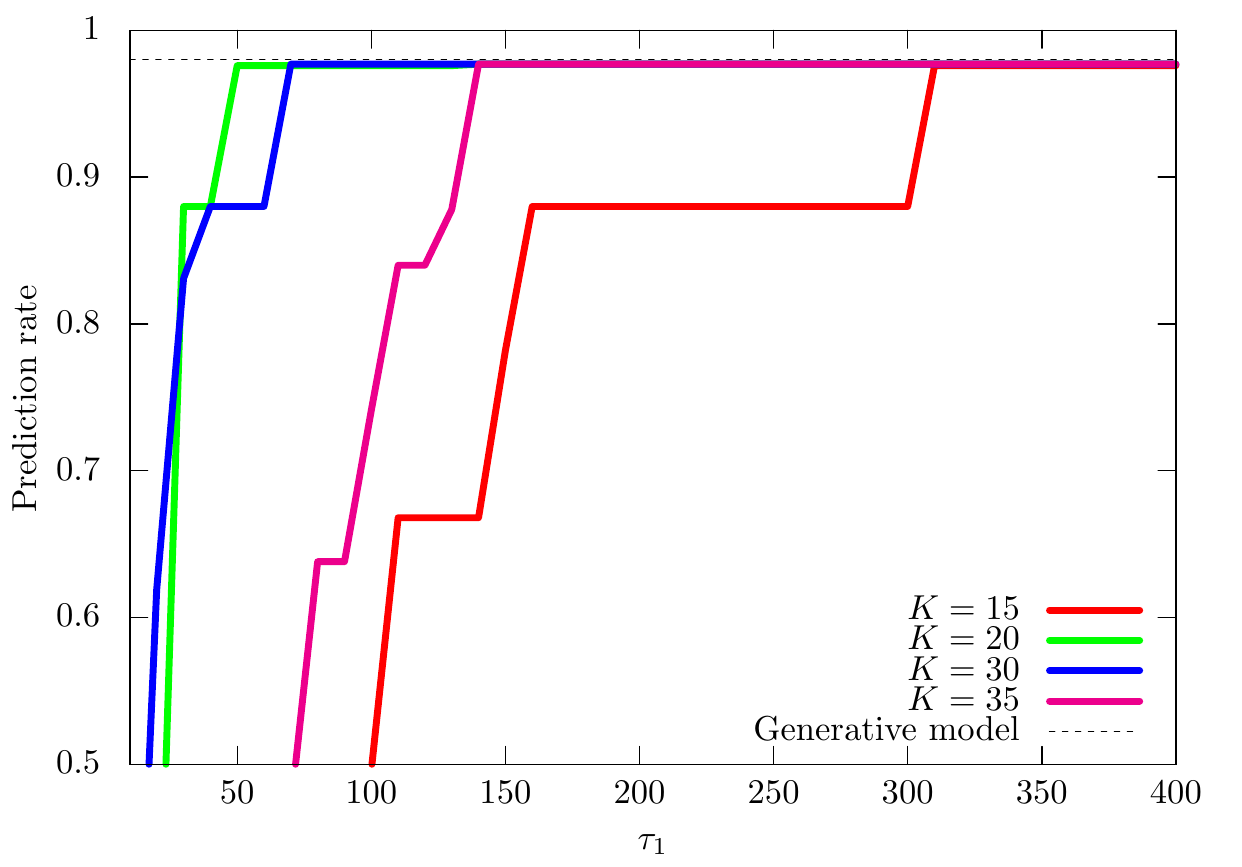}
\includegraphics[scale=0.450]{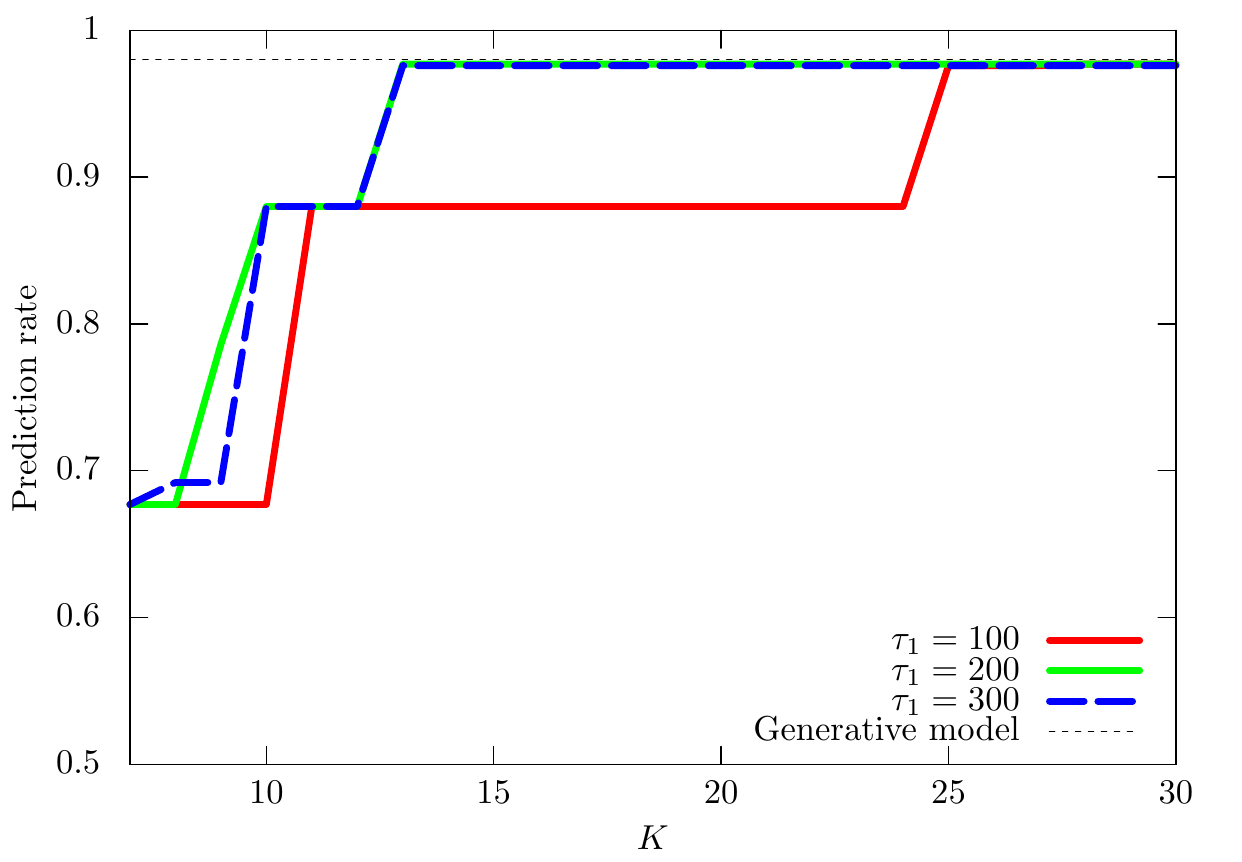}
\includegraphics[scale=0.450]{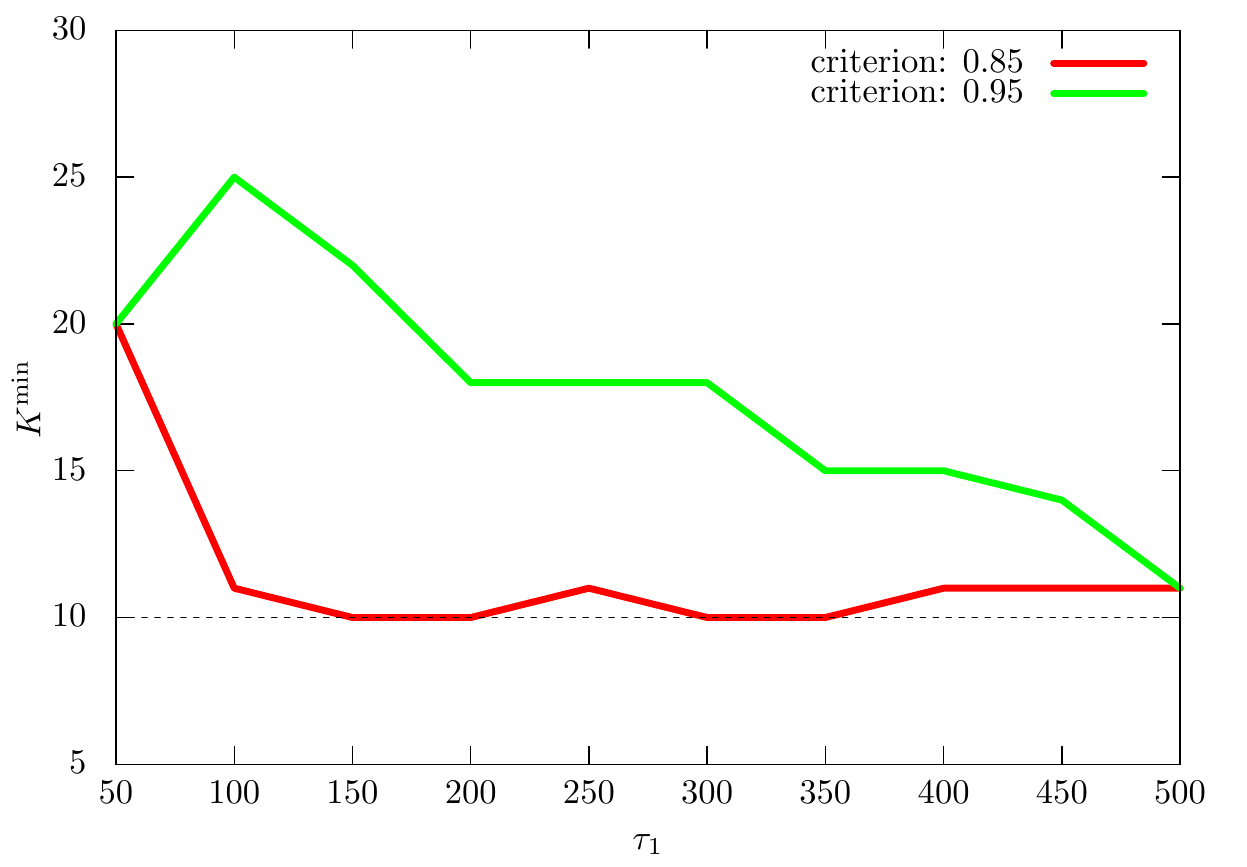}
\includegraphics[scale=0.450]{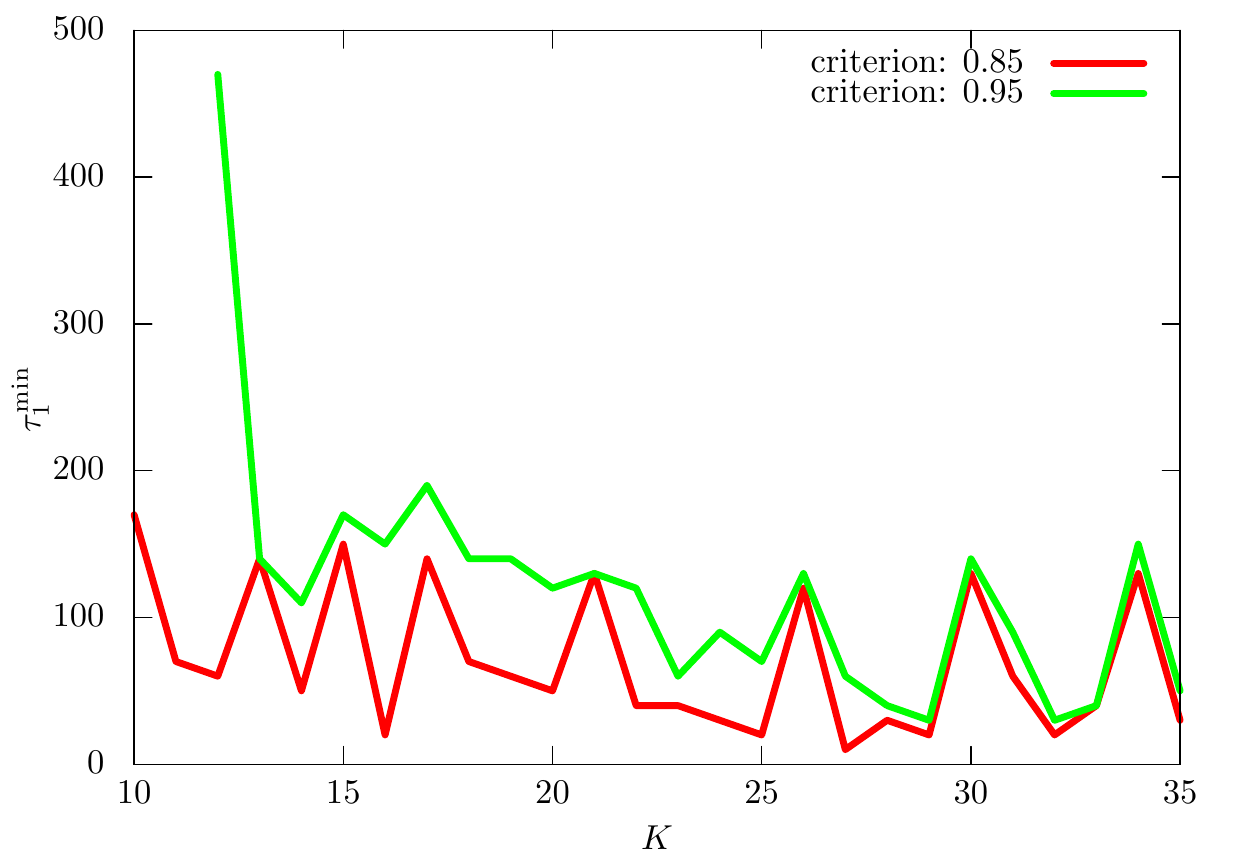}
\caption{\textbf{Tradeoffs between quantum annealing duration $\tau_1$ and $K$, the number of Gaussians in the QAVB algorithm:} In order for QAVB to achieve a state close to the ground state at zero temperature, both $K$  --which determines the expressive power of the MF variational function-- and $\tau_1$ --which determines how slow we anneal-- need to be set. As defined on page \pageref{numerical-analysis},  for a fixed $\tau_1$, $p_\mathrm{suc}^K (K, \tau_1)= \max_{k\leq K} p_\mathrm{suc}(k, \tau_1)$, that is, the maximum accuracy obtained by varying the number of clusters up to $K$ for a fixed $\tau_1$ . Similarly,  $p_\mathrm{suc}^{\tau_1} (K, \tau_1)$ is the maximum accuracy obtained for any $\tau_1^{'}\leq \tau_1$ for any fixed $K$.  Numerical computations for the two-dimensional data illustrated in Fig.~\ref{main_numerical_001_001} are presented here; for this data set the maximum possible success rate as obtained from the generative model  is $0.98$ and is shown  by the black dotted lines. (a) Dependence of $p_\mathrm{suc}^K (K, \tau_1)$ on $\tau_1$ for different $K$'s, (b) that of $p_\mathrm{suc}^{\tau_1} (K, \tau_1)$ on $K$. As these plots show, QAVB can achieve almost optimal performance for a wide range of $K\geq 14$ and $100\leq \tau_1 \leq 300$; two best performing combinations are ($K=20$, $\tau_1=50$) and ($K=14$, $\tau_1=200$). Next we set a target success rate, $p_\mathrm{cr}$, of $0.85$ and $0.95$, respectively. (c) $K^\mathrm{min} (\tau_1)$, which is the minimum value of $K$ to achieve a given $p_\mathrm{cr}$, as $\tau_1$ is varied, and (d) $\tau_1^\mathrm{min} (K)$, which is the minimum value of $\tau_1$ to achieve a given $p_\mathrm{cr}$. For example, for $\tau_1 =500$ one can match the best performance for $K=12$.Note that in (c) and (d) the values are not monotonically decreasing.
Note that when $K$ is set larger than $10$ (actual number of gaussians), the posterior probabilities of only $10$ modes match the corresponding ground truth values, and the rest of the $K-10$ modes have almost zero posterior probabilities.
Similar results are shown for a 3-D dataset in Fig~\ref{main_numerical_003_002}.}
\label{main_numerical_002_001}
\end{figure}

We use Eqs.~\eqref{main_annealing_schedule_QAVB_001_001} and \eqref{main_annealing_schedule_QAVB_001_002} and set $\beta_0 = 30.0$ and $s_0 = 1.0$ for the annealing schedule of the experiments. With these settings, Algorithm~\ref{main_algo_QAVB_001_001} is run only once for any given choices of $K$, $\tau_1$ and $\tau_2$.
\textit{The prediction rate is a function of the hyper-parameters $K$ and $\tau_1$} and we hereafter denote it by $p_\mathrm{suc} (K, \tau_1)$. Clearly, the duration of the QA steps, $\tau_1$, determines how closely one can track the ground state and plays a crucial role. The number of clusters, $K$, determines the number of parameters and hence, the expressive power of $\hat{\rho}^{\Sigma, \theta}$ in approximating the ground state. one of the goals of our experiments is to show the tradeoffs between these two hyper-parameters. Since $p_\mathrm{suc} (K, \tau_1)$ is not monotonic with respect to $K$ and $\tau_1$, we define $p_\mathrm{suc}^K (K, \tau_1) \coloneqq \max_{K' \le K} p_\mathrm{suc} (K', \tau_1)$ and $p_\mathrm{suc}^{\tau_1} (K, \tau_1) \coloneqq \max_{\tau_1' \le \tau_1} p_\mathrm{suc} (K, \tau_1')$.
In Figs.~\ref{main_numerical_002_001}(a) and (b), we plot the dependence of $p_\mathrm{suc}^K (K, \tau_1)$ on $\tau_1$ and the dependence of $p_\mathrm{suc}^{\tau_1} (K, \tau_1')$ on $K$, respectively.
These results show that for sufficiently large $\tau_1$ and $K$, QAVB shows a high prediction rate that almost matches the upper bound set by  the generative model, with full knowledge. \label{numerical-analysis}
Given prediction criterion $p_\mathrm{cr}$, we then define $K^\mathrm{min} (\tau_1) \coloneqq \argmin_K p_\mathrm{suc} (K, \tau_1)$ and $\tau_1^\mathrm{min} (K) \coloneqq \argmin_{\tau_1} p_\mathrm{suc} (K, \tau_1)$ subject to $p_\mathrm{suc} (K, \tau_1) \ge p_\mathrm{cr}$.
In Figs.~\ref{main_numerical_002_001}(c) and (d), we plot $K^\mathrm{min} (\tau_1)$ and $\tau_1^\mathrm{min} (K)$, respectively.
These figures show that, to achieve $p_\mathrm{cr} = 0.85, 0.95$, relatively small $K$ and $\tau_1$ are enough.

\begin{figure}[t]
\includegraphics[scale=0.50]{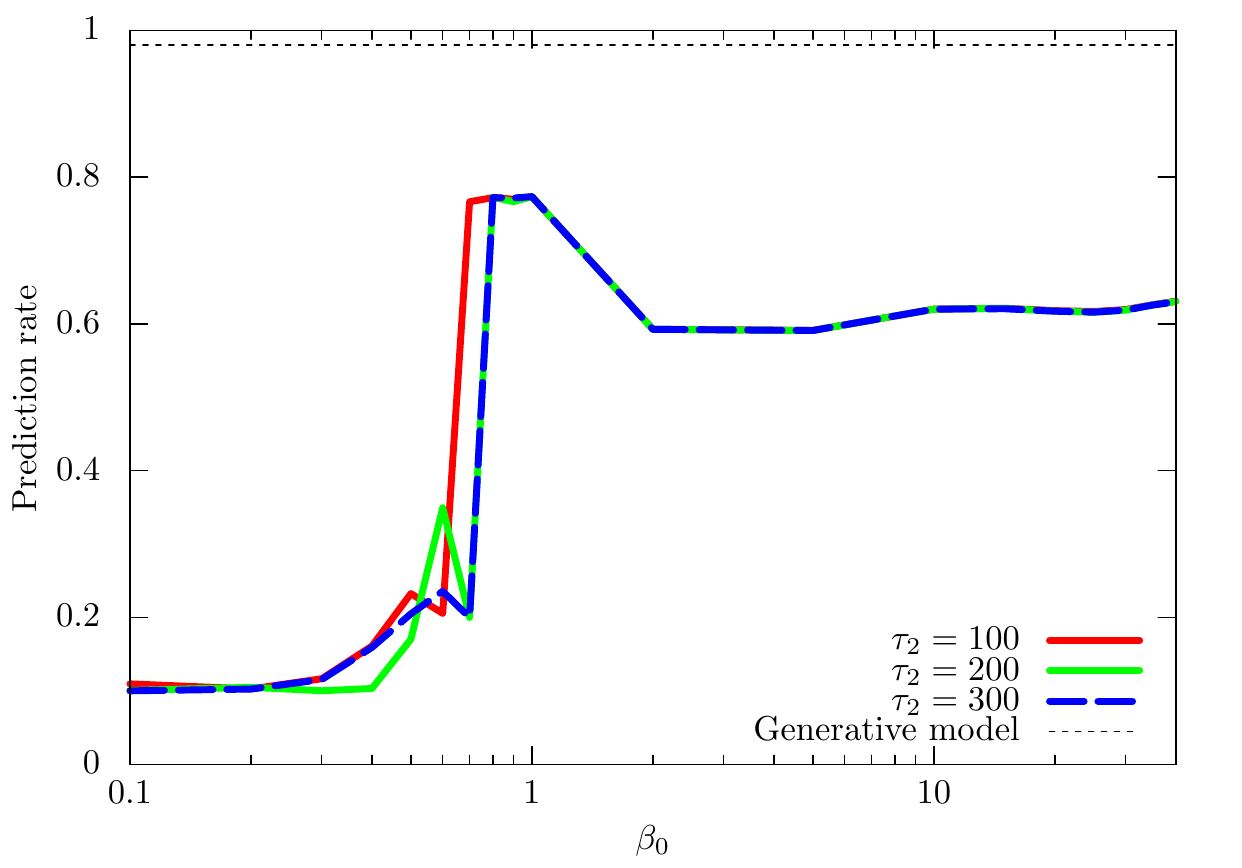}
\includegraphics[scale=0.50]{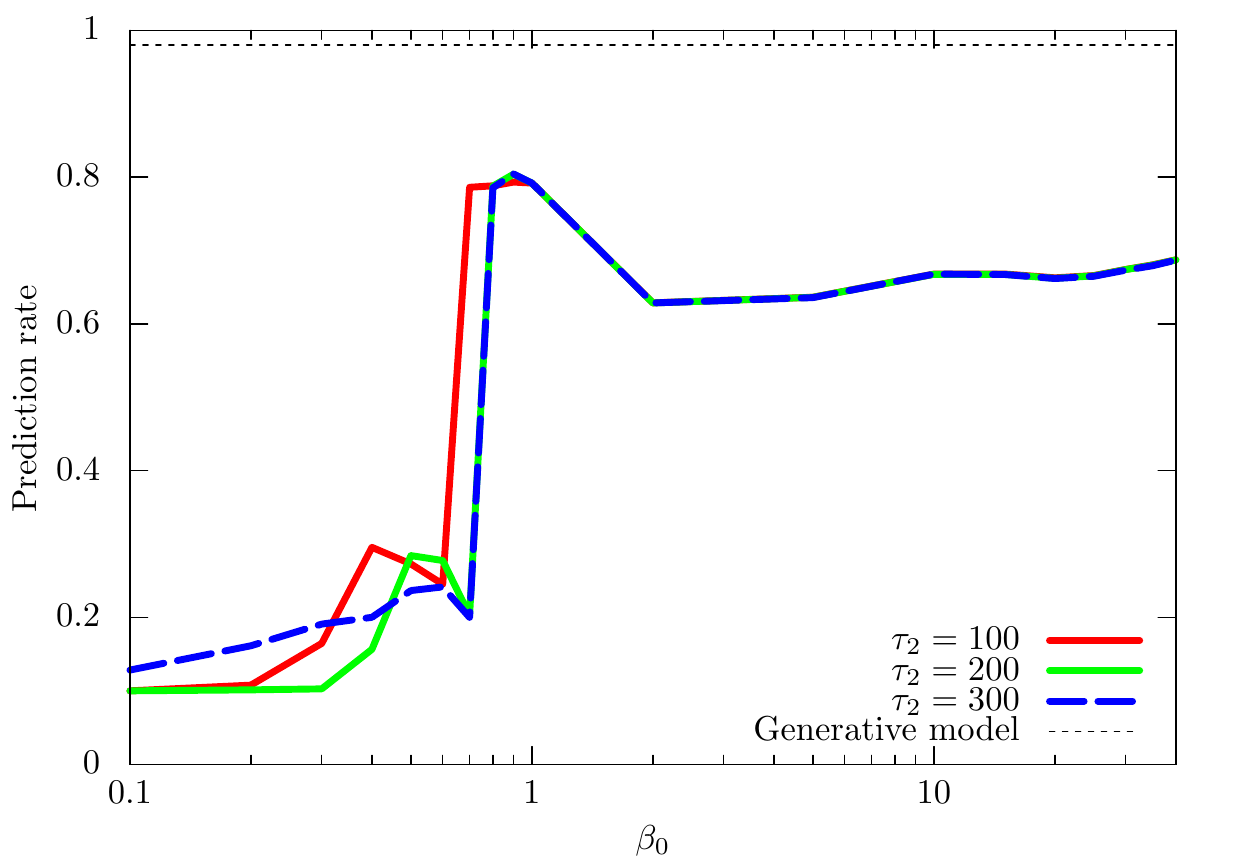}
\includegraphics[scale=0.50]{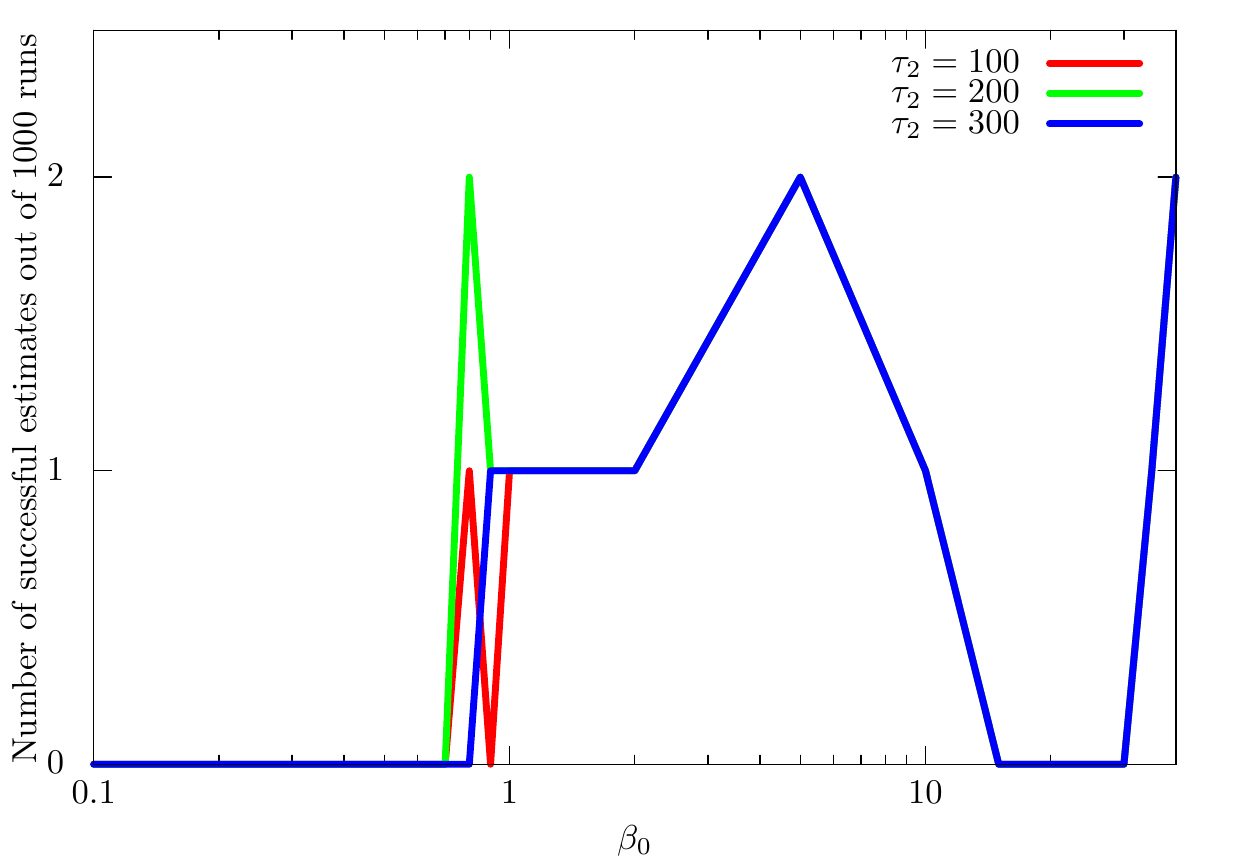}
\includegraphics[scale=0.50]{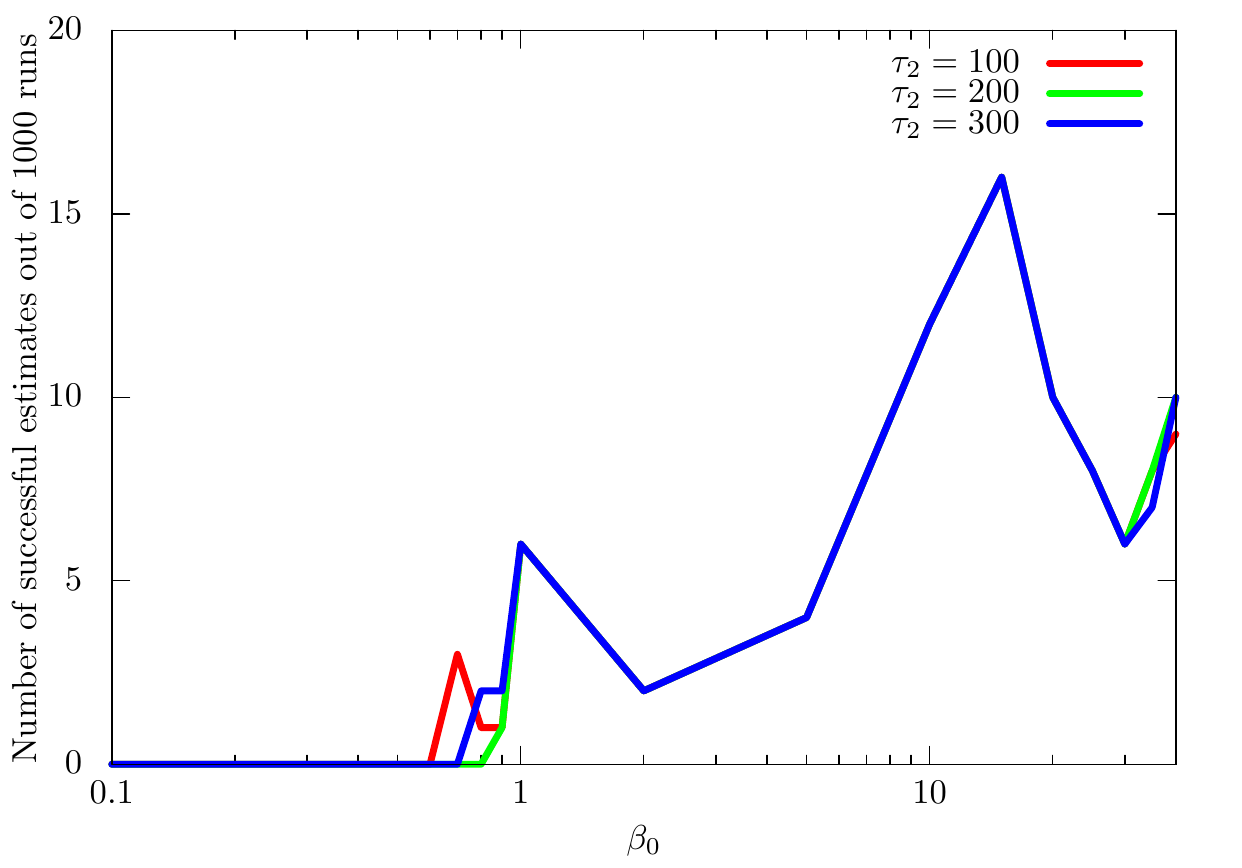}
\caption{\textbf{Comparative average performance of DAVB and its dependence on the initial temperature $\beta_0$}. Note that in DAVB the performance depends strongly on the initial distribution, i.e., $\hat{\rho}_{0}^\Sigma$. For the two dimensional dataset, we profile the dependence of the average prediction rate of DAVB on $\beta_0$ for (a) $K = 20$ and (b) $K = 30$. Number of times that achieve $p_\mathrm{cr} = 0.95$ out of 1000 runs for (c) $K = 20$ and (d) $K = 30$ are also plotted. As is expected, \textit{the average performance is inferior to that of QAVB, which for a wide range of choices of parameters gives a near-optimal result} \textbf{in a single run.} }
\label{main_numerical_002_002}
\end{figure}

Next, we turn our attention to DAVB.
We again use Eq.~\eqref{main_annealing_schedule_QAVB_001_002} for $\beta_t$ in DAVB, and set $\tau_1 = 10$ and $\tau_2 = 100, 200, 300$.
In Figs.~\ref{main_numerical_002_002}(a) and (b), we plot the dependence of the average prediction rate of DAVB on $\beta_0$ for $K = 20, 30$, respectively.
And, in Figs.~\ref{main_numerical_002_002}(c) and (d), we also plot the number of times that achieve $p_\mathrm{cr} = 0.95$ on $\beta_0$ for $K = 20, 30$, respectively.
These figures show that the average prediction rate of DAVB is much lower than that of QAVB and DAVB rarely achieves $p_\mathrm{cr} = 0.95$.

\begin{figure}[t]
\includegraphics[scale=0.50]{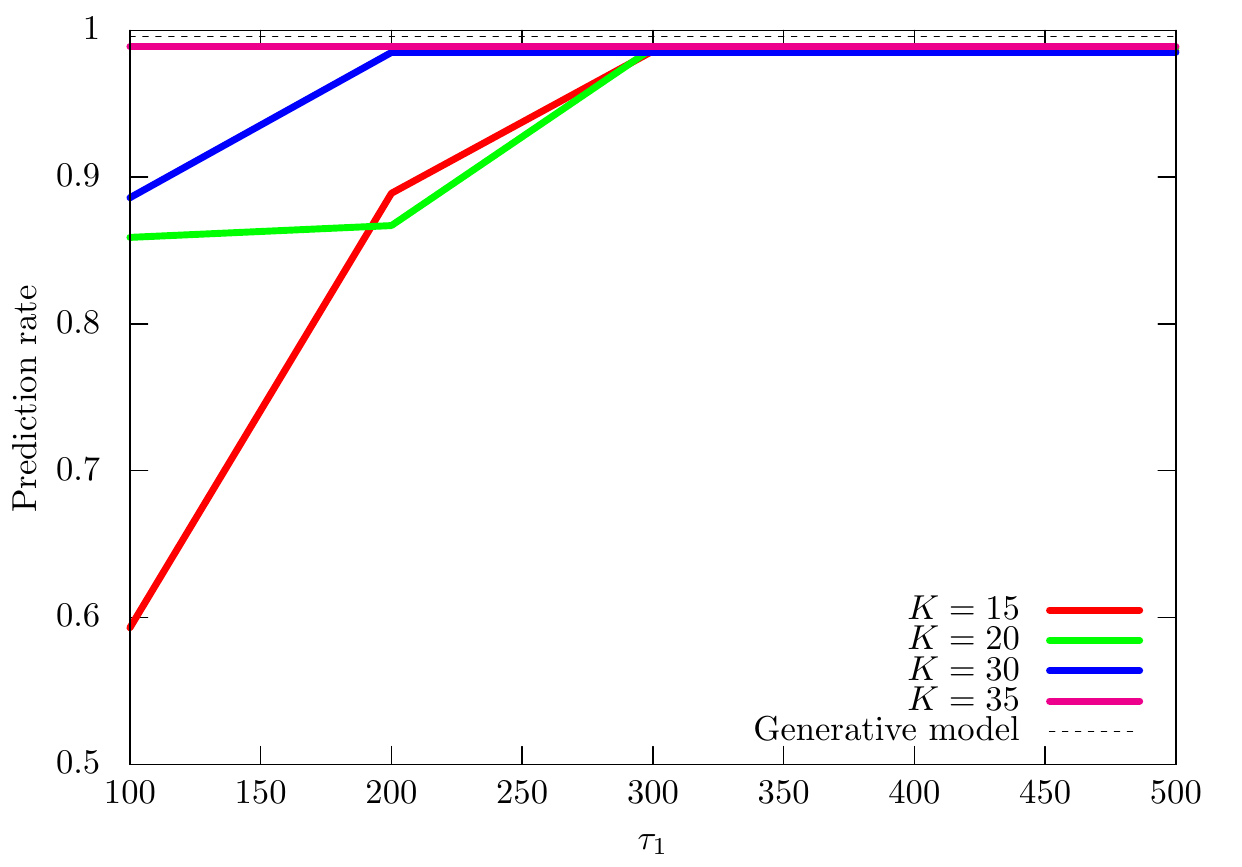}
\includegraphics[scale=0.50]{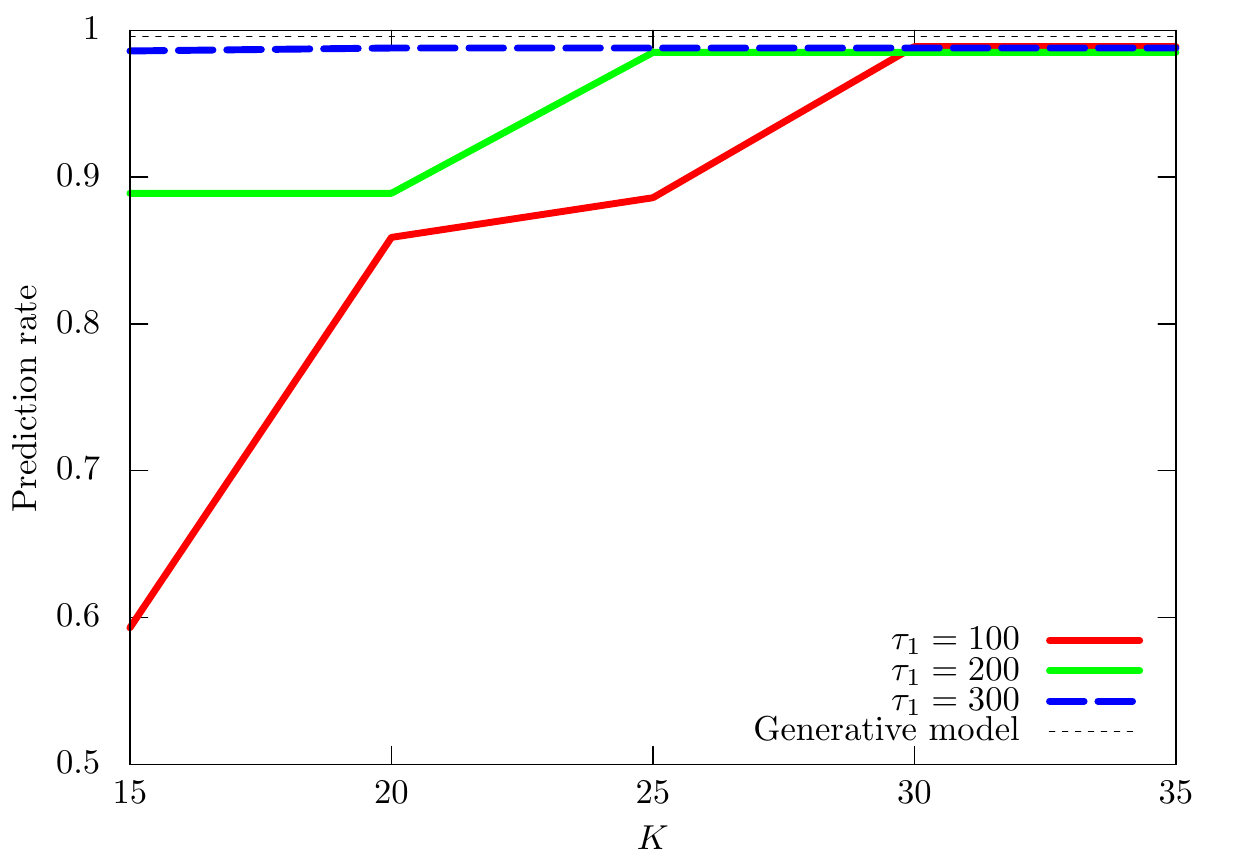}
\caption{Results paralleling that of Fig.~\ref{main_numerical_002_001} are shown here for a 3-D dataset: (a) Dependence of $p_\mathrm{suc}^K (K, \tau_1)$ on $\tau_1$ and (b) that of $p_\mathrm{suc}^{\tau_1} (K, \tau_1)$ on $K$. These exhibit very similar tradeoffs observed in Fig.~\ref{main_numerical_002_001} and gets very close to the optimal performance obtained from the ground truth generative model used to create the dataset (shown by the black dotted line).}
\label{main_numerical_003_001}
\end{figure}

\begin{figure}[t]
\includegraphics[scale=0.50]{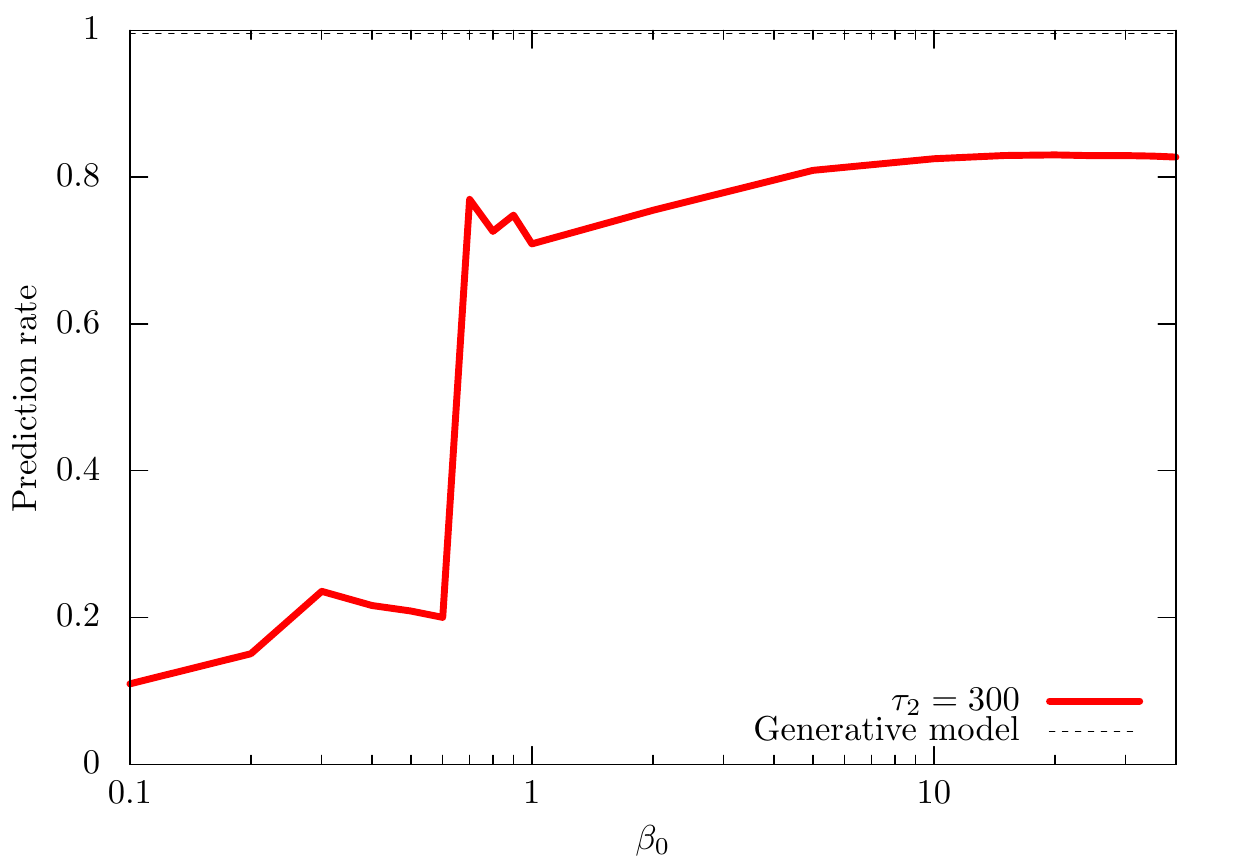}
\includegraphics[scale=0.50]{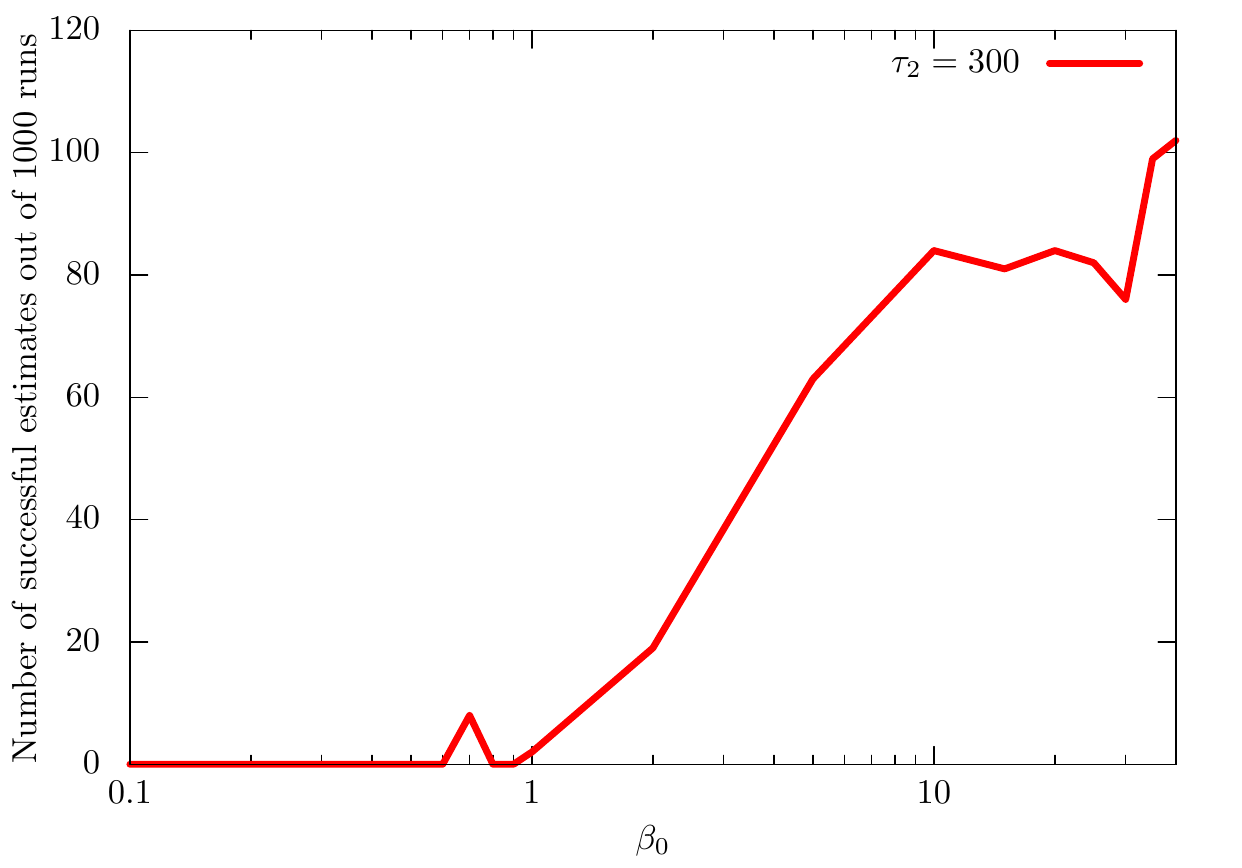}
\caption{Performance metrics of DAVB for the 3-D data analyzed in Fig.~\ref{main_numerical_003_001} are shown here: (a) Dependence of the average prediction rate of DAVB on $\beta_0$. (b) Number of times that achieve prediction rate $0.95$ out of 1000 runs. We set $K = 20$. A single run of QAVB can outperform DAVB. To get a relatively good performance from DAVB, one needs to start at a low temperature $\beta_0>> 1$and increase it to $\beta =1$; even then, the probability of getting a good prediction rate is low.}
\label{main_numerical_003_002}
\end{figure}

We next present numerical results on a three-dimensional dataset.
In Fig.~\ref{main_numerical_003_001}, we plot the dependence of $p_\mathrm{suc}^K (K, \tau_1)$ on $\tau_1$ and the dependence of $p_\mathrm{suc}^{\tau_1} (K, \tau_1)$ on $K$, respectively.
In Figs.~\ref{main_numerical_003_002}(a) and (b), we plot the dependence of the average prediction rate of DAVB and the number of times that achieve $p_\mathrm{cr} = 0.95$ on $\beta_0$, respectively.
Here we set $\tau_1 = 10$, $\tau_2 = 300$, and $K = 20$.
The numerical result on the three-dimensional dataset is consistent with the case of the two-dimensional dataset though it is quantitatively different from the case of the two-dimensional dataset.

\begin{figure}[t]
\includegraphics[scale=0.50]{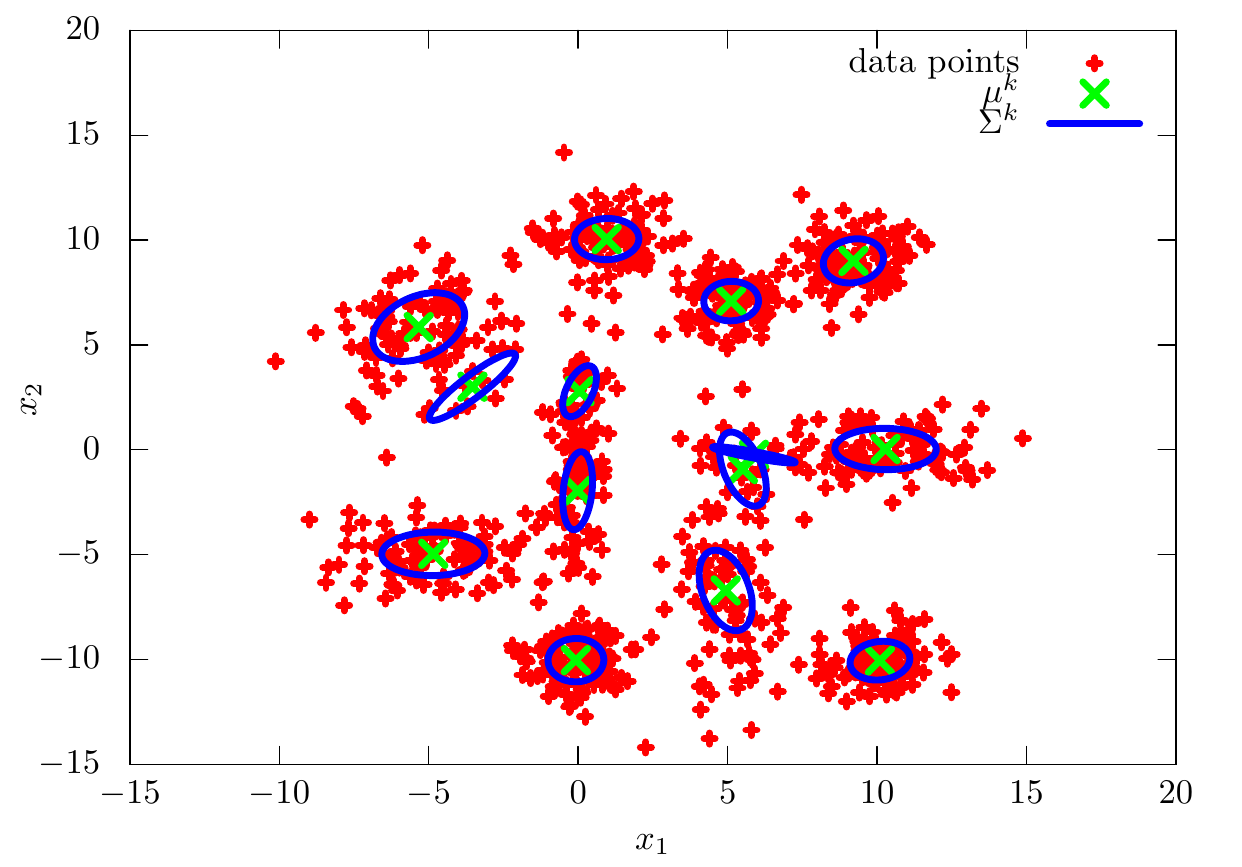}
\includegraphics[scale=0.50]{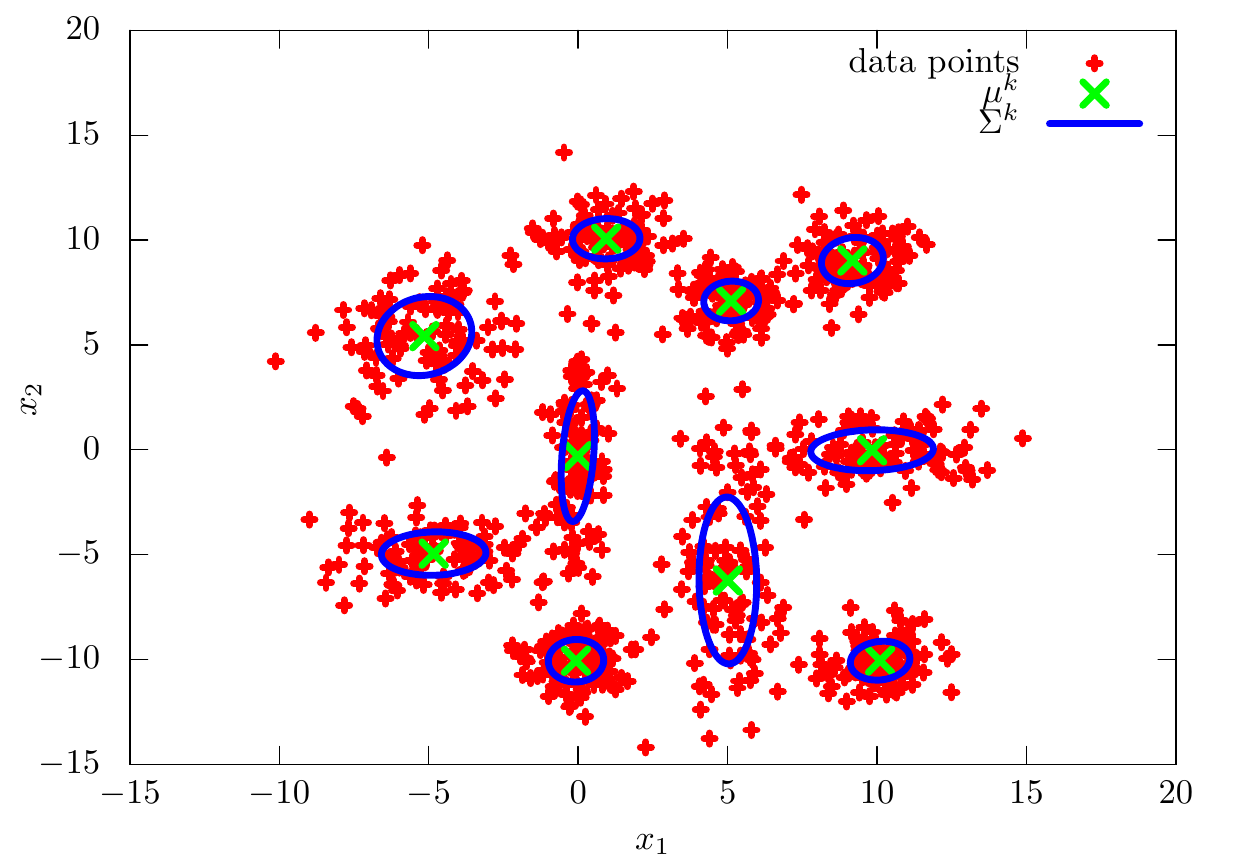}
\caption{\textbf{GMM estimates at the end of the QA step in QAVB and the critical role played by the QA part:} For the 2-D dataset in Fig.~\ref{main_numerical_001_001}(b) we visualize the estimated Gaussians functions, when $\beta_0 = 30$, and $K=20$.
(a) at step 300 ($\beta_{300} = 30.0$), i.e. at the end of the QA step;  and (b) at step 460 ($\beta_{460} = 1.0$), i.e. at the end of the QAVB algorithm. Only the Gaussian functions whose weight is greater than 0.01 are shown; that is when the probability of picking a Gaussian,  $\pi_j > 0.01$, then the $j$-th Gaussian function is shown for $j = 1, 2, \dots, K$. This shows that by the end of the QA step the algorithm has  found an almost-optimal solution,  and increasing temperature only fine tunes these estimates. This is further borne out by the results presented in Table~\ref{main_table_success_rate_001_001}.  }
\label{main_numerical_004_001}
\end{figure}

To understand the dynamics of QAVB, it is instructive to study cluster assignments of QAVB at the end of the QA part of the annealing schedule at $t = \tau_1$ and at convergence.
We show the cluster assignments of QAVB with $\beta_0 = 30.0$ in Figs.~\ref{main_numerical_004_001}(a) and (b).
The estimates at the end of the QA part are almost the same with those at convergence, respectively. Only a few of the clusters in the ground truth data are split.
Thus the process of raising temperature is not that important. This is expected due to the absence of SSB, and it is quite reasonable to focus on the QA part.
Figures~\ref{main_numerical_004_001}(a) and (b) show that, in the case of low temperature, QAVB successfully estimates the ground state while QAVB does not in the case of high temperature.
These results are also consistent with the discussions of the possible mechanism of QAVB.
In Table~\ref{main_table_success_rate_001_001}, we show the success rates of QAVB at convergence and at the end of the QA part and the best possible success rates with full knowledge of the generative model.
Ten datasets were created by using the same generative model to create the dataset shown in Fig.~\ref{main_numerical_001_001} and then the mean and standard deviation of the performance were computed.
The success rate of QAVB at convergence is very close to that of the generative model, and that of QAVB at the end of the QA part is also very close to them.
This observation reflects two points.
The first one is simply that QAVB is successful for the generative model under consideration.
The second one is that the soft clustering defined via the minimization problem of the KL divergence, which is Eq.~\eqref{main_optimization_problem_VB_002_001} at $\beta = 1$ and $s = 0$, and the hard clustering defined via that at $\beta \gg 1$ are similar. In the case of the hard clustering problem, however, the data points are classified into a larger number of clusters. That is, out of the $K$ Gaussians ($K>10$ where the generative model has 10 Gaussians) more than 10 have $\pi_j > 0.01$.
Thus, the success rate of QAVB at the end of the QA part is slightly worse than at convergence.
\begin{table}[t]
\caption{Success rates of QAVB at convergence and at the end of the QA part and the best achievable success rates from the generative model used to create datasets.
We created 10 datasets by using the same generative model used to create Fig.~\ref{main_numerical_001_001} and computed the mean and standard deviation of the performance. We set $K = 20$, $\tau_1 = 300$, and $\tau_2 = 50$. Note that $s_0 = 1.0$ and $\beta_0 = 30.0$.
QAVB achieves a high success rate that is close to the success rate of the generative model at convergence. Furthermore, the success rate of QAVB at the end of the QA part is also very close to that at convergence. This demonstrates that QAVB approximates the ground state well and obtains a good hard-clustering solution at the end of the QA step.
}
\label{main_table_success_rate_001_001}
\begin{tabular}{ccc}
\hline
at convergence & at the end of the QA part & best achievable \\
\hline
$0.9221 \pm 0.0497$ & $0.8919 \pm 0.0496$ & $0.9833 \pm 0.0045$ \\
\hline
\end{tabular}
\end{table}

\section{Adiabatic-theorem-like property: Similarity and differences between QA and QAVB}
In QA, the total Hamiltonian is constructed by the convex combination of a Hamiltonian that describes an optimization problem of interest and a noncommutative Hamiltonian that can be easily diagonalized.
Similar to QA, we construct Eq.~\eqref{main_Gibbs_operator_quantum_001_001} by the convex combination of two Hamiltonians. On the other hand, the main difference is that QA solves the Schr\"odinger equation but QAVB solves the MF equation.
Thus, \textit{the adiabatic theorem~\cite{Sakurai_001} does not directly hold}.
We next analytically examine QAVB and discuss an adiabatic-theorem-like property of QAVB.
First, let us consider the eigenvalue decomposition of Eq.~\eqref{main_Gibbs_operator_quantum_001_001}:
\begin{align}
\ln \hat{f} (\beta, s) &= \sum_{n = 0, 1, 2, \dots} \varepsilon_n (\beta, s) | n; \beta, s; \Sigma, \theta \rangle \langle n; \beta, s; \Sigma, \theta|. \label{main_Gibbs_operator_quantum_002_001}
\end{align}
Here, $\varepsilon_n (\beta, s)$ is the $(n+1)$-th largest eigenvalues with $\beta$ and $s$ for $n = 0, 1, 2, \dots$.
As explained before, QAVB is based on the MF theory; then, it is quite natural to consider the MF approximated form of Eq.~\eqref{main_Gibbs_operator_quantum_002_001}:
\begin{align}
  \ln \hat{f}^\mathrm{MF} (\beta, s) &= \sum_{n = 0, 1, 2, \dots} \varepsilon_n^\mathrm{MF} (\beta, s) | n; \beta, s; \Sigma \rangle \langle n; \beta, s; \Sigma| \nonumber \\
  & \quad \otimes | n; \beta, s; \theta \rangle \langle n; \beta, s; \theta|. \label{main_Gibbs_operator_quantum_002_002}
\end{align}
Here, $\varepsilon_n^\mathrm{MF} (\beta, s)$ is the $(n+1)$-th largest MF eigenvalues with $\beta$ and $s$ for $n = 0, 1, 2, \dots$, and $| n; \beta, s; \Sigma \rangle \otimes | n; \beta, s; \theta \rangle$ is the eigenvectors associated with $\varepsilon_n^\mathrm{MF} (\beta, s)$.
By using Eq.~\eqref{main_Gibbs_operator_quantum_002_002}, the update equations of QAVB, Eqs.~\eqref{main_QAVB_E-step_001_001} and \eqref{main_QAVB_M-step_001_001}, are rewritten as
\begin{align}
\ln \hat{\rho}_{t+1}^\Sigma &= \mathrm{Tr}_{\theta} \left[ \Big(\hat{I}^\Sigma \otimes \hat{\rho}_{t+1}^\theta \Big) \ln \hat{f}^\mathrm{MF} (\beta_t, s_t) \right] + \mathrm{const.}, \label{main_QAVB_MF_E-step_001_001} \\
\ln \hat{\rho}_{t+1}^\theta &= \mathrm{Tr}_{\Sigma} \left[ \Big( \hat{\rho}_t^\Sigma \otimes \hat{I}^\theta \Big) \ln \hat{f}^\mathrm{MF} (\beta_t, s_t) \right] + \mathrm{const.} \label{main_QAVB_MF_M-step_001_001}
\end{align}

Assuming that $\hat{\rho}_t^\Sigma = | 0; \beta_{t-1}, s_{t-1}; \Sigma \rangle \langle 0; \beta_{t-1}, s_{t-1}; \Sigma |$ and $\langle 0; \beta_{t-1}, s_{t-1}; \Sigma | 0; \beta_t, s_t; \Sigma \rangle \approx 1$, Eq.~\eqref{main_QAVB_MF_M-step_001_001} becomes
\begin{align}
\ln \hat{\rho}_{t+1}^\theta &= \mathrm{Tr}_{\Sigma} \Bigg[ \Big( | 0; \beta_{t-1}, s_{t-1}; \Sigma \rangle \langle 0; \beta_{t-1}, s_{t-1}; \Sigma | \otimes \hat{I}^\theta \Big) \nonumber \\
& \quad \times \sum_n \varepsilon_n^\mathrm{MF} (\beta, s) | n; \beta_t, s_t; \Sigma \rangle \langle n; \beta_t, s_t; \Sigma| \nonumber \\
& \quad \otimes | n; \beta_t, s_t; \theta \rangle \langle n; \beta_t, s_t; \theta| \Bigg] + \mathrm{const.} \\
&\approx \mathrm{Tr}_{\Sigma} \Big[ \Big( | 0; \beta_{t-1}, s_{t-1}; \Sigma \rangle \langle 0; \beta_{t-1}, s_{t-1}; \Sigma | \otimes \hat{I}^\theta \Big) \nonumber \\
& \quad \times \varepsilon_0^\mathrm{MF} (\beta, s) | 0; \beta_t, s_t; \Sigma \rangle \langle 0; \beta_t, s_t; \Sigma| \nonumber \\
& \quad \otimes | 0; \beta_t, s_t; \theta \rangle \langle 0; \beta_t, s_t; \theta| \Big] + \mathrm{const.} \\
&= | 0; \beta_t, s_t; \theta \rangle \langle 0; \beta_t, s_t; \theta| + \mathrm{const.},
\end{align}
and a similar computation can also be done for Eq.~\eqref{main_QAVB_MF_E-step_001_001}.
Note that a constant multiple does not affect the physical property of $\hat{\rho}_{t+1}^\theta$.
In the numerical simulations of QAVB, we varied $s$ at fixed $\beta$ at the first QA part; then the above assumption is reasonable.
Thus, the discussion here analytically gives the reason why QAVB shows high performance.
In particular, it explains the mechanism by which QAVB gives the ground state of $\hat{H}_\mathrm{cl}^{\Sigma| \theta}$ in Eq.~\eqref{main_classical_Hamiltonian_001_001}.

\section{Time complexity of QAVB and QAVB as quantum dynamics}
The main focus of this paper is to quantify how much better QAVB can perform compared with VB and provide analytical results on its dynamics. From the viewpoint of practical applications its computational complexity is also an important metric.
In the case of a classical computer, the time complexity of VB with respect to the number of cluster $K$ is $O (K)$ since VB has single loops on $K$.
On the other hand, QAVB requires one to compute the exponentials of $K \times K$ matrices; thus, the time complexity of QAVB with respect to $K$ is $O (K^3)$.
Note that, similarly to VB, the time complexity of QAVB with respect to the number of data points is $O (N)$; thus, it practically works on a classical computers.

The above situation changes if we assume a quantum computer.
The simulations of a quantum system does not increase time complexity compared with that of a classical system. Furthermore, if we can find a local Hamiltonian that describes QAVB, then we can expect a quantum speedup with respect to $K$~\cite{Childs_001}.
From this viewpoint, it is worth considering  physical implementations of QAVB.
From Eqs.~\eqref{main_QAVB_E-step_001_001} and \eqref{main_QAVB_M-step_001_001}, the relationship between $\hat{\rho}_t^\Sigma$ and $\hat{\rho}_{t+1}^\Sigma$ is written as
\begin{align}
\hat{\rho}_{t+1}^\Sigma &= \frac{1}{\mathcal{Z}_{t+1}} \exp \Big( \mathrm{Tr}_{\theta} \Big[ \Big( \hat{I}^\Sigma \nonumber \\
& \quad \otimes \exp \Big( \mathrm{Tr}_{\Sigma} \Big[ ( \hat{\rho}_t^\Sigma \otimes \hat{I}^\theta ) \ln \hat{f} (\beta_t, s_t) \Big] \Big) \Big) \ln \hat{f} (\beta_t, s_t) \Big] \Big), \label{main_dynamics_QAVB_Sigma_001_001}
\end{align}
where
\begin{align}
\mathcal{Z}_{t+1} &= \mathrm{Tr}_\Sigma \Big[ \exp \Big( \mathrm{Tr}_{\theta} \Big[ \Big( \hat{I}^\Sigma \nonumber \\
& \quad \otimes \exp \Big( \mathrm{Tr}_{\Sigma} \Big[ ( \hat{\rho}_t^\Sigma \otimes \hat{I}^\theta ) \ln \hat{f} (\beta_t, s_t) \Big] \Big) \Big) \ln \hat{f} (\beta_t, s_t) \Big] \Big) \Big].
\end{align}
In Eq.~\eqref{main_dynamics_QAVB_Sigma_001_001}, two types of operations are involved: the exponential operation and partial trace.
We show that these two operators are both CPTP because it can be realized in a physical process~\cite{Lidar_001}.

First, we discuss that the exponential map is CP when the input density operator is positive semidefinite.
Here, we basically follow Ref.~\cite{Accardi_001}.
It is enough to say that the composite operator $\exp \circ A_n$ is positive for $n = 1, 2, \dots$ where $A_n$ is an arbitrary $n$-dimensional operator.
As described in Ref.~\cite{Accardi_001}, the family of positive definite operators is closed under point-wise addition and point-wise multiplication; thus $\exp \circ A_n$ is positive for $n = 1, 2, \dots$; thus, the exponential map is completely positive.
Note that a Hamiltonian is not necessarily positive semidefinite, but we can always add a constant shift such that the Hamiltonian becomes positive semidefinite.
The map of interest is TP because of the partition function, though the exponential map itself is not TP.

Next, we turn our attention to partial trace.
We can say that partial trace operation is completely positive by constructing a Kraus operator directly.
Let us consider $\hat{K}_\alpha \coloneqq \hat{I}_\mathrm{A} \otimes \langle \alpha |$.
In general, a density operator for subsystems A and B has the form $\hat{\rho}_\mathrm{A B} \coloneqq \sum_{i j \mu \nu} \lambda_{i j \mu \nu} | i \rangle \langle j | \otimes | \mu \rangle \langle \nu |$,
and taking partial trace with respect to subsystem B yields $\mathrm{Tr}_\mathrm{B} [ \hat{\rho}_\mathrm{A B} ] = \sum_\alpha \sum_{i j \mu \nu} \lambda_{i j \mu \nu} | i \rangle \langle j | \langle \alpha | \mu \rangle \langle \nu | \alpha \rangle = \sum_{i j \mu} \lambda_{i j \mu \mu} | i \rangle \langle j |$.
On the other hand, $\hat{K}_\alpha$ leads to $\sum_\alpha \hat{K}_\alpha \hat{\rho}_\mathrm{A B} \hat{K}_\alpha^\dagger = \sum_{i j \mu} \lambda_{i j \mu \mu} | i \rangle \langle j | = \mathrm{Tr}_\mathrm{B} [ \hat{\rho}_\mathrm{A B} ]$.
Thus we have shown that $\hat{K}_\alpha$ is the Kraus operator for partial trace.
For more details, refer to Ref.~\cite{Lidar_001}.
Thus, we have shown that Eq.~\eqref{main_dynamics_QAVB_Sigma_001_001} is CPTP.
In other words, QAVB is physically implementable.

\section{Discussions}
In this paper, we have analyzed the dynamics of QAVB by developing an analytical framework and providing numerical simulations to support the analytical results. In particular, we developed a theoretical framework to understand why there is a quantum advantage in variational bayesian inference. Next, via numerical analysis, we confirmed that the QA part of QAVB is essential by showing that an estimate at the end of the QA part is almost the same as an estimate at convergence at finite temperature.
Thus, an optimal solution to the VB problem is essentially obtained at the end of the QA part, and increasing temperature does not affect the estimates very much.  We also showed that the estimate at the end of the QA part of annealing part corresponds to the hard clustering assignment.
Second, we developed  an adiabatic-theorem-like result that shows that the QA also holds in  the case of the MF dynamics. Then we explained that this generalized QA framework is why QAVB is efficient and gives a quantum advantage.
Finally, we discussed the physical realizability of QAVB by showing that QAVB can be expressed as a CPTP map.
This discussion tells us that QAVB can be realized in a quantum system.
We expect  this work to motivate physics-inspired algorithms and further research on the emerging fields at the intersection of physics and machine learning.

%
%

\appendix

\section{Introduction of Appendices}


This supplemental information (SI) covers background material such as, the Gaussian Mixture Model (GMM), and the conjugate priors used for various parameters in the GMM.
It also covers the update equations that result from using the mean field approximation in varitional Bayes (VB).
In addition, it introduces concepts from quantum mechanics, statistical mechanics, and quantum statistical mechanics that would make them more accessible to researchers with backgrounds in Machine Learning and AI.
It also introduces the framework of canonical distributions, and how it leads to deterministic annealing (DA) and DAVB.
Finally, it introduces a path integral formulation of QAVB.
Some of these background concepts are covered in Refs.~\cite{Miyahara_001, Miyahara_002} , and we have provided a more integrated view in this SI.

\section{Canonical distribution} \label{supp_canonical_distribution_001}

Let us consider $N_\mathrm{tot}$ systems together and assume that the total energy of the systems is $E_\mathrm{tot}$.
We denote, by $N_k$, the number of the systems in the $k$-th energy level for $n = 0, 1, 2, \dots, \infty$; then we have the following constraints:
\begin{align}
  N_\mathrm{tot} &= \sum_{k=0}^\infty N_k, \label{supp_eq_constraint_numbers_001_001} \\
  E_\mathrm{tot} &= \sum_{k=0}^\infty N_k \varepsilon_k. \label{supp_eq_constraint_energy_001_001}
\end{align}
On the other hand, the number of the configurations of the $N_\mathrm{tot}$ systems that satisfy Eqs.~\eqref{supp_eq_constraint_numbers_001_001} and \eqref{supp_eq_constraint_energy_001_001} is given by
\begin{align}
  N_\mathrm{con} &= \frac{N_\mathrm{tot}!}{\prod_{k=0}^\infty N_k!}. \label{supp_eq_number_states_001_001}
\end{align}
Then the most probable state in this setup is the configuration that maximizes Eq.~\eqref{supp_eq_number_states_001_001} under Eqs.~\eqref{supp_eq_constraint_numbers_001_001} and \eqref{supp_eq_constraint_energy_001_001}.
To solve this maximization problem, we use the method of Lagrange multipliers; the Lagrange function becomes
\begin{align}
  \ln N_\mathrm{con} - \alpha \sum_{k=0}^\infty N_k - \beta \sum_{k=0}^\infty N_k \varepsilon_k. \label{supp_eq_Lagrange_multiplier_001_001}
\end{align}

For $N \gg 0$, Stirling’s formula reads
\begin{align}
  N_\mathrm{tot}! &\approx N_\mathrm{tot} (\ln N_\mathrm{tot} - 1). \label{supp_eq_stirling_001_001}
\end{align}
Applying Eq.~\eqref{supp_eq_stirling_001_001} to Eq.~\eqref{supp_eq_Lagrange_multiplier_001_001}, we obtain
\begin{align}
  \ln N_\mathrm{con} - \alpha \sum_{k=0}^\infty N_k - \beta \sum_{k=0}^\infty N_k \varepsilon_k &= \ln \frac{N_\mathrm{tot}!}{\prod_{k=0}^\infty N_k!} - \alpha \sum_{k=0}^\infty N_k - \beta \sum_{k=0}^\infty N_k \varepsilon_k \\
  &\approx \ln \frac{N_\mathrm{tot} (\ln N_\mathrm{tot} - 1)}{\prod_{k=0}^\infty N_k (\ln N_k - 1)} - \alpha \sum_{k=0}^\infty N_k - \beta \sum_{k=0}^\infty N_k \varepsilon_k.
\end{align}
Then the variational conditions obtained from Eq.~\eqref{supp_eq_Lagrange_multiplier_001_001} with respect to $\{ N_k \}_{n=0}^\infty$ are, for $n = 0, 1, 2, \dots$,
\begin{align}
  0 &= - \delta N_k (\ln N_k + \alpha + \beta \varepsilon_k).
\end{align}
Then we obtain, for $n = 1, 2, \dots, \infty$.
\begin{align}
  N_k &= e^{- \alpha} e^{- \beta \varepsilon_k}.
\end{align}

Then, by introducing $p_k^\mathrm{can} \coloneqq \frac{N_k}{N}$, we have
\begin{align}
  p_k^\mathrm{can} &= \frac{e^{- \beta \varepsilon_k}}{\mathcal{Z}_\beta}, \label{supp_eq_canonical_weight_001_001}
\end{align}
where
\begin{align}
  \mathcal{Z}_\beta &\coloneqq \sum_{k=0}^\infty e^{- \beta \varepsilon_k}.
\end{align}
Eq.~\eqref{supp_eq_canonical_weight_001_001} is called the canonical distribution.
Note that the canonical distribution is not the unique equilibrium state of a system that is attached to a heat bath at inverse temperature $\beta$.
We also note that the quantum counterpart of Eq.~\eqref{supp_eq_canonical_weight_001_001}, which we call the Gibbs state is given by
\begin{align}
  \hat{\rho}_\beta &= \frac{1}{\mathcal{Z}_\beta} \sum_{k=0}^\infty e^{- \beta \varepsilon_k} | k \rangle \langle k |.
\end{align}
In Eq.~\eqref{supp_eq_canonical_weight_001_001}, we have derived the canonical distribution from the energy spectrum, that is, the Hamiltonian.
In QAVB, we use the argument in a reversed way; that is, we define the Hamiltonian of VB from a probability distribution that has parameters to be estimated:
\begin{align}
  \varepsilon_k &= - \frac{1}{\beta} \ln p_k^\mathrm{can} - \frac{1}{\beta} \ln \mathcal{Z}_\beta. \label{supp_eq_energy_level_from_canonical_001_001}
\end{align}
Note that, in QAVB, we set $\beta = 1$ when we define the Hamiltonian.
The quantum counterpart of Eq.~\eqref{supp_eq_energy_level_from_canonical_001_001} is
\begin{align}
  \hat{H} &= \sum_{k=0}^\infty \varepsilon_k | k \rangle \langle k |,
\end{align}
with Eq.~\eqref{supp_eq_energy_level_from_canonical_001_001}.

\section{Gaussian mixture model} \label{supp_sec_GMM_001}

In the main paper, to demonstrate the performance of QAVB, we consider the estimation problem of the parameters and the number of clusters of the GMM studied in Refs.~\cite{Bishop_001, Murphy_001}.
The joint probability distribution of the GMM over an observable variable $y_i$ and a hidden variable $\sigma_i$ conditioned by a set of parameters $\theta$ is given by
\begin{align}
p^{y, \sigma| \theta} (y_i, \sigma_i| \theta) &= \sum_{k=1}^K \pi^k \mathcal{N} (y_i| \mu^k, (\Lambda^k)^{-1}) \delta_{k, \sigma_i}, \label{supp_joint_GMM_001_001}
\end{align}
where $\delta_{k, \sigma_i}$ is the Kronecker delta function, $\{ \pi^k \}_{k=1}^K$ are the mixing coefficients of the GMM, and $\mathcal{N} (y_i| \mu^k, (\Lambda^k)^{-1})$ is a Gaussian distribution whose mean and precision, which is the inverse of its covariance, are $\mu^k$ and $\Lambda^k$, respectively~\footnote{We have not gone into detail about the prior distribution of the GMM $p_\mathrm{pr}^\theta (\theta)$ because we do not quantize it in this paper. See Refs.~\cite{Bishop_001, Murphy_001} if the reader is not familiar with it.}.
Here, we have assumed that each hidden variable $\sigma_i$ takes $1, \dots, K$; that is, $S^{\sigma_i} = \{k\}_{k=1}^K$ for each $i$.
Note that, by using the one-hot notation~\cite{Bishop_001, Murphy_001}, we can construct an equivalent quantization scheme~\cite{Miyahara_003, Miyahara_004}.
To simplify the notation, we denote $\{ \pi^k \}_{k=1}^K$, $\{ \mu^k \}_{k=1}^K$, and $\{ \Lambda^k \}_{k=1}^K$ by $\pi$, $\mu$, and $\Lambda$, respectively, and we refer to $\{\pi, \mu, \Lambda\}$ collectively as $\theta$.

Taking the logarithm of Eq.~\eqref{supp_joint_GMM_001_001}, we define the Hamiltonian of the GMM for $\sigma_i$ with $y_i = y_i^\mathrm{obs}$ as
\begin{align}
H_\mathrm{cl}^{\sigma_i| \theta} &= - \ln p^{y, \sigma| \theta} (y_i^\mathrm{obs}, \sigma_i | \theta).
\end{align}
Then the Hamiltonian of the GMM for $\Sigma = \{ \sigma_i \}_{i=1}^N$ with $Y = Y^\mathrm{obs}$ is given by $H_\mathrm{cl}^{\Sigma| \theta} = \sum_{i=1}^N H_\mathrm{cl}^{\sigma_i| \theta}$.
Using a classical Hamiltonian $\hat{H}_\mathrm{cl}^{\Sigma| \theta}$, we can also define the quantum representation of $H_\mathrm{cl}^{\Sigma| \theta}$.

To introduce quantum fluctuations into $\hat{H}_\mathrm{cl}^{\Sigma| \theta}$, we add a noncommutative term $\hat{H}_\mathrm{qu}^{\Sigma} = \sum_{i=1}^N \hat{H}_\mathrm{qu}^{\sigma_i}$ that satisfies $[\hat{H}_\mathrm{cl}^{\Sigma | \theta}, \hat{H}_\mathrm{qu}^\Sigma] \ne 0$.
In this paper, we adopt
\begin{align}
\hat{H}_\mathrm{qu}^{\sigma_i} &= \bigg( \overset{i-1}{\underset{j = 1}{\otimes}} \hat{I}^{\sigma_j} \bigg) \otimes \Vast( \sum_{\substack{k = 1, \dots, K, \\ l = k \pm 1}} \Ket{ \sigma_i = l } \Bra{ \sigma_i = k } \Vast) \otimes \bigg( \overset{N}{\underset{j = i+1}{\otimes}} \hat{I}^{\sigma_j} \bigg) \otimes \hat{I}^\theta,
\end{align}
where $\Ket{ \sigma_i = 0 } = \Ket{ \sigma_i = K }$, and $\Ket{ \sigma_i = K + 1 } = \Ket{ \sigma_i = 1 }$.
We note that the form of $\hat{H}_\mathrm{qu}^{\sigma_i}$ is not limited to the above definition and has arbitrariness in general.

\section{Prior and posterior distributions of the GMM} \label{supp_sec_GMM_prior_posterior_001_001}

In Sec.~\ref{supp_sec_GMM_001}, we explained the GMM and its Hamiltonian.
From the viewpoint of Bayes inference, prior and posterior distributions are quite important.
Thus, we review those of the GMM here.

\subsection{Prior distributions of the GMM} \label{supp_sub-sec_GMM_priors_001_001}

Conjugate prior distributions of a model yield posterior distributions that have the same functional forms with them.
Thus, we often use conjugate prior distributions as prior distributions in VB, and, in this case, we can transform the update rules of distributions of parameters into the update rules of hyperparameters, which characterize the conjugate prior distribution and the associated posterior distribution.

This section is devoted to introducing the conjugate prior distributions of the GMM.
The explanation is divided into two since the prior distribution of the mixture coefficients $\pi$ can be described independently from the other parameters, and the mean $\mu$ and precision $\Lambda$ depend on each other.

\subsubsection{Prior distribution of $\pi$}

Let us denote, by $p_\mathrm{pr}^\pi (\pi)$, the prior distribution of $\pi$.
The prior distribution of the mixture coefficients $\pi$ is given by the Dirichlet distribution:
\begin{align}
p_\mathrm{pr}^\pi (\pi) &= \mathcal{D} (\pi | \alpha_\mathrm{pr} ), \label{supp_eq_prior_pi_001_001}
\end{align}
where $\alpha_\mathrm{pr} = \{ \alpha_\mathrm{pr}^k \}_{k=1}^K$, $C( \alpha_\mathrm{pr})$ is the normalization constant, and the Dirichlet distribution reads
\begin{align}
\mathcal{D} (\pi | \alpha ) & = C( \alpha ) \prod_{k=1}^K \left( \pi^k \right)^{\alpha^k - 1},
\end{align}
where $\alpha = \{ \alpha^k \}_{k=1}^K$.
We will see that, in this case, the posterior distribution becomes the Dirichlet distribution again.

\subsubsection{Prior distributions of $\mu$ and $\Lambda$}

Next, we turn our attention to the prior distribution of $\mu$ and $\Lambda$.
The prior distribution of them can be expressed as the product of the prior distributions of $\mu$ and $\Lambda$:
\begin{align}
p_\mathrm{pr}^{\mu, \Lambda} ( \mu, \Lambda ) &= p_\mathrm{pr}^{\mu | \Lambda} ( \mu | \Lambda) p_\mathrm{pr}^\Lambda (\Lambda) \\
&= \prod_{k=1}^K p_\mathrm{pr}^{\mu^k| \Lambda^k} ( \mu^k | \Lambda^k ) p_\mathrm{pr}^{\Lambda^k} (\Lambda^k),
\end{align}
where
\begin{align}
p_\mathrm{pr}^{\mu^k| \Lambda^k} ( \mu^k | \Lambda^k ) &= \mathcal{N} (\mu^k | m_\mathrm{pr}^k, (\gamma_\mathrm{pr}^k \Lambda^k)^{-1}), \label{supp_eq_prior_mu_001_001} \\
p_\mathrm{pr}^{\Lambda^k} (\Lambda^k) &= \mathcal{W} (\Lambda^k | W_\mathrm{pr}^k, \nu_\mathrm{pr}^k). \label{supp_eq_prior_Lambda_001_001}
\end{align}
Here, the Wishart distribution is defined as
\begin{align}
\mathcal{W} ( \Lambda^k | W^k, \nu^k) &= B | \Lambda^k |^{(\nu^k - D - 1) / 2} \exp \left( - \frac{1}{2} \mathrm{Tr} \left[ (W^k)^{-1} \Lambda^k \right] \right),
\end{align}
with
\begin{align}
B(W^k, \nu^k) &= | W^k |^{- \nu / 2} \left( 2^{\nu D / 2} \pi^{D(D-1) / 4} \prod_{i=1}^D \Gamma \left( \frac{\nu^k + 1 - i}{2} \right) \right)^{-1}, \label{supp_eq_Wihsart_normalization_001_001}
\end{align}
and $D$ is the number of dimensions.
Note that the Wishart distribution is defined when $\nu^k > D - 1$ for $k = 1, \dots, K$.
The Gamma function $\Gamma (z)$ in Eq.~\eqref{supp_eq_Wihsart_normalization_001_001} is given by
\begin{align}
\Gamma (z) &= \int_0^\infty du \, u^{z-1} e^{-u},
\end{align}
for $\mathrm{Re} [z] \ge 0$.

\subsection{Posterior distribution of $\Sigma$ in VB}

Here, we explicitly write down the update rule of hidden variables $\Sigma = \{ \sigma_i \}_{i=1}^N$ in the GMM.
The posterior distribution of hidden variables $\Sigma = \{ \sigma_i \}_{i=1}^N$ is, in general, updated as
\begin{align}
q_{t+1}^{\Sigma} (\Sigma) &\propto \exp \bigg( \int_{\theta \in S^\theta} d\theta\, q_{t+1}^\theta(\theta) \ln p^{Y, \Sigma, \theta} (Y, \Sigma, \theta) \bigg). \label{supp_eq_VB_E-step_Sigma_001_001}
\end{align}
To simplify expressions, we introduce
\begin{align}
\mathbb{E}_{q^\theta (\theta)} \left[ \dots \right] &\coloneqq \int_{\theta \in S^\theta} d \theta \, q^\theta (\theta) [ \dots ]. \label{supp_eq_expectation_theta_001_001}
\end{align}
Then, Eq.~\eqref{supp_eq_VB_E-step_Sigma_001_001} for the GMM is written as
\begin{align}
\ln q_{t+1}^{\sigma_i} (\sigma_i) &= \mathbb{E}_{q_{t+1}^\theta (\theta)} \Big[ \ln p^{y, \sigma, \theta} (y_i, \sigma_i, \theta) \Big] \\
&= \mathbb{E}_{q_{t+1}^\pi (\pi)} \left[ \ln p^{\sigma| \pi} (\sigma_i| \pi) \right] + \mathbb{E}_{q_{t+1}^{\mu, \Lambda} (\mu, \Lambda)} \left[ \ln p^{y| \sigma, \mu, \Lambda} (y_i| \sigma_i, \mu, \Lambda) \right] + \mathrm{const.}
\end{align}
By simple calculations, we also have
\begin{align}
\ln q_{t+1}^{\sigma_i} (\sigma_i) &= - \sum_{\sigma_i \in S^\sigma} \mathbb{E}_{q_{t+1}^\theta (\theta)} \Big[ H_\mathrm{cl}^{\sigma_i| \theta} \Big] \delta_{k, \sigma_i} + \mathrm{const.}, \label{supp_eq_VB_E-step_sigma_001_002}
\end{align}
where $H_\mathrm{cl}^{\sigma_i = k} = - \ln \pi^k \mathcal{N} (y_i | \mu^k, (\Lambda^k)^{-1})$.
Taking the exponential of both sides of Eq.~\eqref{supp_eq_VB_E-step_sigma_001_002} and normalizing, we finally obtain the update rules of hidden variables in the GMM:
\begin{align}
q_{t+1}^{\sigma_i} (\sigma_i) &= \sum_{k=1}^K \delta_{k, \sigma_i} r_i^k,
\end{align}
where
\begin{align}
r_i^k &= \frac{\exp \Big( - \mathbb{E}_{q_{t+1}^\theta (\theta)} \Big[ H_\mathrm{cl}^{\sigma_i = k| \theta} \Big] \Big)}{\sum_{\sigma_i \in S^\sigma} \exp \Big( - \mathbb{E}_{q_{t+1}^\theta (\theta)} \Big[ H_\mathrm{cl}^{\sigma_i| \theta} \Big] \Big)},
\end{align}
for $i = 1, 2, \dots, N$.

\subsection{Posterior distributions of $\theta$ in VB}

The update rule for the parameter $\theta$ is written as
\begin{align}
q_{t+1}^{\theta} (\theta) &\propto p_\mathrm{pr}^\theta (\theta) \exp \Bigg( \sum_{\Sigma \in S^{\Sigma}} q_t^{\Sigma} (\Sigma) \ln p^{Y, \Sigma| \theta} (Y, \Sigma| \theta) \Bigg) \\
&= p_\mathrm{pr}^\theta (\theta) \prod_{i=1}^N \exp \Bigg( \sum_{\sigma_i \in S^\sigma} q_t^{\sigma_i}(\sigma_i) \ln p^{y, \sigma| \theta} (y_i, \sigma_i| \theta) \Bigg). \label{supp_eq_VB_M-step_theta_001_001}
\end{align}

We will write down an explicit formula of the update rules of the GMM using Eq.~\eqref{supp_eq_VB_M-step_theta_001_001}.
For later convenience, we first define some statistics as follows:
\begin{subequations}
\begin{align}
N^k &= \sum_{i=1}^N r_i^k, \\
\bar{y}^k &= \frac{1}{N_k} \sum_{i=1}^N r_i^k y_i, \\
S^k &= \frac{1}{N_k} \sum_{i=1}^N r_i^k (y_i - \bar{y}^k) (y_i - \bar{y}^k)^\intercal.
\end{align}
\end{subequations}
Using Eq.~\eqref{supp_eq_VB_M-step_theta_001_001}, the posterior distribution of $\theta$ is computed as
\begin{align}
\ln q_{t+1}^{\theta} (\pi, \mu, \Lambda) &= \ln p_\mathrm{pr}^\pi (\pi) + \sum_{k=1}^K \Big[ \ln p_\mathrm{pr}^{\mu^k| \Lambda^k} (\mu^k| \Lambda^k) + \ln p_\mathrm{pr}^{\Lambda^k} (\Lambda^k) \Big] + \sum_{i=1}^N \mathbb{E}_{q_t^{\sigma_i} (\sigma_i)} \left[ \ln p^{\sigma | \pi} (\sigma_i | \pi) \right] \nonumber \\
& \quad + \sum_{i=1}^N \sum_{k=1}^K \mathbb{E}_{q_t^{\sigma_i} (\sigma_i)} \Big[ \ln p^{y| \sigma, \pi, \mu, \Lambda}(y_i| \sigma_i, \pi, \mu, \Lambda) \Big] + \mathrm{const.} \label{supp_eq_VB_M-step_GMM_001_001}
\end{align}
Similarly to Eq.~\eqref{supp_eq_expectation_theta_001_001}, we have also defined
\begin{align}
\mathbb{E}_{q^{\sigma_i} (\sigma_i)} \left[ \dots \right] &\coloneqq \sum_{\sigma_i \in S^\sigma} q^{\sigma_i} (\sigma_i) [ \dots ].
\end{align}
To go further, we decompose $p_\mathrm{pr}^\theta (\pi, \mu, \Lambda) = p_\mathrm{pr}^\pi (\pi) p^{\mu, \Lambda} (\mu, \Lambda)$ as in Sec.~\ref{supp_sub-sec_GMM_priors_001_001}, and show the details as follows.

\subsubsection{Posterior distribution of $\pi$}

We first consider the posterior distribution of $\pi$.
From Eqs.~\eqref{supp_eq_prior_pi_001_001} and \eqref{supp_eq_VB_M-step_GMM_001_001}, we obtain the posterior distributions as
\begin{align}
\ln q_{t+1}^{\pi} (\pi) &= \ln p_\mathrm{pr}^\pi (\pi) + \sum_{i=1}^N \mathbb{E}_{q_t^{\sigma_i} (\sigma_i)} \left[ \ln p^{\sigma | \pi} (\sigma_i | \pi) \right] + \mathrm{const.} \\
&= \sum_{k=1}^K (\alpha_\mathrm{pr}^k - 1) \ln \pi^k + \sum_{k=1}^K \Bigg( \sum_{i=1}^N r_i^k \Bigg) \ln \pi^k + \mathrm{const.} \\
&= \sum_{k=1}^K (\alpha_\mathrm{pr}^k - 1) \ln \pi^k + \sum_{k=1}^K N^k \ln \pi^k + \mathrm{const.},
\end{align}
in which we have used
\begin{align}
\mathbb{E}_{q^{\sigma_i} (\sigma_i)} \Big[ \delta_{k, \sigma_i} \Big] &= r_i^k, \\
\sum_{i=1}^N \mathbb{E}_{q^{\sigma_i} (\sigma_i)} \left[ \ln p^{\sigma | \pi} (\sigma_i | \pi) \right] &= \Bigg( \sum_{i=1}^N r_i^k \Bigg) \ln \pi^k \\
&= N^k \ln \pi^k.
\end{align}
Then we get
\begin{align}
q_{t+1}^{\pi} (\pi) &= \mathcal{D} (\pi | \alpha), \label{supp_eq_posterior_pi_001_001}
\end{align}
where $\alpha = \{\alpha^k\}_{k=1}^K$ and
\begin{align}
\alpha^k &= \alpha_\mathrm{pr}^k + N^k, \label{supp_eq_posterior_pi_001_002}
\end{align}
for $k = 1, 2, \dots, K$.
Note that Eq.~\eqref{supp_eq_posterior_pi_001_001} has the same functional form with Eq.~\eqref{supp_eq_prior_pi_001_001}.
This property is important to simplify the calculations of VB.

\subsubsection{Posterior distributions of $\mu$ and $\Lambda$}

Next, we derive the update rules for posterior distributions of $\mu$ and $\Lambda$.
From Eqs.~\eqref{supp_eq_prior_mu_001_001}, \eqref{supp_eq_prior_Lambda_001_001}, and \eqref{supp_eq_VB_M-step_GMM_001_001}, we obtain the posterior distributions as
\begin{align}
& \ln q_{t+1}^{\mu, \Lambda} (\mu, \Lambda) \nonumber \\
& \quad = \sum_{k=1}^K \Big[ \ln p_\mathrm{pr}^{\mu^k| \Lambda^k} (\mu^k| \Lambda^k) + \ln p_\mathrm{pr}^{\Lambda^k} (\Lambda^k) \Big] + \sum_{i=1}^N \mathbb{E}_{q_t^{\sigma_i} (\sigma_i)} \Big[ \ln p^{y| \sigma, \pi, \mu, \Lambda}(y_i| \sigma_i, \pi, \mu, \Lambda) \Big] + \mathrm{const.} \\
& \quad = \sum_{k=1}^K \Big[ \ln \mathcal{N} (\mu^k | m_\mathrm{pr}^k, (\gamma_\mathrm{pr}^k \Lambda^k)^{-1}) + \ln \mathcal{W} (\Lambda^k | W_\mathrm{pr}^k, \nu_\mathrm{pr}^k) \Big] + \sum_{i=1}^N \sum_{k=1}^K \mathbb{E}_{q_t^{\sigma_i} (\sigma_i)} \Big[ \delta_{k, \sigma_i} \Big] \ln \mathcal{N} (y_i| \mu^k, (\Lambda^k)^{-1}) + \mathrm{const.} \\
& \quad = \sum_{k=1}^K \Big[ \ln \mathcal{N} (\mu^k | m_\mathrm{pr}^k, (\gamma_\mathrm{pr}^k \Lambda^k)^{-1}) + \ln \mathcal{W} (\Lambda^k | W_\mathrm{pr}^k, \nu_\mathrm{pr}^k) \Big] + \sum_{i=1}^N \sum_{k=1}^K r_i^k \ln \mathcal{N} (y_i| \mu^k, (\Lambda^k)^{-1}) + \mathrm{const.}
\end{align}
Then, with simple calculations, we have the updated distribution over the parameters $\mu$ and $\Sigma$ given by
\begin{subequations}
\begin{align}
q_{t+1}^{\mu^k| \Lambda^k} ( \mu^k | \Lambda^k ) &= \mathcal{N} (\mu^k | m^k, (\gamma^k \Lambda^k)^{-1}), \\
q_{t+1}^{\Lambda^k} (\Lambda^k) &= \mathcal{W} (\Lambda^k | W^k, \nu^k),
\end{align}
\end{subequations}
where
\begin{subequations}
\begin{align}
\gamma^k &= \gamma_\mathrm{pr}^k + N^k, \label{supp_eq_posterior_mu_021_041} \\
m^k &= \frac{1}{\gamma^k} ( \gamma_\mathrm{pr}^k m_\mathrm{pr}^k + N^k \bar{y}^k),\label{supp_eq_posterior_mu_021_042} \\
(W^k)^{-1} &= {(W_\mathrm{pr}^k)}^{-1} + N^k S^k + \frac{\gamma_\mathrm{pr}^k N^k}{\gamma^k} (\bar{y}^k - m_\mathrm{pr}^k) (\bar{y}^k - m_\mathrm{pr}^k)^\intercal, \label{supp_eq_posterior)_Lambda_021_051} \\
\nu^k &= \nu_\mathrm{pr}^k + N^k. \label{supp_eq_posterior)_Lambda_021_061}
\end{align}
\end{subequations}
When we use the implementations of VB and QAVB described, we set $\nu_0^k = D - 1$ since $N^k > 0$ almost surely if initial distributions of $\Sigma$ is properly initialized.

\section{Posterior distribution in DAVB}

In the previous section, we introduced the conjugate prior distribution and the associated posterior distribution of the GMM in the case of VB.
Here, we consider the posterior distributions of the GMM in the case of DAVB.

\subsection{Quick review of DAVB}

We here try to briefly explain the key idea of DAVB~\cite{Katahira_001}.
In DAVB, we introduce thermal fluctuations into probability distributions by assuming the canonical distribution; then the total joint probability distribution becomes
\begin{align}
p_\mathrm{SA}^{Y, \Sigma, \theta} (Y, \Sigma, \theta; \beta^\mathrm{pr}, \beta) &\coloneqq \Big[ p_\mathrm{pr}^\theta (\theta) \Big]^{\beta^\mathrm{pr}} \Big[ p^{Y, \Sigma| \theta} (Y, \Sigma| \theta) \Big]^\beta, \label{supp_eq_distribution_DAVB_001_001}
\end{align}
We stress that Eq.~\eqref{supp_eq_distribution_DAVB_001_001} has two inverse temperatures for the prior distribution and the GMM: $\beta^\mathrm{pr}$ and $\beta$.
It will turn out to be reasonable to introduce two different inverse temperatures.

The update equations of $\Sigma$ and $\theta$ for DAVB is then computed by minimizing the following KL divergence:
\begin{align}
\mathrm{KL} \left( q^{\Sigma, \theta} (\Sigma, \theta) \middle\| p_\mathrm{SA}^{\Sigma, \theta | Y} (\Sigma, \theta | Y; \beta^\mathrm{pr}, \beta) \right) &\coloneqq - \sum_{\Sigma \in S^\Sigma} \int_{\theta \in S^\theta} d\theta \, q^{\Sigma, \theta} (\Sigma, \theta) \ln \frac{p_\mathrm{SA}^{\Sigma, \theta | Y} (\Sigma, \theta | Y; \beta^\mathrm{pr}, \beta)}{q^{\Sigma, \theta} (\Sigma, \theta)}.
\end{align}

\subsection{Posterior distribution of $\sigma_i$ in DAVB}

Following the same procedure employed in VB, we also obtain
\begin{align}
\ln q_{t+1}^{\sigma_i} (\sigma_i) &= - \beta \sum_{\sigma_i \in S^\sigma} \mathbb{E}_{q_{t+1}^\theta (\theta)} \Big[ H_\mathrm{cl}^{\sigma_i| \theta} \Big] \delta_{k, \sigma_i} + \mathrm{const.}, \label{supp_eq_update_DAVB_sigma_005_001}
\end{align}
where $H_\mathrm{cl}^{\sigma_i = k} = - \ln \pi^k \mathcal{N} (y_i | \mu^k, (\Lambda^k)^{-1})$.
Taking the exponential of Eq.~\eqref{supp_eq_update_DAVB_sigma_005_001}, we obtain
\begin{align}
q_{t+1}^{\sigma_i} (\sigma_i) &= \sum_{k=1}^K \delta_{k, \sigma_i} r_i^k,
\end{align}
where
\begin{align}
r_i^k &= \frac{\exp \Big( - \beta \mathbb{E}_{q_{t+1}^\theta (\theta)} \Big[ H_\mathrm{cl}^{\sigma_i = k| \theta} \Big] \Big)}{\sum_{\sigma_i \in S^\sigma} \exp \Big( - \beta \mathbb{E}_{q_{t+1}^\theta (\theta)} \Big[ H_\mathrm{cl}^{\sigma_i| \theta} \Big] \Big)}.
\end{align}

\subsection{Posterior distribution of $\theta$ in DAVB}

Next, we turn our attention to the update rules of $\theta$ in the case of DAVB.

\subsubsection{Posterior distribution of $\pi$ in DAVB}

We first consider the update rule of $\pi$.
The posterior distribution of $\pi$ in the case of DAVB is computed as
\begin{align}
\ln q_{t+1}^{\pi} (\pi) &= \beta^\mathrm{pr} \ln p_\mathrm{pr}^\pi (\pi) + \beta \sum_{i=1}^N \mathbb{E}_{q_t^{\sigma_i} (\sigma_i)} \left[ \ln p^{\sigma | \pi} (\sigma_i | \pi) \right] + \mathrm{const.} \\
&= \sum_{k=1}^K \beta^\mathrm{pr} (\alpha_\mathrm{pr}^k - 1) \ln \pi^k + \beta \sum_{k=1}^K \Bigg( \sum_{i=1}^N r_i^k \Bigg) \ln \pi^k + \mathrm{const.} \\
&= \sum_{k=1}^K \beta^\mathrm{pr} (\alpha_\mathrm{pr}^k - 1) \ln \pi^k + \beta \sum_{k=1}^K N^k \ln \pi^k + \mathrm{const.} \\
&= \sum_{k=1}^K \Big[ \Big\{ \beta^\mathrm{pr} (\alpha_\mathrm{pr}^k - 1) + \beta N^k + 1 \Big\} - 1 \Big] \ln \pi^k + \mathrm{const.}
\end{align}
Thus, Eq.~\eqref{supp_eq_posterior_pi_001_002}, which is the update rule of $\pi$ in VB, becomes
\begin{align}
\alpha^k &= \beta^\mathrm{pr} (\alpha_\mathrm{pr}^k - 1) + \beta N^k + 1.
\end{align}

\subsubsection{Posterior distributions of $\mu$ and $\Sigma$ in DAVB}

Next, we compute the posterior distribution of $\mu$ and $\Lambda$ in the case of DAVB:
\begin{align}
& \ln q_{t+1}^{\mu, \Lambda} (\mu, \Lambda) \nonumber \\
& \quad = \beta^\mathrm{pr} \sum_{k=1}^K \Big[ \ln p_\mathrm{pr}^{\mu^k| \Lambda^k} (\mu^k| \Lambda^k) + \ln p_\mathrm{pr}^{\Lambda^k} (\Lambda^k) \Big] + \beta \sum_{i=1}^N \mathbb{E}_{q_t^{\sigma_i} (\sigma_i)} \Big[ \ln p^{y| \sigma, \pi, \mu, \Lambda}(y_i| \sigma_i, \pi, \mu, \Lambda) \Big] + \mathrm{const.} \\
& \quad = \beta^\mathrm{pr} \sum_{k=1}^K \Big[ \ln \mathcal{N} (\mu^k | m_\mathrm{pr}^k, (\gamma_\mathrm{pr}^k \Lambda^k)^{-1}) + \ln \mathcal{W} (\Lambda^k | W_\mathrm{pr}^k, \nu_\mathrm{pr}^k) \Big] + \beta \sum_{i=1}^N \sum_{k=1}^K \mathbb{E}_{q_t^{\sigma_i} (\sigma_i)} \Big[ \delta_{k, \sigma_i} \Big] \ln \mathcal{N} (y_i| \mu^k, (\Lambda^k)^{-1}) + \mathrm{const.} \\
& \quad = \beta^\mathrm{pr} \sum_{k=1}^K \Big[ \ln \mathcal{N} (\mu^k | m_\mathrm{pr}^k, (\gamma_\mathrm{pr}^k \Lambda^k)^{-1}) + \ln \mathcal{W} (\Lambda^k | W_\mathrm{pr}^k, \nu_\mathrm{pr}^k) \Big] + \beta \sum_{k=1}^K \Bigg( \sum_{i=1}^N r_i^k \Bigg) \ln \mathcal{N} (y_i| \mu^k, (\Lambda^k)^{-1}) + \mathrm{const.} \\
& \quad = \beta^\mathrm{pr} \sum_{k=1}^K \Big[ \ln \mathcal{N} (\mu^k | m_\mathrm{pr}^k, (\gamma_\mathrm{pr}^k \Lambda^k)^{-1}) + \ln \mathcal{W} (\Lambda^k | W_\mathrm{pr}^k, \nu_\mathrm{pr}^k) \Big] + \beta \sum_{k=1}^K N^k \ln \mathcal{N} (y_i| \mu^k, (\Lambda^k)^{-1}) + \mathrm{const.}
\end{align}
Then the posterior distribution of $\mu^k$ conditioned by $\Lambda^k$ is written as
\begin{align}
q_{t+1}^{\mu^k| \Lambda^k} (\mu^k| \Lambda^k) &= \mathcal{N} (\mu^k| m^k, (\gamma^k \Lambda^k)^{-1}),
\end{align}
where
\begin{align}
\gamma^k &= \beta^\mathrm{pr} \gamma_\mathrm{pr}^k + \beta N^k, \label{supp_eq_posterior_gamma_031_041} \\
m^k &= \frac{1}{\gamma^k} ( \beta^\mathrm{pr} \gamma_\mathrm{pr}^k m_\mathrm{pr}^k + \beta N^k \bar{y}^k). \label{supp_eq_posterior_gamma_031_042}
\end{align}
Thus, Eqs.~\eqref{supp_eq_posterior_mu_021_041} and \eqref{supp_eq_posterior_mu_021_042} are transformed to Eqs.~\eqref{supp_eq_posterior_gamma_031_041} and \eqref{supp_eq_posterior_gamma_031_042}.
Furthermore, Bayes' theorem leads to the following calculation:
\begin{align}
\ln q_{t+1}^{\Lambda^k} (\Lambda^k) &= \ln q_{t+1}^{\mu^k, \Lambda^k} (\mu^k, \Lambda^k) - \ln q_{t+1}^{\mu^k| \Lambda^k} (\mu^k| \Lambda^k) \\
& = \beta^\mathrm{pr} \ln \mathcal{N} (\mu^k | m_\mathrm{pr}^k, (\gamma_\mathrm{pr}^k \Lambda^k)^{-1}) + \beta^\mathrm{pr} \ln \mathcal{W} (\Lambda^k | W_\mathrm{pr}^k, \nu_\mathrm{pr}^k) \nonumber \\
& \quad + \beta \sum_{i=1}^N r_i^k \ln \mathcal{N} (y_i| \mu^k, (\Lambda^k)^{-1}) - \ln \mathcal{N} (\mu^k| m^k, (\gamma^k \Lambda^k)^{-1}) + \mathrm{const.} \\
& = \beta^\mathrm{pr} \frac{1}{2} \ln | \Lambda^k | - \beta^\mathrm{pr} \frac{1}{2} (\mu^k - m_\mathrm{pr}^k)^\intercal (\gamma _0^k \Lambda^k) (\mu^k - m_\mathrm{pr}^k) + \beta^\mathrm{pr} \frac{1}{2} ( \nu_\mathrm{pr}^k - D - 1) \ln | \Lambda^k | - \beta^\mathrm{pr} \frac{1}{2} \mathrm{Tr} [(W_\mathrm{pr}^k)^{-1} \Lambda^k] \nonumber \\
& \quad + \beta N^k \frac{1}{2} \ln \Lambda^k - \beta \frac{1}{2} \sum_{i=1}^N r_i^k (y_i - \mu^k) \Lambda^k (y_i - \mu^k) - \frac{1}{2} \ln | \Lambda^k | + \frac{1}{2} (\mu^k - m^k) (\gamma^k \Lambda^k) (\mu^k - m^k) + \mathrm{const.},
\end{align}
Thus, the posterior distribution of $\Lambda^k$ has the form
\begin{align}
\ln q_{t+1}^{\Lambda^k} (\Lambda^k) &= \frac{1}{2}(\nu^k - D -1) \ln | \Lambda^k | - \frac{1}{2} \mathrm{Tr} [(W^k)^{-1} \Lambda^k],
\end{align}
where
\begin{subequations}
\begin{align}
(W^k)^{-1} &= \beta^\mathrm{pr} \gamma_\mathrm{pr}^k (\mu^k - m_\mathrm{pr}^k) (\mu^k - m_\mathrm{pr}^k)^\intercal + \beta^\mathrm{pr} (W_\mathrm{pr}^k)^{-1} + \beta \sum_{i=1}^N r_i^k (y_i - \mu^k) (y_i - \mu^k)^\intercal - \gamma^k (\mu^k - m^k) (\mu^k - m^k)^\intercal \\
&= \beta^\mathrm{pr} \gamma_\mathrm{pr}^k (\mu^k - m_\mathrm{pr}^k) (\mu^k - m_\mathrm{pr}^k)^\intercal + \beta^\mathrm{pr} (W_\mathrm{pr}^k)^{-1} \nonumber \\
& \quad + \beta N^k S^k + \beta N^k (\bar{y}^k - \mu^k) (\bar{y} - \mu^k)^\intercal - \gamma^k (\mu^k - m^k) (\mu^k - m^k)^\intercal \\
&= \beta^\mathrm{pr} (W_\mathrm{pr}^k)^{-1} + \beta N^k S^k + \frac{\beta^\mathrm{pr} \beta \gamma_\mathrm{pr}^k N^k}{\gamma^k} (\bar{y}^k - m_\mathrm{pr}^k) (\bar{y}^k - m_\mathrm{pr}^k)^\intercal, \label{supp_eq_posterior_Lambda_031_051} \\
\nu^k &= \beta^\mathrm{pr} \nu_\mathrm{pr}^k + \beta N^k + (1 - \beta^\mathrm{pr}) D. \label{supp_eq_posterior_Lambda_031_061}
\end{align}
\end{subequations}
We have succeeded in transforming Eqs.~\eqref{supp_eq_posterior)_Lambda_021_051} and \eqref{supp_eq_posterior)_Lambda_021_061} into Eqs.~\eqref{supp_eq_posterior_Lambda_031_051} and \eqref{supp_eq_posterior_Lambda_031_061}.

\section{Posterior distribution in QAVB}

In this section, we derive the posterior distributions in the generalized version of QAVB.

\subsection{Posterior distribution of $\sigma_i$ in QAVB}

We have shown the update rule of $\sigma_i$ of VB in Eq.~\eqref{supp_eq_VB_E-step_sigma_001_002}.
In the case of the generalized version of QAVB, the update rule for $\sigma_i$ becomes
\begin{align}
\ln \hat{\rho}_{t+1}^{\sigma_i} &= \mathrm{Tr}_\theta \left[ \hat{\rho}_{t+1}^\theta \left\{ - \beta s^\mathrm{cl} \hat{H}_\mathrm{cl}^{\sigma_i| \theta} - \beta s^\Sigma \hat{H}_\mathrm{qu}^{\sigma_i} \right\} \right] + \mathrm{const.} \label{supp_eq_update_equation_QAVB_sigma_005_001}
\end{align}
where
\begin{align}
\hat{H}_\mathrm{cl}^{\sigma_i| \theta} &= \int_{\sigma_i \in S^\sigma} H_\mathrm{cl}^{\sigma_i| \theta} \hat{P}^{\sigma_i, \theta}, \\
H_\mathrm{cl}^{\sigma_i = k} &= - \ln \pi^k \mathcal{N} (y_i | \mu^k, (\Lambda^k)^{-1}),
\end{align}
$\hat{P}^{\sigma_i, \theta} \coloneqq \Ket{\sigma_i, \theta} \Bra{\sigma_i, \theta}$, and $S^\sigma = \{k\}_{k=1}^K$.
Taking the exponential of Eq.~\eqref{supp_eq_update_equation_QAVB_sigma_005_001}, we obtain
\begin{align}
\hat{\rho}_{t+1}^{\sigma_i} &\propto \exp \Big( \mathrm{Tr}_\theta \left[ \hat{\rho}_{t+1}^\theta \left\{ - \beta s^\mathrm{cl} \hat{H}_\mathrm{cl}^{\sigma_i| \theta} - \beta s^\Sigma \hat{H}_\mathrm{qu}^{\sigma_i} \right\} \right] \Big),
\end{align}
where the normalization condition is determined to satisfy the trace condition $\mathrm{Tr}_{\sigma_i} [\hat{\rho}^{\sigma_i}] = 1$.

\subsection{Posterior distribution of $\theta$ in QAVB}

Here, we write down the explicit formulas of posterior distributions of the GMM in the generalized version of QAVB.
As explained in Sec.~\ref{supp_sec_GMM_prior_posterior_001_001}, the Hamiltonian of the conjugate prior distribution of the GMM can be decomposed as
\begin{align}
H_\mathrm{pr}^\theta &= H_\mathrm{pr}^\pi + H_\mathrm{pr}^{\mu| \Lambda} + H_\mathrm{pr}^\Lambda.
\end{align}
Furthermore, due to the definition of the conjugate prior distributions, the Hamiltonians of the posterior distributions of QAVB have the same form
\begin{align}
H_\mathrm{cl}^\theta &= H_\mathrm{cl}^\pi + H_\mathrm{cl}^{\mu| \Lambda} + H_\mathrm{cl}^\Lambda.
\end{align}
In this section, we describe the update rules using the above Hamiltonian formulations.

\subsection{Definitions of Quantum states and operators of $\theta$}

For later convenience, we define the quantum states of $\theta = \{\pi, \mu, \Lambda\}$, $\pi$, $\mu$, and $\Lambda$ by $\Ket{\theta}$, $\Ket{\pi}$, $\Ket{\mu}$, and $\Ket{\Lambda}$ by $\Ket{\theta}$, $\Ket{\pi}$, $\Ket{\mu}$, and $\Ket{\Lambda}$, respectively.
These ket vectors satisfy
\begin{align}
\Ket{\theta} &= \Ket{\pi} \otimes \Ket{\mu} \otimes \Ket{\Lambda}.
\end{align}
Next, we define the projection operators of $\theta$, $\pi$, $\mu$, and $\Lambda$ denoted by $\hat{P}^\theta$, $\hat{P}^\pi$, $\hat{P}^\mu$, $\hat{P}^\Lambda$, respectively:
\begin{subequations}
\begin{align}
\hat{P}^\theta &= \Ket{\theta} \Bra{\theta}, \\
\hat{P}^\pi &= \Ket{\pi} \Bra{\pi}, \\
\hat{P}^\mu &= \Ket{\mu} \Bra{\mu}, \\
\hat{P}^\Lambda &= \Ket{\Lambda} \Bra{\Lambda}.
\end{align}
\end{subequations}
Similarly, we also define the identity operators of $\theta$, $\pi$, $\mu$, and $\Lambda$ by $\hat{I}^\theta$, $\hat{I}^\pi$, $\hat{I}^\mu$, $\hat{I}^\Lambda$, respectively:
\begin{subequations}
\begin{align}
\hat{I}^\theta &= \int d\theta \, \Ket{\theta} \Bra{\theta}, \\
\hat{I}^\pi &= \int d\pi \, \Ket{\pi} \Bra{\pi}, \\
\hat{I}^\mu &= \int d\mu \, \Ket{\mu} \Bra{\mu}, \\
\hat{I}^\Lambda &= \int d\Lambda \, \Ket{\Lambda} \Bra{\Lambda}.
\end{align}
\end{subequations}

We also denote, by $| \mu^k \rangle$ and $| \Lambda^k \rangle$, the quantum states of $\mu^k$ and $\Lambda^k$, respectively.
Then the projection operators of $\mu^k$ and $\Lambda^k$ are, respectively, given by
\begin{subequations}
\begin{align}
\hat{P}^{\mu_k} &= \Ket{\mu_k} \Bra{\mu_k}, \\
\hat{P}^{\Lambda_k} &= \Ket{\Lambda_k} \Bra{\Lambda_k}.
\end{align}
\end{subequations}
Furthermore, the identity operators of $\mu^k$ and $\Lambda^k$ reads, respectively,
\begin{subequations}
\begin{align}
\hat{I}^\mu &= \overset{K}{\underset{k=1}{\otimes}} \hat{I}^{\mu^k}, \\
\hat{I}^\Lambda &= \overset{K}{\underset{k=1}{\otimes}} \hat{I}^{\Lambda^k}.
\end{align}
\end{subequations}
Note that the following equalities hold:
\begin{subequations}
\begin{align}
\hat{P}^\mu &= \overset{K}{\underset{k=1}{\otimes}} \hat{P}^{\mu^k}, \\
\hat{P}^\Lambda &= \overset{K}{\underset{k=1}{\otimes}} \hat{P}^{\Lambda^k}.
\end{align}
\end{subequations}

\subsubsection{Posterior distribution of $\pi$ in QAVB}

The update equation of $\pi$ of the GMM in the case of the generalized version of QAVB reads
\begin{align}
\ln \hat{\rho}_{t+1}^{\pi} &= - \hat{H}_\mathrm{cl}^\pi - \beta s^\theta \hat{H}_\mathrm{qu}^\pi + \mathrm{const.}, \label{supp_eq_update_equation_QAVB_pi_005_001}
\end{align}
where
\begin{subequations}
\begin{align}
\hat{H}_\mathrm{cl}^\pi &\coloneqq \int_{S^\pi} d \pi \, H_\mathrm{cl}^\pi \hat{P}^{\Sigma, \theta}, \\
H_\mathrm{cl}^\pi &\coloneqq - \ln \Big( \mathcal{D} (\pi | \alpha) \Big), \\
\alpha^k &\coloneqq \beta^\mathrm{pr} (\alpha_\mathrm{pr}^k - 1) + \beta s^\mathrm{cl} N^k + 1.
\end{align}
\end{subequations}
We impose the commutation relation on $\hat{H}_\mathrm{qu}^\pi$, $[\hat{H}_\mathrm{qu}^\pi, \hat{I}^\Sigma \otimes \hat{\pi} \otimes \hat{I}^\mu \otimes \hat{I}^\Lambda] \ne 0$.
Here, $\hat{P}^\pi = \Ket{\pi} \Bra{\pi}$,
$\hat{\pi} = \int_{\pi \in S^\pi} d \pi \, \pi \hat{P}^\pi$, and $S^\pi$ is the domain of $\pi$.
Taking the exponential of Eq.~\eqref{supp_eq_update_equation_QAVB_pi_005_001}, we obtain
\begin{align}
\hat{\rho}_{t+1}^{\pi} &\propto \exp \Big( - \hat{H}_\mathrm{cl}^\pi - \beta s^\theta \hat{H}_\mathrm{qu}^\pi \Big).
\end{align}

\subsubsection{Posterior distributions of $\mu$ and $\Sigma$ in QAVB}

Similarly, the update rules of $\mu$ is written as
\begin{align}
\ln \hat{\rho}_{t+1}^{\mu^k| \Lambda^k} &= - \hat{H}_\mathrm{cl}^{\mu^k| \Lambda^k} - \beta s^\theta \hat{H}_\mathrm{qu}^{\mu^k} + \mathrm{const.}, \label{supp_eq_update_equation_QAVB_mu_005_001}
\end{align}
where
\begin{subequations}
\begin{align}
\hat{H}_\mathrm{cl}^{\mu^k| \Lambda^k} &\coloneqq \int_{\mu^k \in S^{\mu^k}} d \mu^k \, \int_{\Lambda^k \in S^{\Lambda^k}} d \Lambda^k \, H_\mathrm{cl}^{\mu^k| \Lambda^k} \hat{P}^{\Sigma, \theta}, \\
H_\mathrm{cl}^{\mu^k| \Lambda^k} &\coloneqq - \ln \Big( \mathcal{N} (\mu^k | m^k, (\gamma^k \Lambda^k)^{-1}) \Big), \\
\gamma^k &\coloneqq \beta^\mathrm{pr} \gamma_\mathrm{pr}^k + \beta s^\mathrm{cl} N^k,, \\
m^k &\coloneqq \frac{1}{\gamma^k} ( \beta^\mathrm{pr} \gamma_\mathrm{pr}^k m_\mathrm{pr}^k + \beta s^\mathrm{cl} N^k \bar{y}^k).
\end{align}
\end{subequations}
We impose the commutation relation on $\hat{H}_\mathrm{qu}^{\mu^k}$:
\begin{align}
\bigg[ \hat{H}_\mathrm{qu}^{\mu^k}, \hat{I}^\Sigma \otimes \hat{I}^\pi \otimes \bigg( \overset{k-1}{\underset{j=1}{\otimes}} \hat{I}^{\mu^j} \bigg) \otimes \hat{\mu}^k \otimes \bigg( \overset{K}{\underset{j=k+1}{\otimes}} \hat{I}^{\mu^j} \bigg) \otimes \hat{I}^\Lambda \bigg] \ne 0.
\end{align}
Here, $\hat{P}^{\mu^k, \Lambda^k} \coloneqq \Ket{\mu^k, \Lambda^k} \Bra{\mu^k, \Lambda^k}$,
with $\hat{\mu}^k \coloneqq \int_{\mu^k \in S^{\mu^k}} d \mu^k \, \mu^k \hat{P}^{\mu^k}$ with $\hat{P}^{\mu^k} \coloneqq \Ket{\mu^k} \Bra{\mu^k}$, and $S^{\mu^k}$ and $S^{\Lambda^k}$ are the domains of $\mu^k$ and $\Lambda^k$, respectively.
Taking the exponential of Eq.~\eqref{supp_eq_update_equation_QAVB_mu_005_001}, we obtain
\begin{align}
\hat{\rho}_{t+1}^{\mu^k| \Lambda^k} &\propto \exp \Big( - \hat{H}_\mathrm{cl}^{\mu^k| \Lambda^k} - \beta s^\theta \hat{H}_\mathrm{qu}^{\mu^k} \Big).
\end{align}

The update rule of $\Lambda$ is written as
\begin{align}
\ln \hat{\rho}_{t+1}^{\Lambda^k} &= - \hat{H}_\mathrm{cl}^{\Lambda^k} - \beta s^\theta \hat{H}_\mathrm{qu}^{\Lambda^k} + \mathrm{const.}, \label{supp_eq_update_equation_QAVB_Lambda_005_001}
\end{align}
where
\begin{subequations}
\begin{align}
\hat{H}_\mathrm{cl}^{\Lambda^k} &\coloneqq \int_{\Lambda^k \in S^{\Lambda^k}} d\Lambda^k \, H_\mathrm{cl}^{\Lambda^k} \hat{P}^{\Sigma, \theta}, \\
H_\mathrm{cl}^{\Lambda^k} &\coloneqq - \ln \Big( \mathcal{W} (\Lambda^k | W^k, \nu^k) \Big) \\
W^k &\coloneqq \beta^\mathrm{pr} (W_\mathrm{pr}^k)^{-1} + \beta s^\mathrm{cl} N^k S^k + \frac{\beta^\mathrm{pr} \beta s^\mathrm{cl} \gamma_\mathrm{pr}^k N^k}{\gamma^k} (\bar{y}^k - m_\mathrm{pr}^k) (\bar{y}^k - m_\mathrm{pr}^k)^\intercal, \\
\nu^k &\coloneqq \beta^\mathrm{pr} \nu_\mathrm{pr}^k + \beta s^\mathrm{cl} N^k + (1 - \beta^\mathrm{pr}) D.
\end{align}
\end{subequations}
Here $\hat{P}^{\Lambda^k} \coloneqq \Ket{\Lambda^k} \Bra{\Lambda^k}$, $\hat{\Lambda}^k \coloneqq \int_{\Lambda^k \in S^{\Lambda^k}} d \Lambda^k \, \Lambda^k \hat{P}^{\Lambda^k}$, $S^{\Lambda^k}$ is the domain of $\Lambda^k$, and $\hat{H}_\mathrm{qu}^{\Lambda^k}$ is a Hamiltonian that satisfies
\begin{align}
\bigg[ \hat{H}_\mathrm{qu}^{\mu^k}, \hat{I}^\Sigma \otimes \hat{I}^\pi \otimes \hat{I}^\mu \otimes \bigg( \overset{k-1}{\underset{j=1}{\otimes}} \hat{I}^{\Lambda^j} \bigg) \otimes \hat{\Lambda}^k \otimes \bigg( \overset{K}{\underset{j=k+1}{\otimes}} \hat{I}^{\Lambda^j} \bigg) \bigg] \ne 0.
\end{align}
Taking the exponential of Eq.~\eqref{supp_eq_update_equation_QAVB_Lambda_005_001}, we obtain
\begin{align}
\hat{\rho}_{t+1}^{\Lambda^k} &\propto \exp \Big( - \hat{H}_\mathrm{cl}^{\Lambda^k} - \beta s^\theta \hat{H}_\mathrm{qu}^{\Lambda^k} \Big).
\end{align}

\section{Quantization of $\theta$ in QAVB}

So far we have quantized $\Sigma$ and now we explain the generalized version of QAVB and address the reason why we do not use it.

\subsection{Generalized version of QAVB}

By adding $\hat{H}_\mathrm{qu}^\theta$ that satisfies $[\hat{H}_\mathrm{qu}^\theta, \hat{\theta}] \ne 0$ to Eq.~\eqref{main_Gibbs_operator_quantum_001_001} and introducing the inverse temperature for the prior Hamiltonian, we obtain the generalized Gibbs operator given by
\begin{align}
\hat{g} (\beta^\mathrm{pr}, \beta, s^\mathrm{cl}, s^\Sigma, s^\theta) &\coloneqq \exp \Big( - \beta^\mathrm{pr} \hat{H}_\mathrm{pr}^\theta - \beta \big( s^\mathrm{cl} \hat{H}_\mathrm{cl}^{\Sigma| \theta} + s^\Sigma \hat{H}_\mathrm{qu}^{\Sigma} + s^\theta \hat{H}_\mathrm{qu}^\theta \big) \Big). \label{supp_eq_generalized_Gibbs_operator_QAVB_001_001}
\end{align}
In Eq.~\eqref{supp_eq_generalized_Gibbs_operator_QAVB_001_001}, $\theta$ is also quantized in addition to $\Sigma$ since the noncommutative term on $\theta$ is added.
If we make use of Eq.~\eqref{supp_eq_generalized_Gibbs_operator_QAVB_001_001} instead of Eq.~\eqref{main_Gibbs_operator_quantum_001_001}, we can formulate the fully quantized version of QAVB.
Note that Eqs.~\eqref{main_Gibbs_operator_quantum_001_001} and \eqref{supp_eq_generalized_Gibbs_operator_QAVB_001_001} are identical when $s^\mathrm{cl} = 1 - s$, $s^\Sigma = s$, and $s^\theta = 0$.

If we use Eq.~\eqref{supp_eq_generalized_Gibbs_operator_QAVB_001_001} instead of Eq.~\eqref{main_Gibbs_operator_quantum_001_001} in QAVB, we can obtain the update equations of the fully quantized version of QAVB as follows.
The derivation of the update equations is unchanged whether or not $\theta$ is quantized; then the update equations for the operator are written as
\begin{align}
		\hat{\rho}_{t+1}^{\Sigma} &\propto \exp \Big( \mathrm{Tr}_{\theta} \Big[ \Big( \hat{I}^\Sigma \otimes \hat{\rho}_{t+1}^\theta \Big) \ln \hat{g} (\beta_t, s_t^\mathrm{cl}, s_t^\Sigma, s_t^{\theta}) \Big] \Big), \\
		\hat{\rho}_{t+1}^{\theta} &\propto \exp \Big( \mathrm{Tr}_{\Sigma} \Big[ \Big( \hat{\rho}_t^\Sigma \otimes \hat{I}^\theta \Big) \ln \hat{g} (\beta_t, s_t^\mathrm{cl}, s_t^\Sigma, s_t^{\theta}) \Big] \Big).
\end{align}
Here we note that the numerical cost still scales linearly with the number of data points $N$.
In this paper, we call this algorithm the generalized version of QAVB.

\subsection{Why we do not consider the generalized version of QAVB}

In QAVB, we choose $\beta^\mathrm{pr} = 1$, $s^\mathrm{cl} = 1-s$, $s^\Sigma = s$, and $s^\theta = 0$.
In this subsection, we state the reasons why we adopted $\beta^\mathrm{pr} = 1$ and $s^\theta = 0$.
First, if we set $\beta^\mathrm{pr} > 1$, the posterior distribution may break the necessary condition of the Wishart distribution in the case of the GMM.
That is the Wishart distribution can no longer act as a conjugate prior.
Thus, we set $\beta^\mathrm{pr} = 1$.
Second, when we set $s^\theta \ne 0$, we have to numerically solve the quantum Hamiltonian of parameters because we do not know the prior distributions associated with the quantum Hamiltonian of parameters and its computational cost is considered to be huge.

We have stated the reasons why we limited ourselves to the case of $\beta^\mathrm{pr} = 1$ and $s^\theta = 0$.
Fortunately, QAVB with $\beta^\mathrm{pr} = 1$ and $s^\theta = 0$ works well; so, we do not need to consider the general version of QAVB.

\section{QAVB in the path integral formulation} \label{supp_sec_Feynman_path-integral_001_001}

In general, quantum mechanics can also be formulated via the path integral formulation proposed by Feynman~\cite{Feynman_001}.
This section aims to reformulate QAVB by using the path integral formulation.

\subsection{The quantum relative entropy and update equations of QAVB in the path integral formulation}

So far, we have formulated QAVB in the operator formulation.
On the other hand, there exists another formulation for quantum mechanics, that is, Feynman's path integral formulation~\cite{Feynman_001, Suzuki_001, Takahashi_001, Takahashi_002}.
Here, we propose the path integral formulation of QAVB for a better understanding and a prerequisite of to compare with the work of Sato \textit{et al.} in Refs.~\cite{Sato_001, Sato_002}.

Using a variational function defined on paths $\{ \Sigma_j \}_{j=1}^M$ and $\{\theta_j\}_{j=1}^M$,
\begin{align}
\tilde{q}^{\Sigma, \theta} (\{ \Sigma_j \}_{j=1}^M, \{ \theta_j \}_{j=1}^M) &\coloneqq \prod_{j=1}^M \Braket{ \Sigma_j, \theta_j | \left[ \hat{\rho}^{\Sigma, \theta} \right]^\frac{1}{M} | \Sigma_{j-1}, \theta_{j-1}},
\end{align}
with $\Sigma_0 = \Sigma_M$ and $\theta_0 = \theta_M$,
we define the KL divergence on paths to be minimized in the path integral formulation of QAVB as follows:
\begin{align}
& \widetilde{\mathrm{KL}} \left( \tilde{q}^{\Sigma, \theta} (\{ \Sigma_j \}_{j=1}^M, \{ \theta_j \}_{j=1}^M) \middle\| \frac{\tilde{g} (\{\Sigma_j\}_{j=1}^M, \{\theta_j\}_{j=1}^M; \beta^\mathrm{pr}, \beta, s^\mathrm{cl}, s^\Sigma, s^\theta) }{\tilde{\mathcal{Z}} (\beta^\mathrm{pr}, \beta, s^\mathrm{cl}, s^\Sigma, s^\theta)} \right) \nonumber \\
& \quad \coloneqq - \sum_{\Sigma_{1}, \dots, \Sigma_M \in S^\Sigma} \Bigg[ \prod_{j=1}^M \int_{\theta_j \in S^\theta} d\theta_{j} \, \Bigg] \tilde{q}^{\Sigma, \theta} (\{ \Sigma_j \}_{j=1}^M, \{ \theta_j \}_{j=1}^M) \nonumber \\
& \qquad \times \Bigg[ \ln \tilde{q}^{\Sigma, \theta} (\{ \Sigma_j \}_{j=1}^M, \{ \theta_j \}_{j=1}^M) - \ln \frac{\tilde{g} (\{\Sigma_j\}_{j=1}^M, \{\theta_j\}_{j=1}^M; \beta^\mathrm{pr}, \beta, s^\mathrm{cl}, s^\Sigma, s^\theta) }{\tilde{\mathcal{Z}} (\beta^\mathrm{pr}, \beta, s^\mathrm{cl}, s^\Sigma, s^\theta)} \Bigg]. \label{supp_eq_KL_QAVB_path-integral_001_001}
\end{align}
Here, we have used
\begin{align}
&\tilde{g} (\{\Sigma_j\}_{j=1}^M, \{\theta_j\}_{j=1}^M; \beta^\mathrm{pr}, \beta, s^\mathrm{cl}, s^\Sigma, s^\theta) \nonumber \\
& \quad \coloneqq \prod_{j=1}^M \Braket{ \Sigma_j, \theta_{j} | \Big[ e^{ \frac{1}{M} \hat{K}_\mathrm{cl}^{\Sigma, \theta} (\beta^\mathrm{pr}, \beta, s^\mathrm{cl})} e^{ \frac{- \beta}{M} s^\Sigma \hat{H}_\mathrm{qu}^\Sigma } e^{ \frac{- \beta}{M} s^\theta \hat{H}_\mathrm{qu}^\theta } \Big] | \Sigma_{j-1}, \theta_{j-1} },
\end{align}
where
\begin{align}
  \hat{K}_\mathrm{cl}^{\Sigma, \theta} (\beta^\mathrm{pr}, \beta, s^\mathrm{cl}) &\coloneqq - \beta^\mathrm{pr} \hat{H}_\mathrm{pr}^\theta - \beta s^\mathrm{cl} \hat{H}_\mathrm{cl}^{\Sigma| \theta},
\end{align}
and
\begin{align}
\tilde{\mathcal{Z}} (\beta^\mathrm{pr}, \beta, s^\mathrm{cl}, s^\Sigma, s^\theta) &\coloneqq - \sum_{\Sigma_1, \dots, \Sigma_M \in S^\Sigma} \Bigg[ \prod_{j=1}^M \int_{\theta_j \in S^\theta} d\theta_{j} \, \Bigg] \tilde{g} (\{\Sigma_j\}_{j=1}^M, \{\theta_j\}_{j=1}^M; \beta^\mathrm{pr}, \beta, s^\mathrm{cl}, s^\Sigma, s^\theta), \label{supp_eq_partition-function_path-integral_001_021}
\end{align}
where $\Ket{\Sigma_0, \theta_{0} } = \Ket{ \Sigma_M, \theta_{M} }$ and $M$ represents the number of beads in the Trotter dimension.

Here we mention a property of the partition function in the path integral formulation, Eq.~\eqref{supp_eq_partition-function_path-integral_001_021}.
From Eq.~\eqref{supp_eq_generalized_Gibbs_operator_QAVB_001_001}, the exact partition function is computed as
\begin{align}
  \mathcal{Z} (\beta^\mathrm{pr}, \beta, s^\mathrm{cl}, s^\Sigma, s^\theta) &\coloneqq \mathrm{Tr}_{\Sigma, \theta} [\hat{g} (\beta^\mathrm{pr}, \beta, s^\mathrm{cl}, s^\Sigma, s^\theta)]. \label{supp_eq_partition-function_per-i_001_001}
\end{align}
Equation~\eqref{supp_eq_partition-function_path-integral_001_021} is an approximate form of the partition function and converges to the exact partition function with $M$ infinite.
That is, in the limit $M \rightarrow \infty$, Eq.~\eqref{supp_eq_partition-function_path-integral_001_021} leads to Eq.~\eqref{supp_eq_partition-function_per-i_001_001}:
\begin{align}
\mathcal{Z} (\beta^\mathrm{pr}, \beta, s^\mathrm{cl}, s^\Sigma, s^\theta) &= \lim_{M \rightarrow \infty} \tilde{\mathcal{Z}} (\beta^\mathrm{pr}, \beta, s^\mathrm{cl}, s^\Sigma, s^\theta).
\end{align}

As explained in the previous section, we rewrite minimization of Eq.~\eqref{supp_eq_KL_QAVB_path-integral_001_001} with the mean-field approximation as follows.
Taking the logarithm of Eq.~\eqref{supp_eq_partition-function_path-integral_001_021}, we define the free energy function in the path integral formulation as
\begin{align}
\tilde{\mathcal{F}} (\beta^\mathrm{pr}, \beta, s^\mathrm{cl}, s^\Sigma, s^\theta) &\coloneqq \ln \tilde{\mathcal{Z}} (\beta^\mathrm{pr}, \beta, s^\mathrm{cl}, s^\Sigma, s^\theta). \label{supp_eq_free-energy_001_096}
\end{align}
Note that Eq.~\eqref{supp_eq_free-energy_001_096} is a constant.
Next, we employ the mean-field approximation.
In the mean-field approximation of the path integral formulation, we assume a mean-field variational function $\tilde{q}^{\Sigma, \theta} (\{ \Sigma_j \}_{j=1}^M, \{ \theta_j \}_{j=1}^M) = \tilde{q}^{\Sigma}(\{ \Sigma_j \}_{j=1}^M) \tilde{q}^{\theta}(\{ \theta_{j} \}_{j=1}^M)$.
Thus the free energy function~\eqref{supp_eq_free-energy_001_096} can be decomposed into two parts:
\begin{align}
&\tilde{\mathcal{F}}(\beta^\mathrm{pr}, \beta, s^\mathrm{cl}, s^\Sigma, s^\theta) = \tilde{\mathcal{G}} \Big( \tilde{q}^{\Sigma}(\{ \Sigma_j \}_{j=1}^M) \tilde{q}^{\theta}(\{ \theta_{j} \}_{j=1}^M) \Big) \nonumber \\
& \quad + \widetilde{\mathrm{KL}} \left( \tilde{q}^{\Sigma}(\{ \Sigma_j \}_{j=1}^M) \tilde{q}^{\theta}(\{ \theta_{j} \}_{j=1}^M) \middle\| \frac{\tilde{g} (\{\Sigma_j\}_{j=1}^M, \{\theta_j\}_{j=1}^M; \beta^\mathrm{pr}, \beta, s^\mathrm{cl}, s^\Sigma, s^\theta) }{\tilde{\mathcal{Z}} (\beta^\mathrm{pr}, \beta, s^\mathrm{cl}, s^\Sigma, s^\theta)} \right),
\end{align}
with
\begin{align}
&\tilde{\mathcal{G}} \Big( \tilde{q}^{\Sigma}(\{ \Sigma_j \}_{j=1}^M) \tilde{q}^{\theta}(\{ \theta_{j} \}_{j=1}^M) \Big) \nonumber \\
& \quad \coloneqq \tilde{\mathcal{G}}^{\mathrm{cl}} \Big( \tilde{q}^{\Sigma}(\{ \Sigma_j \}_{j=1}^M) \tilde{q}^{\theta}(\{ \theta_{j} \}_{j=1}^M) \Big) + \tilde{\mathcal{G}}^\Sigma \Big( \tilde{q}^{\Sigma}(\{ \Sigma_j \}_{j=1}^M) \Big) + \tilde{\mathcal{G}}^{\theta} \Big( \tilde{q}^{\theta}(\{ \theta_{j} \}_{j=1}^M) \Big), \label{supp_eq_functional_xxx_001_001}
\end{align}
where
\begin{align}
&\tilde{\mathcal{G}}^{\mathrm{cl}} \Big( \tilde{q}^{\Sigma}(\{ \Sigma_j \}_{j=1}^M) \tilde{q}^{\theta}(\{ \theta_{j} \}_{j=1}^M) \Big) \nonumber \\
& \quad \coloneqq \sum_{\Sigma_1, \dots, \Sigma_M \in S^\Sigma} \Bigg[ \prod_{j=1}^M \int_{\theta_j \in S^\theta} d\theta_{j} \, \Bigg] \tilde{q}^\Sigma ( \{ \Sigma_j \}_{j=1}^M ) \tilde{q}^{\theta}(\{ \theta_{j} \}_{j=1}^M) \nonumber \\
& \qquad \times \Bigg( \sum_{j=1}^M \frac{1}{M} \Braket{\Sigma_j, \theta_j | \hat{K}_\mathrm{cl}^{\Sigma, \theta} (\beta^\mathrm{pr}, \beta, s^\mathrm{cl}) | \Sigma_j, \theta_j} - \ln \tilde{q}^{\Sigma}(\{ \Sigma_j \}_{j=1}^M) - \ln \tilde{q}^{\theta}(\{ \theta_{j} \}_{j=1}^M) \Bigg),
\end{align}
\begin{align}
\tilde{\mathcal{G}}^\Sigma \Big( \tilde{q}^{\Sigma}(\{ \Sigma_j \}_{j=1}^M) \Big) &\coloneqq \sum_{\Sigma_1, \dots, \Sigma_M \in S^\Sigma} \tilde{q}^{\Sigma}(\{ \Sigma_j \}_{j=1}^M) \sum_{j=1}^M \ln \Braket{ \Sigma_j | \Big[ e^{\frac{- \beta}{M} s^\Sigma \hat{H}_\mathrm{qu}^\Sigma } \Big] | \Sigma_{j-1} },
\end{align}
and
\begin{align}
\tilde{\mathcal{G}}^{\theta} \Big( \tilde{q}^{\theta}(\{ \theta_{j} \}_{j=1}^M) \Big) &\coloneqq \Bigg[ \prod_{j=1}^M \int_{\theta_j \in S^\theta} d\theta_{j} \, \Bigg] \tilde{q}^{\theta}(\{ \theta_{j} \}_{j=1}^M) \sum_{j=1}^M \ln \Braket{\theta_j | \Big[ e^{\frac{- \beta}{M} s^\theta \hat{H}_\mathrm{qu}^\theta} \Big] | \theta_{j - 1}}.
\end{align}
Note that minimization of Eq.~\eqref{supp_eq_KL_QAVB_path-integral_001_001} is identical to maximization of Eq.~\eqref{supp_eq_functional_xxx_001_001}.

Then, taking the functional derivative of Eq.~\eqref{supp_eq_KL_QAVB_path-integral_001_001} with respect to $\tilde{q}^{\Sigma}(\{ \Sigma_j \}_{j=1}^M)$ and $\tilde{q}^{\theta}(\{ \theta_{j} \}_{j=1}^M)$, we obtain the update equations of QAVB in the path integral formulation,
\begin{align}
&\tilde{q}_{t+1}^\Sigma(\{ \Sigma_j \}_{j=1}^M) \nonumber \\
& \, \propto \exp \Bigg( \Bigg[ \prod_{j=1}^M \int_{\theta_j \in S^\theta} d\theta_{j} \, \Bigg] \tilde{q}_{t+1}^{\theta}(\{ \theta_{j} \}_{j=1}^M) \sum_{j=1}^M \frac{1}{M} \Braket{\Sigma_j, \theta_j | \hat{K}_\mathrm{cl}^{\Sigma, \theta} (\beta^\mathrm{pr}, \beta, s^\mathrm{cl}) | \Sigma_j, \theta_j} \Bigg) \nonumber \\
& \quad \times \prod_{j=1}^M \Braket{ \Sigma_j | \Big[ e^{\frac{- \beta }{M} s^\Sigma \hat{H}_\mathrm{qu}^\Sigma} \Big] | \Sigma_{j-1} }, \label{supp_eq_E-step_path-integral_001_001}
\end{align}
and
\begin{align}
&\tilde{q}_{t+1}^{\theta}(\{ \theta_{j} \}_{j=1}^M) \nonumber \\
& \, \propto \exp \Bigg( \sum_{\Sigma_1, \dots, \Sigma_M \in S^\Sigma} \tilde{q}_t^\Sigma(\{ \Sigma_j \}_{j=1}^M) \sum_{j=1}^M \frac{1}{M} \Braket{\Sigma_j, \theta_j | \hat{K}_\mathrm{cl}^{\Sigma, \theta} (\beta^\mathrm{pr}, \beta, s^\mathrm{cl}) | \Sigma_j, \theta_j} \Bigg) \nonumber \\
& \quad \times \prod_{j=1}^M \Braket{\theta_j | \Big[ e^{\frac{- \beta}{M} s^\theta \hat{H}_\mathrm{qu}^\theta} \Big] | \theta_{j - 1}}, \label{supp_eq_M-step_path-integral_001_001}
\end{align}
where $t$ denotes the number of iterations, $\Sigma_0 = \Sigma_M$ and $\theta_0 = \theta_M$ are satisfied, and normalization constant is determined such that Eqs.~\eqref{supp_eq_E-step_path-integral_001_001} and \eqref{supp_eq_M-step_path-integral_001_001} satisfy the condition of probability distributions on paths.
In practical calculations, we can evaluate Eqs.~\eqref{supp_eq_E-step_path-integral_001_001} and \eqref{supp_eq_M-step_path-integral_001_001} for any models by performing quantum Monte Carlo methods~\cite{Landau01, Newman01, Gubernatis01}.

So far, we have derived the update equations of QAVB in the path integral formulation.
In the end of this section, we show that update equations in the two formulations are identical.
Namely, we prove that Eqs.~\eqref{supp_eq_E-step_path-integral_001_001} and \eqref{supp_eq_M-step_path-integral_001_001} are equivalent to Eqs.~\eqref{main_QAVB_E-step_001_001} and \eqref{main_QAVB_M-step_001_001} as follows.
By relabeling the index $j$ in $\{ \theta_j \}_{j=1}^M$ and using $\left[ \prod_{j=1}^{M-1} \int_{\theta_j \in S^\theta} d\theta_{j} \, \right] \tilde{q}_{t+1}^{\theta}(\{ \theta_{j} \}_{j=1}^M) = \Braket{\theta_M | \hat{\rho}_{t+1}^\theta | \theta_M}$, Eq.~\eqref{supp_eq_E-step_path-integral_001_001} becomes
\begin{align}
&\tilde{q}_{t+1}^\Sigma(\{ \Sigma_j \}_{j=1}^M) \nonumber \\
& \quad \propto \exp \Bigg( \Bigg[ \prod_{j=1}^M \int_{\theta_j \in S^\theta} d\theta_{j} \, \Bigg] \tilde{q}_{t+1}^{\theta}(\{ \theta_{j} \}_{j=1}^M) \sum_{j=1}^M \frac{1}{M} \Braket{\Sigma_j, \theta_j | \hat{K}_\mathrm{cl}^{\Sigma, \theta} (\beta^\mathrm{pr}, \beta, s^\mathrm{cl}) | \Sigma_j, \theta_j} \Bigg) \nonumber \\
& \qquad \times \prod_{j=1}^M \Braket{ \Sigma_j | \Big[ e^{\frac{- \beta }{M} s^\Sigma \hat{H}_\mathrm{qu}^\Sigma} \Big] | \Sigma_{j-1} } \\
& \quad = \exp \Bigg( \int_{\theta_M \in S^\theta} d\theta_{M} \, \Braket{\theta_M | \hat{\rho}_{t+1}^\theta | \theta_M} \sum_{j=1}^M \frac{1}{M} \Braket{\Sigma_j, \theta_j | \hat{K}_\mathrm{cl}^{\Sigma, \theta} (\beta^\mathrm{pr}, \beta, s^\mathrm{cl}) | \Sigma_j, \theta_j} \Bigg) \nonumber \\
& \qquad \times \prod_{j=1}^M \Braket{ \Sigma_j | \Big[ e^{\frac{- \beta }{M} s^\theta \hat{H}_\mathrm{qu}^\Sigma} \Big] | \Sigma_{j-1} }.
\end{align}
We repeat the same calculation for Eq.~\eqref{supp_eq_M-step_path-integral_001_001}:
\begin{align}
&\tilde{q}_{t+1}^{\theta}(\{ \theta_{j} \}_{j=1}^M) \nonumber \\
& \quad \propto \exp \Bigg( \sum_{\Sigma_1, \dots, \Sigma_M \in S^\Sigma} \tilde{q}_t^\Sigma(\{ \Sigma_j \}_{j=1}^M) \sum_{j=1}^M \frac{1}{M} \Braket{\Sigma_j, \theta_j | \hat{K}_\mathrm{cl}^{\Sigma, \theta} (\beta^\mathrm{pr}, \beta, s^\mathrm{cl}) | \Sigma_j, \theta_j} \Bigg) \nonumber \\
& \qquad \times \prod_{j=1}^M \Braket{\theta_j | \Big[ e^{\frac{- \beta}{M} s^\theta \hat{H}_\mathrm{qu}^\theta} \Big] | \theta_{j - 1}} \label{supp_eq_update_equation_theta_path-integral_001_003} \\
& \quad = \exp \Bigg( \sum_{\Sigma_M \in S^\Sigma} \Braket{\Sigma_M | \hat{\rho}_t^\Sigma | \Sigma_M} \sum_{j=1}^M \frac{1}{M} \Braket{\Sigma_j, \theta_j | \hat{K}_\mathrm{cl}^{\Sigma, \theta} (\beta^\mathrm{pr}, \beta, s^\mathrm{cl}) | \Sigma_j, \theta_j} \Bigg) \nonumber \\
& \qquad \times \prod_{j=1}^M \Braket{\theta_j | \Big[ e^{\frac{- \beta}{M} s^\theta \hat{H}_\mathrm{qu}^\theta} \Big] | \theta_{j - 1}}. \label{supp_eq_update_equation_theta_path-integral_001_004}
\end{align}
Here, we have used $\sum_{\Sigma_1, \dots, \Sigma_{M-1} \in S^\Sigma} \tilde{q}_t^\Sigma(\{ \Sigma_j \}_{j=1}^M) = \Braket{\Sigma_M | \hat{\rho}_t^\Sigma | \Sigma_M}$ from Eq.~\eqref{supp_eq_update_equation_theta_path-integral_001_003} to Eq.~\eqref{supp_eq_update_equation_theta_path-integral_001_004}.
Marginalizing $\Sigma_1, \dots, \Sigma_{M-1}$ in Eq.~\eqref{supp_eq_update_equation_theta_path-integral_001_003} and $\theta_1, \dots, \theta_{M-1}$ in Eq.~\eqref{supp_eq_update_equation_theta_path-integral_001_004} leads to Eqs.~\eqref{supp_eq_E-step_path-integral_001_001} and \eqref{supp_eq_M-step_path-integral_001_001}.
Therefore, we conclude the update equations in the operator formulation and the path integral formulation are equivalent.

\subsection{A prior attempt at designing a QA extension of VB using path integral formulation} \label{supp_sec_discussion_Sato_002_001}

An initial attempt was made in 2010 by Sato \textit{et al.}~\cite{Sato_001, Sato_002} to follow the path integral approach and create a QA extension of VB.
However, the objective function used in Ref.~\cite{Sato_001, Sato_002} does not equal the VB objective function in the classical limit.
Thus, their algorithm does not minimize the VB objective function and is not a proper quantum extension of VB.
In fact, in Refs.~\cite{Miyahara_002} the errors committed by Sato \textit{et al.}~\cite{Sato_001, Sato_002} have been pointed, and one correct way of deriving the updates that \cite{Sato_001, Sato_002} had intended has been formulated.
On the other hand, the operator-based QAVB algorithm we analyze in this paper exactly corresponds to VB in the classical limit.

The main focus of this paper is to explain why and how the QAVB algorithm --based on the operator formulation of quantum mechanics -- works and the main mechanisms for a quantum advantage.
The dynamics of the correct path-integral based QAVB algorithms might be different from the QAVB analyzed in this paper, and we leave the analysis of the dynamics of such alternate algorithms as future work.

\bibliography{paper_qavb_mechanism_999_001_dropbox}

\end{document}